\documentclass{article}
\pdfoutput=1
\usepackage[utf8]{inputenc} % allow utf-8 input
\usepackage[T1]{fontenc}    % use 8-bit T1 fonts
\usepackage{url}            % simple URL typesetting
\usepackage{booktabs}       % professional-quality tables
\usepackage{amsfonts}       % blackboard math symbols
\usepackage{nicefrac}       % compact symbols for 1/2, etc.
\usepackage{microtype}      % microtypography
\usepackage{graphicx}
\usepackage{subcaption}
\usepackage{booktabs} % for professional tables
\usepackage[square,sort,comma,numbers]{natbib}

\usepackage{here}
\usepackage{graphicx}
\usepackage{caption}
\usepackage[ruled,noend]{algorithm2e}
\usepackage{amsmath,amssymb,amsfonts,amsbsy,amsfonts,latexsym}
\usepackage{amsthm}
\usepackage{multirow}
\usepackage{makecell}
\usepackage[labelfont=bf,textfont=it,belowskip=0pt,aboveskip=5pt,tableposition=top]{caption}
\usepackage{xcolor}
\usepackage{colortbl}
\usepackage{multirow}

\definecolor{colorA}{RGB}{189,201,225}
\definecolor{colorB}{RGB}{103,169,207}
\definecolor{colorC}{RGB}{ 28,144,153}
\definecolor{colorD}{RGB}{  1,108, 89}

\newcolumntype{R}{>{\columncolor{gray!40}}r}
\newcolumntype{L}{>{\columncolor{gray!40}}l}
\newcolumntype{C}{>{\columncolor{gray!40}}c}

\usepackage{tabularx,colortbl,xcolor}
\usepackage{multirow}
\usepackage[normalem]{ulem}
\useunder{\uline}{\ul}{}

\newcommand\Ga{}

\newcommand\Gc{\rowcolor{gray!30}}
\newcommand\Gd{\rowcolor{gray!45}}

\usepackage{enumitem}

\usepackage{xparse}% http://ctan.org/pkg/xparse

\graphicspath{{figs}}
\DeclareGraphicsExtensions{.pdf,.png}
\usepackage{pgfplots, pgfplotstable}

\usepackage[multidot]{grffile}
\usepackage[final,nonatbib]{nips_2018}
 \usepackage{hyperref}       % hyperlinks

\DeclareMathOperator{\sign}{sign}

\newcommand{\M}{\mathcal M}
\newcommand{\J}{\mathcal J}
\newcommand{\B}{\mathcal B}

\def\B{{\bf B}}

\def\g{{\bf g}}

\def\H{{\bf H}}

\def\x{{\bf x}}

\def\y{{\bf y}}

\def\0{{\bf 0}}
\def\1{{\bf 1}}

\def\M{\mathcal M}
\def\D{\mathcal D}

\def\tha{\mbox{\boldmath$\theta$\unboldmath}}

\newtheorem{theorem}{Theorem}

\newtheorem{lemma}[theorem]{Lemma}
\newtheorem{proposition}[theorem]{Proposition}

\newtheorem{assumption}{Assumption}

\newcommand{\nexp}{\text{\sc{e}-}} % Exponent notation

\NewDocumentCommand{\var}{O{s} m O{}}{%
  \ensuremath{#1_{#2}^{#3}}% add \vphantom{<bizarre sup>}
}
\usepackage{siunitx}

\newcommand{\secref}[1]{\S\ref{#1}}

\newcommand{\p} {\partial}

\definecolor{light-gray}{gray}{0.80}

\newcommand{\mrowrot}[2]{
\parbox[t]{2mm}{\multirow{#1}{*}{\rotatebox[origin=c]{90}{#2}}}
}

\renewcommand\paragraph{\subsubsection*}

\newcommand\eref{Eq.~\ref}
\newcommand\fref{Fig.~\ref}
\newcommand\tref{Table~\ref}
\SetKwInOut{Parameter}{parameter}

\newcommand\eps{ \epsilon}
\newcommand\MS{{\mathcal{M}}}
\def\x{{\bf x}}
\def\H{{\bf H}}
\def\g{{\bf g}}

\def\0{{\bf 0}}
\def\s{{\bf s}}

\def\p{{\bf p}}

\def\R{{\mathbb R}}

\newcommand*\samethanks[1][\value{footnote}]{\footnotemark[#1]}
\newcommand\atsign{@}

\title{Hessian-based Analysis of Large Batch Training and Robustness to Adversaries}

\author{Zhewei Yao$^1$\thanks{Equal contribution}~~~Amir Gholami$^1$\samethanks~~~Qi Lei$^2$~~~Kurt Keutzer$^1$~~~Michael W. Mahoney$^1$\\
$^1$ University of California at Berkeley, \{zheweiy, amirgh, keutzer and mahoneymw\}\atsign berkeley.edu\\
$^2$ University of Texas at Austin, leiqi\atsign ices.utexas.edu
}

\begin{document}
% \nipsfinalcopy is no longer used

\maketitle

\begin{abstract}
Large batch size training of Neural Networks has been shown to incur accuracy
loss when trained with the current methods.  The exact underlying reasons for
this are still not completely understood.  Here, we study large batch size
training through the lens of the Hessian operator and robust optimization. In
particular, we perform a Hessian based study to analyze exactly how the landscape of
the loss function changes when training with large batch size. We compute the
true Hessian spectrum, without approximation, by back-propagating the second
derivative. Extensive experiments on multiple networks show that saddle-points are
not the cause for generalization gap of large batch size training, and the results
consistently show that large batch converges to points with noticeably higher Hessian
spectrum. Furthermore, we show that robust training allows one to favor flat areas,
as points with large Hessian spectrum show poor robustness to adversarial perturbation.
We further study this relationship, and provide empirical and theoretical proof that the inner loop for robust training is a saddle-free optimization problem \textit{almost everywhere}.
We present detailed experiments with five
different network architectures, including a residual network, tested on MNIST, CIFAR-10, and CIFAR-100
datasets.
We have open sourced our method which can be accessed at~\cite{HFcode}.

\end{abstract}

\vspace{-2mm}
\section{Introduction}\label{sec:intro}

During the training of a Neural Network (NN), we are given a set of 
input data $\x$ with the corresponding labels $y$
drawn from an unknown distribution $\mathcal{P}$. In practice, we
only observe a set of discrete examples drawn from $\mathcal{P}$, and train
the NN to learn this unknown distribution.
This is typically a non-convex optimization problem, in which the choice of hyper-parameters
would highly affect the convergence properties. In particular, it has been observed that
using large batch size for training often results in convergence to points with poor convergence
properties. The main motivation for using large batch is the increased opportunities for data parallelism
which can be used to reduce training time~\cite{goyal2017accurate,gholami2017integrated}.
Recently, there have been several works that have proposed different methods to avoid the performance
loss with large batch~\cite{goyal2017accurate,smith2017don,you2017scaling}. However, these
methods do not work for all networks and datasets. This has motivated us to revisit the original problem
and study how the optimization with large batch size affects the convergence behavior.

% ---------------------------------------------%
% ---------------------------------------------%
\begin{figure*}[tbp]
\begin{center}
  \includegraphics[width=.4\textwidth]{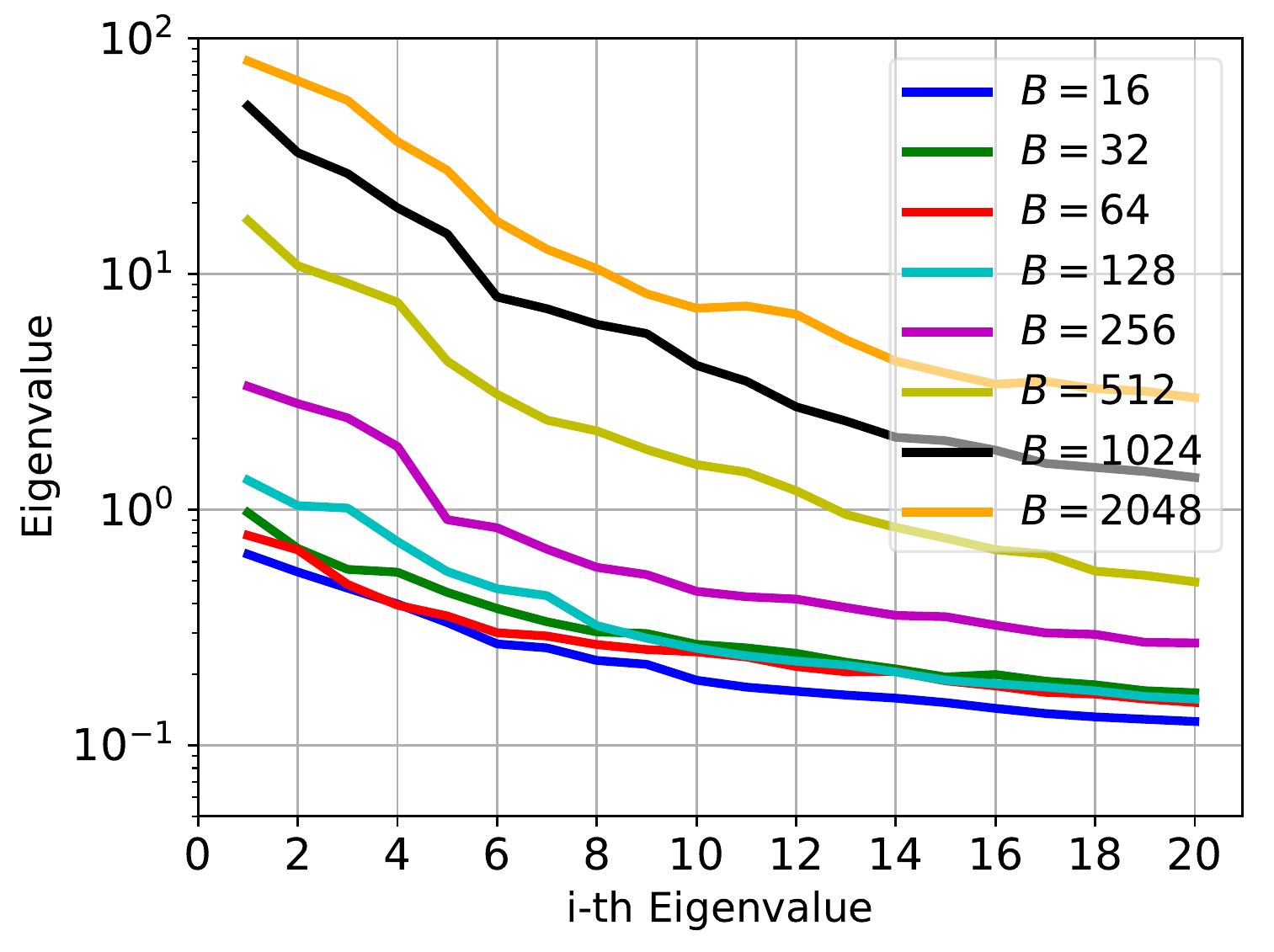}
  \includegraphics[width=.4\textwidth]{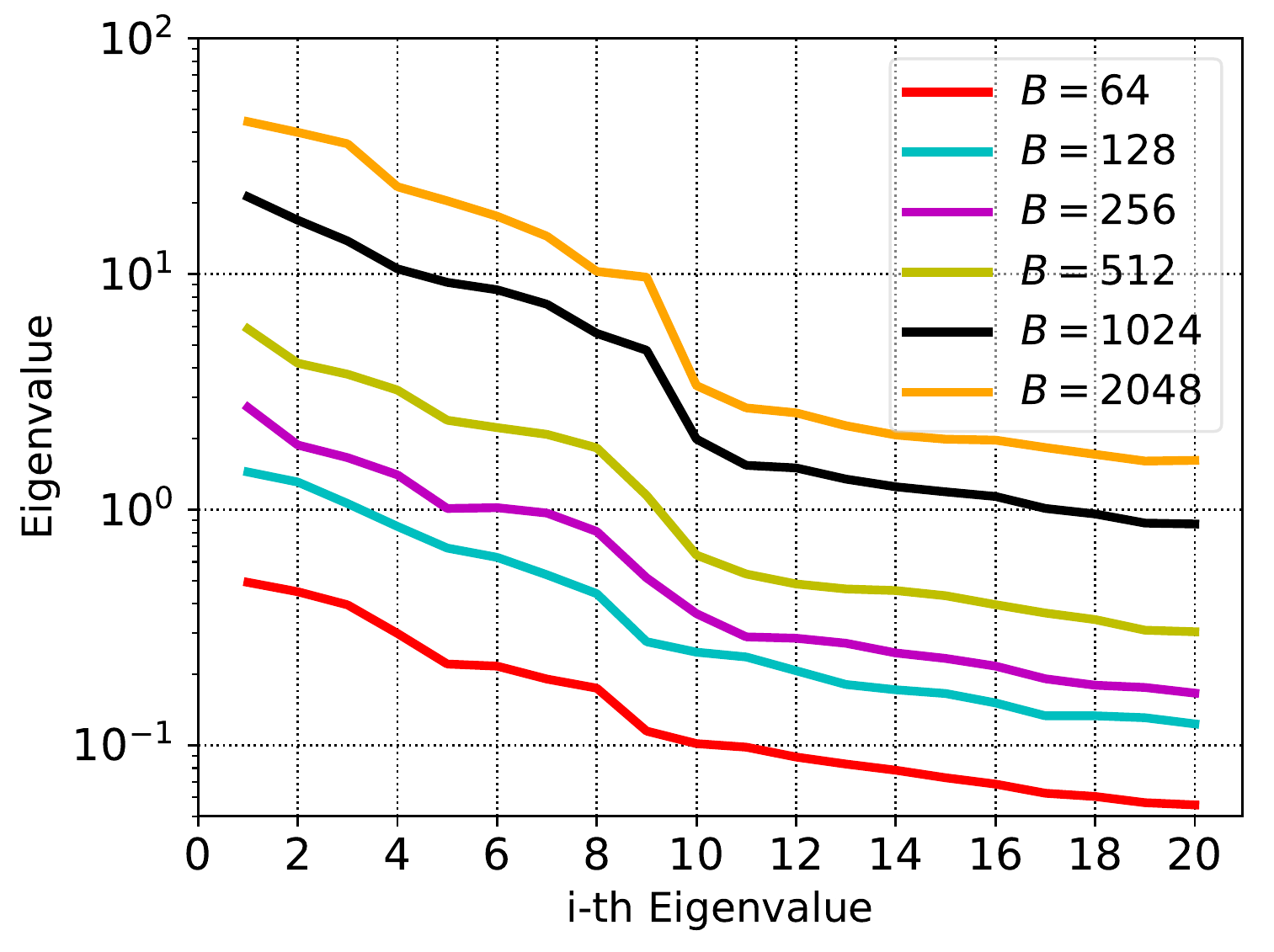}
\end{center}
\caption{
  Top 20 eigenvalues of the Hessian is shown for C1 on CIFAR-10 (left) and M1 on MNIST (right) datasets.
  The spectrum is computed using power iteration with relative error of $1\nexp{4}$.
}
\label{f:top20eigs}
\end{figure*}
% ---------------------------------------------%
% ---------------------------------------------%

We first start by analyzing how the Hessian spectrum and gradient change during training for small
batch and compare it to large batch size and then draw connection with robust training.

In particular, we aim to answer the following questions:
\noindent\textbf{Q1} 

How is the training for large batch size different than small batch size? Equivalently,
what is the difference between the local geometry of the neighborhood that the model converges
when large batch size is used as compared to small batch?

\begin{figure*}[tbp]
\begin{center}
  \includegraphics[width=.245\textwidth]{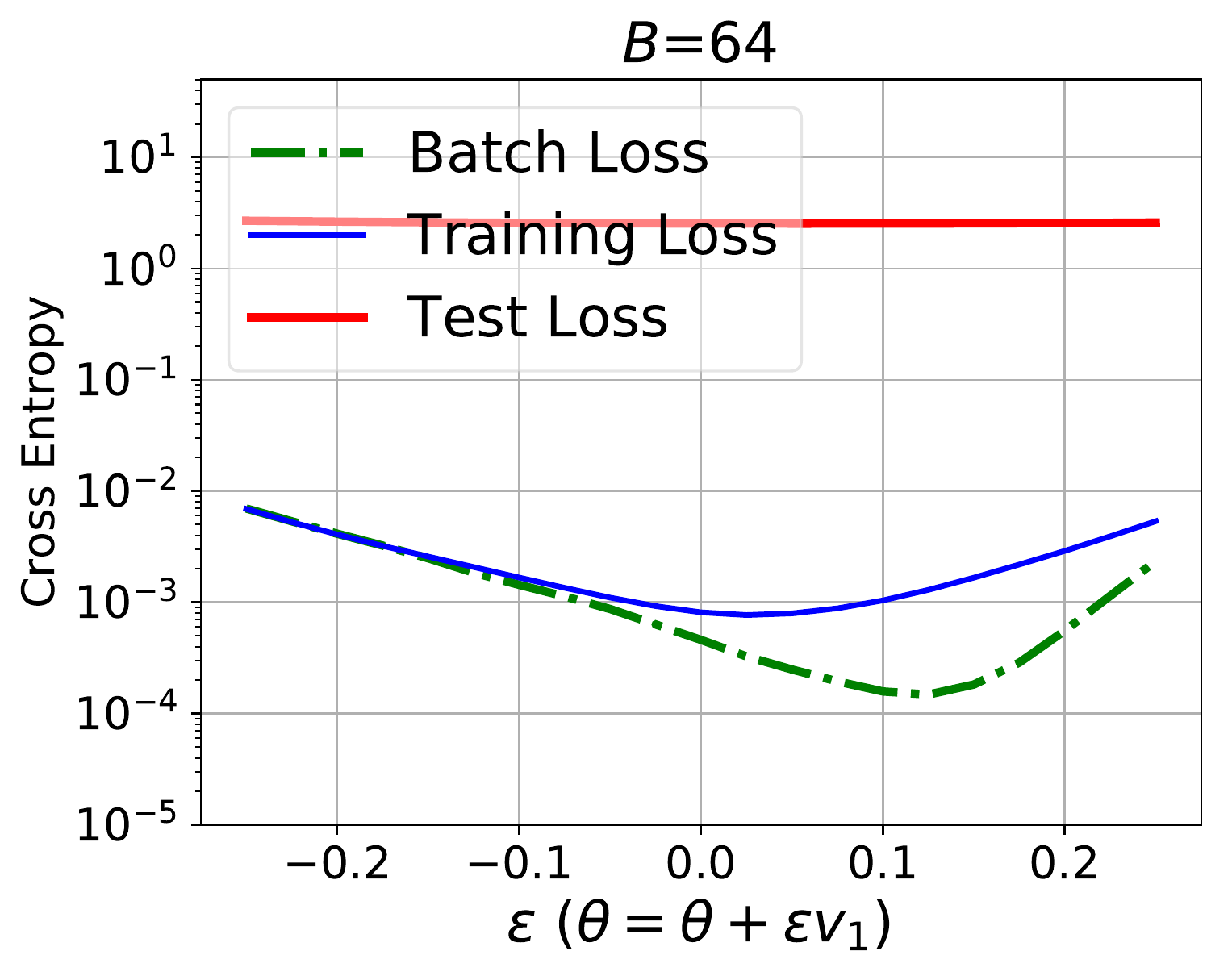}
  \includegraphics[width=.245\textwidth]{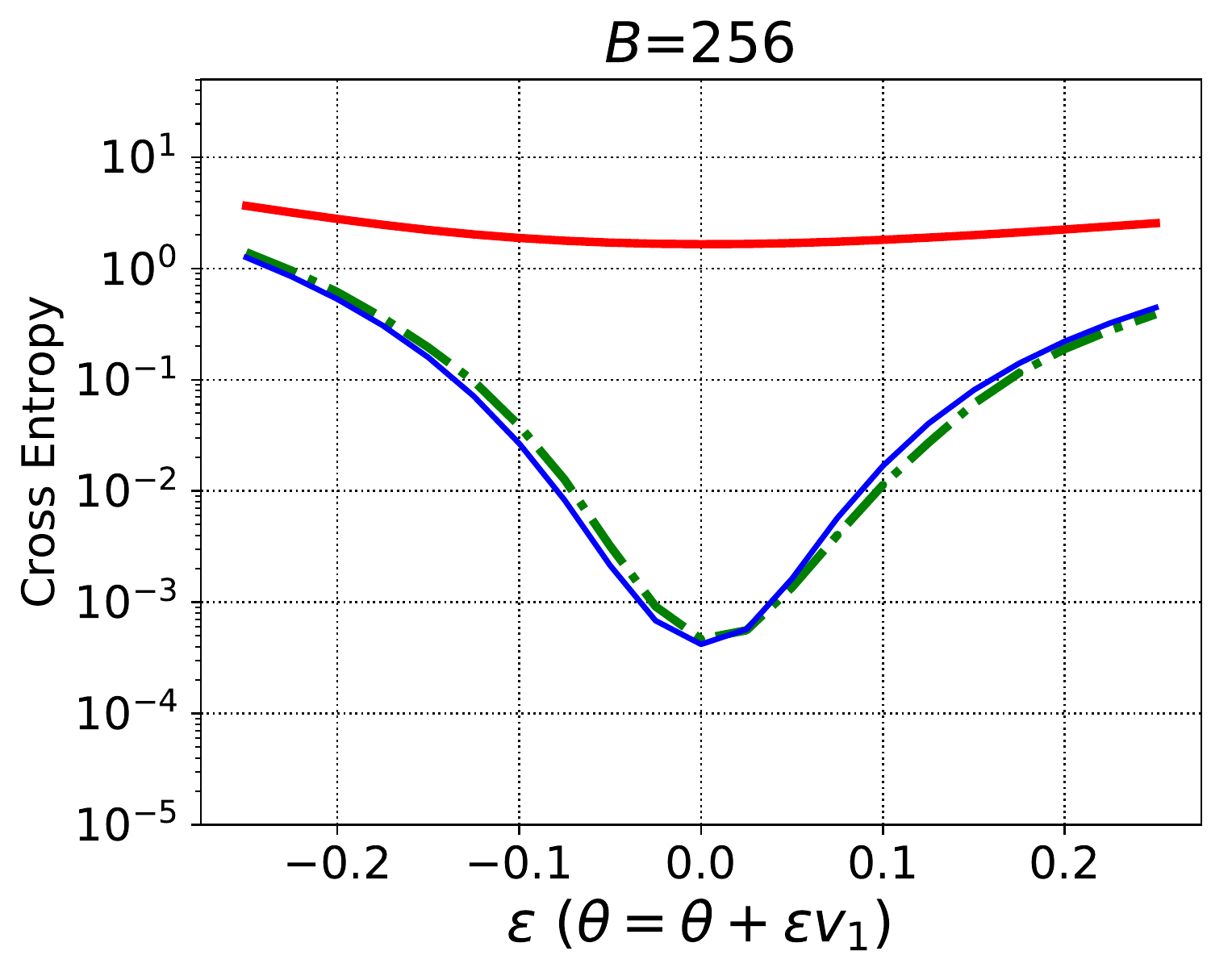}
  \includegraphics[width=.245\textwidth]{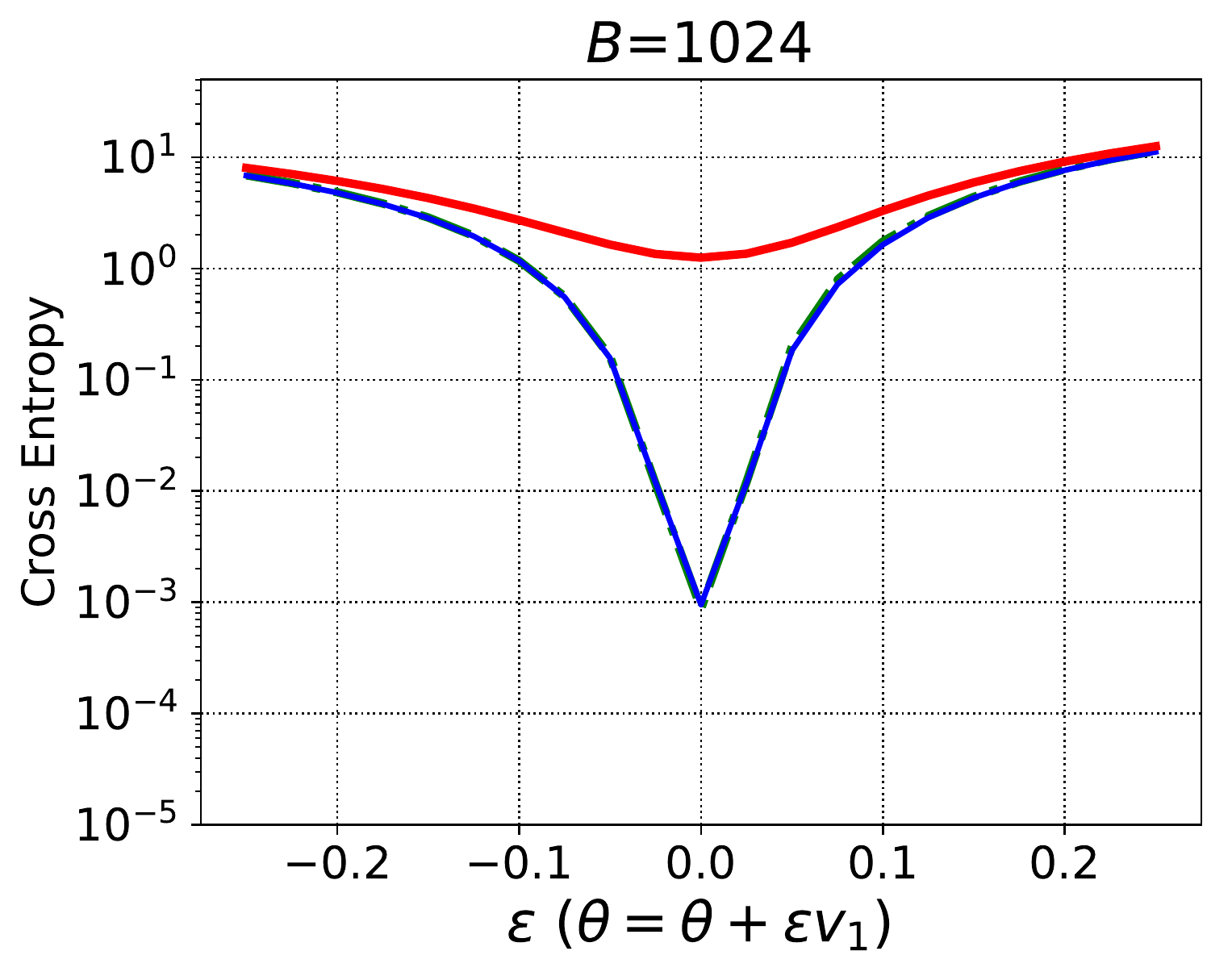}
  \includegraphics[width=.245\textwidth]{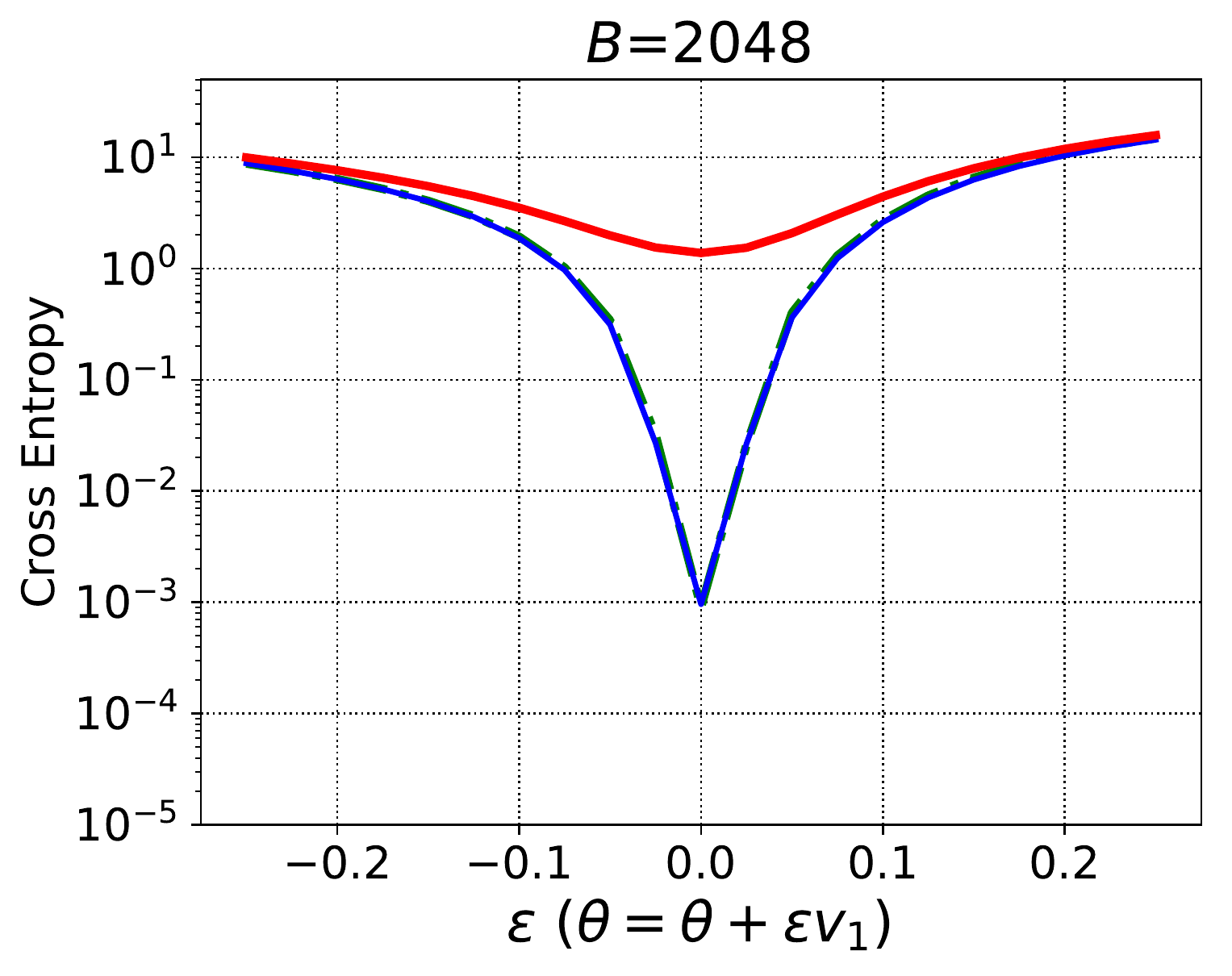}\\
\end{center}
\caption{
  The landscape of the loss is shown along the dominant eigenvector, $v_1$, of the Hessian for C1 on CIFAR-10 dataset.
  Here $\epsilon$ is a scalar that perturbs the model parameters along $v_1$.
}
\label{f:landscape_largebatch_qalex}
\end{figure*}

\textbf{A1}
We backpropagate the second-derivative and compute its spectrum during training. The results show that
despite the arguments regarding prevalence of saddle-points plaguing optimization~\cite{dauphin2014identifying,ge2015escaping},
that is actually not the problem with large batch size training, even when batch size is increased to the gradient descent limit.
In~\cite{flat}, an approximate numerical method was used to approximate the maximal eigenvalue at a point. Here,
by directly computing the spectrum of the true Hessian, we show that large batch size progressively
gets trapped in areas with noticeably larger spectrum (and not just the dominant eigenvalue).
For details please see~\secref{sec:large_batch}, especially Figs.~\ref{f:top20eigs},~\ref{f:landscape_largebatch_qalex} and~\ref{fig:qalex_h_logger}.

\noindent\textbf{Q2} 
What is the connection between robust optimization and large batch size training?
Equivalently, how does the batch size affect the robustness of the model to adversarial perturbation?

\noindent\textbf{A2} 
We show that robust optimization is antithetical to large batch training, in the sense that it favors
areas with small spectrum (aka flat minimas). We show that points converged with large batch size
are significantly more prone to adversarial attacks as compared
to a model trained with small batch size.
Furthermore, we show that robust training progressively favors the opposite, leading
to points with flat spectrum and robust to adversarial perturbation.
We provide empirical and theoretical proof that the inner loop of the robust optimization,
where we find the worst case, is a saddle-free optimization problem \textit{almost everywhere}.
Details are discussed in~\secref{sec:large_batch_adv}, especially Table~\ref{t:qalex_train_batch}, ~\ref{t:lenet_train_batch} and Figs.~\ref{fig:qalex_h_logger}, \ref{fig:cifar_adv_topeig}.

\textbf{Limitations:}
We believe it is critical for every paper to clearly state limitations. In this
work, we have made an effort to avoid reporting just the best results, and
repeated all the experiments at least three times and found all the findings to
be consistent. Furthermore, we performed the tests on multiple datasets and
multiple models, including a residual network, to avoid getting results that
may be specific to a particular test. 
The main limitation is that we do not propose a solution for large batch training.
We do offer analytical insights into the relationship between large batch and robust training, but we do not fully resolve the problem of large batch training. 
There have been several approaches to increasing batch size proposed so far~\cite{goyal2017accurate,smith2017don,you2017scaling}, but they
only work for particular cases and require extensive hyper-parameter tuning.
We are performing an in-depth follow up study to use the results of this paper
to better guide large batch size training.

\textbf{Related Work.}
Deep neural networks have achieved good performance for a wide range of applications.
The diversity of the different problems that a DNN can be used for, has been related to
their efficiency in function approximation~\cite{montufar2014number,delalleau2011shallow,le2010deep,anthony2009neural}.
However the work of ~\cite{zhang2016understanding} showed that not only the network can perform well
on a real dataset, but it can also memorize randomly labeled data very well.
Moreover, the performance of the network is highly dependent on
the hyper-parameters used for training.
In particular, recent studies have shown that Neural Networks can easily be fooled by
imperceptible perturbations to input data~\cite{fgsm}.
Moreover, multiple studies have found that large batch size training
suffers from poor generalization capability~\cite{goyal2017accurate,you2017scaling}.

Here we focus on the latter two aspects of training neural networks.
\cite{flat} presented results showing that large batches converge to a ``sharper minima''.
It was argued that even if the sharp minima has the same training loss as the flat one,
small discrepancies between the test data and the training data can easily lead to poor generalization performance~\cite{flat,dinh2017sharp}.
The fact that ``flat minimas'' generalize well goes back to the earlier work of~\cite{hochreiter1997flat}.
The authors relate flat minima to the theory of minimum description length~\cite{rissanen1978modeling}, and proposed
an optimization method to actually favor flat minimas.
There have been several similar attempts to change the optimization algorithm to find ``better''
regions~\cite{desjardins2015natural,chaudhari2016entropy}.
For instance, \cite{chaudhari2016entropy} proposed entropy-SGD,
which uses Langevin dynamics to augment the loss functional to favor flat regions of the ``energy landscape''.
The notion of flat/sharpness does not have a precise definition. A detailed comparison
of different metrics is discussed in~\cite{dinh2017sharp}, where the authors
show that sharp minimas can also generalize well.
The authors also argued that the sharpness can be arbitrarily changed by reparametrization of the weights.
However, this won't happen
when considering the same model and just changing the training hyper-parameters which is the case here.
In~\cite{smith2017don,smithbayesian}, the authors
proposed that the training can be viewed as a stochastic differential equation, and argued that the optimum
batch size is proportional to the training size and the learning rate. 

As our results show, there is an interleaved connection by studying when NNs do not work well.
\cite{szegedy2013intriguing,fgsm} found that they can easily fool a NN with very good generalization by
slightly perturbing the inputs. The perturbation magnitude is most of the time imperceptible to human
eye, but can completely change the networks prediction.
They introduced an effective
adversarial attack algorithm known as Fast Gradient Sign Method (FGSM).
They related the vulnerability of the Neural Network to linear classifiers and showed that
RBF models, despite achieving much smaller generalization performance, are considerably more robust to FGSM attacks.
The FGSM method was then extended in~\cite{fgsm10} to an iterative FGSM, which
performs multiple gradient ascend steps to compute the adversarial
perturbation. Adversarial attack based on iterative FGSM was found to be
stronger than the original one step FGSM.  Various defenses have been proposed
to resist adversarial
attacks~\cite{metzen2017detecting,gong2017adversarial,grosse2017statistical,bhagoji2017dimensionality,feinman2017detecting}.
We will later show that there is an interleaved connection between robustness of the model and
the large batch size problem.

The structure of this paper is as follows: We present the results by first analyzing how the spectrum changes during training, and test the generalization performances of
the model for different batch sizes in ~\secref{sec:large_batch}. In~\secref{sec:large_batch_adv},
we discuss details of how adversarial attack/training is performed. In particular,
we provide theoretical proof that finding adversarial perturbation is a saddle-free problem under certain conditions, and test the robustness of the model for different batch sizes.
Also, we present results showing how robust training affects the spectrum with empirical studies.
Finally, in~\secref{sec:conclusion} we provide concluding remarks.

\section{Large Batch, Generalization Gap and Hessian Spectrum} \label{sec:large_batch}
\textbf{Setup:} The architecture for the networks used is
reported in~\tref{t:model_def}. In the text, we refer to each architecture by the abbreviation
used in this table.
Unless otherwise specified, each of the batch sizes are trained until a training loss of 0.001 or better is achieved.
Different batches are trained under the same conditions, and
no weight decay or dropout is used.

We first focus on large batch size training versus small batch and report 
the results for large batch training for C1 network on CIFAR-10 dataset, and M1 network on MNIST
are shown in~\tref{t:qalex_train_batch}, and \tref{t:lenet_train_batch}, respectively.
As one can see,
after a certain point increasing batch size
results in performance degradation on the test dataset. This is in line with results in the literature~\cite{flat,goyal2017accurate}.

As discussed before, one popular argument about large batch size's poor
generalization accuracy has been that large batches tend to get attracted to
``sharp'' minimas of the training loss. In~\cite{flat} an approximate metric was
used to measure curvature of the loss function for a given model parameter.
Here, we directly compute the Hessian spectrum,
 using power iteration by back-propagating the matvec of the Hessian~\cite{hessianfree}.
Unless otherwise noted, we continue the power iterations
until a relative error of $1\nexp{4}$ reached for each eigenvalue.
The Hessian spectrum for different batches is
shown in~\fref{f:top20eigs}. Moreover, the value of the dominant eigenvalue,
denoted by $\lambda_1^\theta$, is reported in~\tref{t:qalex_train_batch}, and~\tref{t:resnet_train_batch}, respectively
(Additional result for MNIST tested using LeNet-5 is given in appendix. Please see~\tref{t:lenet_train_batch}).
From~\fref{f:top20eigs}, we can clearly see that for all the experiments, large batches have a noticeably larger Hessian spectrum
both in the dominant eigenvalue as well as the rest of the 19 eigenvalues.
However, note that curvature is a very local measure.
It would be more informative to study how the loss functional behaves in a neighborhood around the point that the model has converged.
To visually demonstrate this, we have plotted
how the total loss changes when the model parameters are perturbed along the dominant eigenvector as
shown in~\fref{f:landscape_largebatch_qalex}, and~\fref{f:landscape_largebatch_lenet} for C1 and M1 models, respectively.
We can clearly see that the large batch size models have been 
attracted to areas with higher curvature for both the test and training losses.

% ---------------------------------------------%
% ---------------------------------------------%
\begin{figure*}[tbp]
\begin{center}
  \includegraphics[width=.245\textwidth]{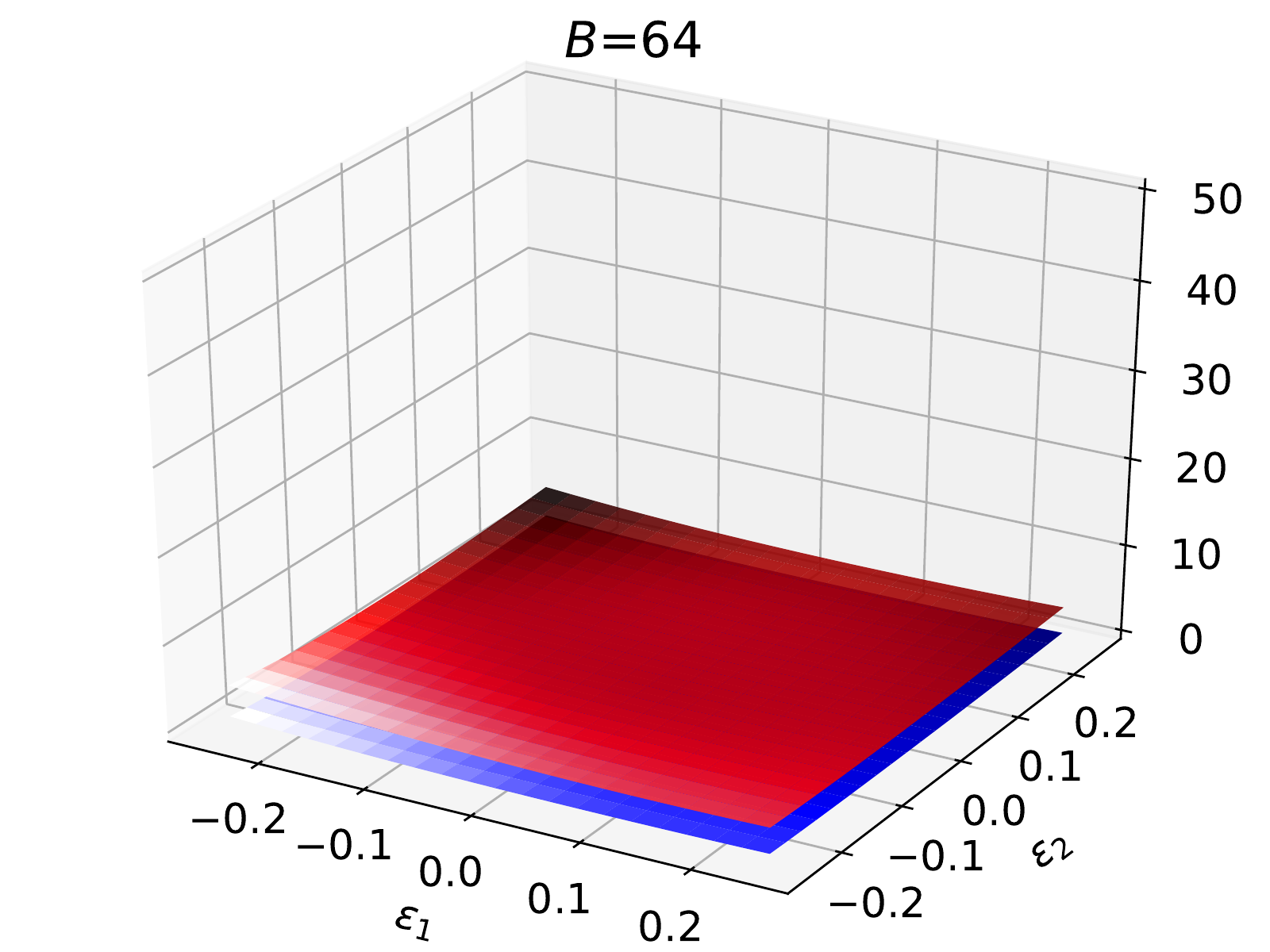}
  \includegraphics[width=.245\textwidth]{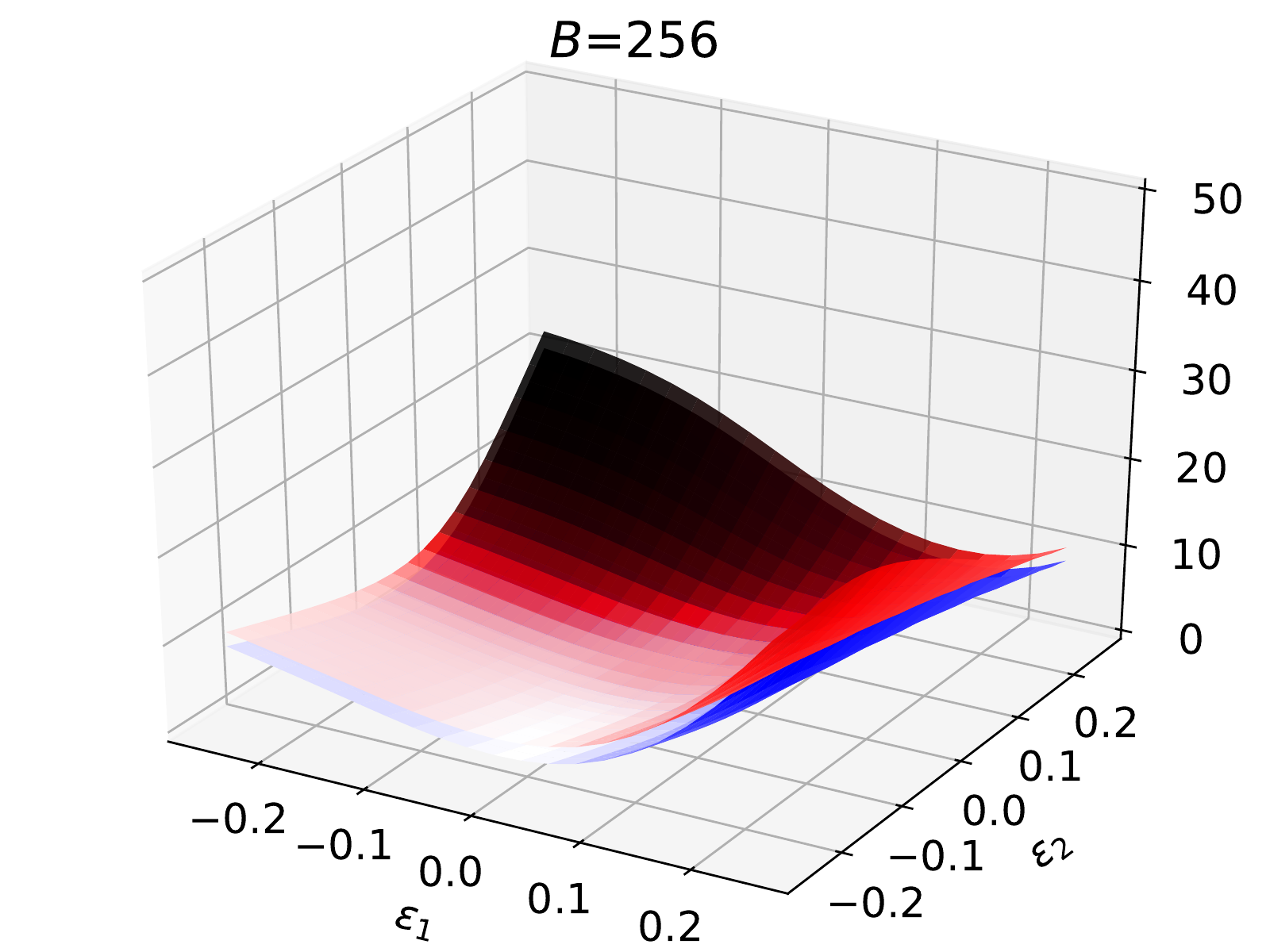}
  \includegraphics[width=.245\textwidth]{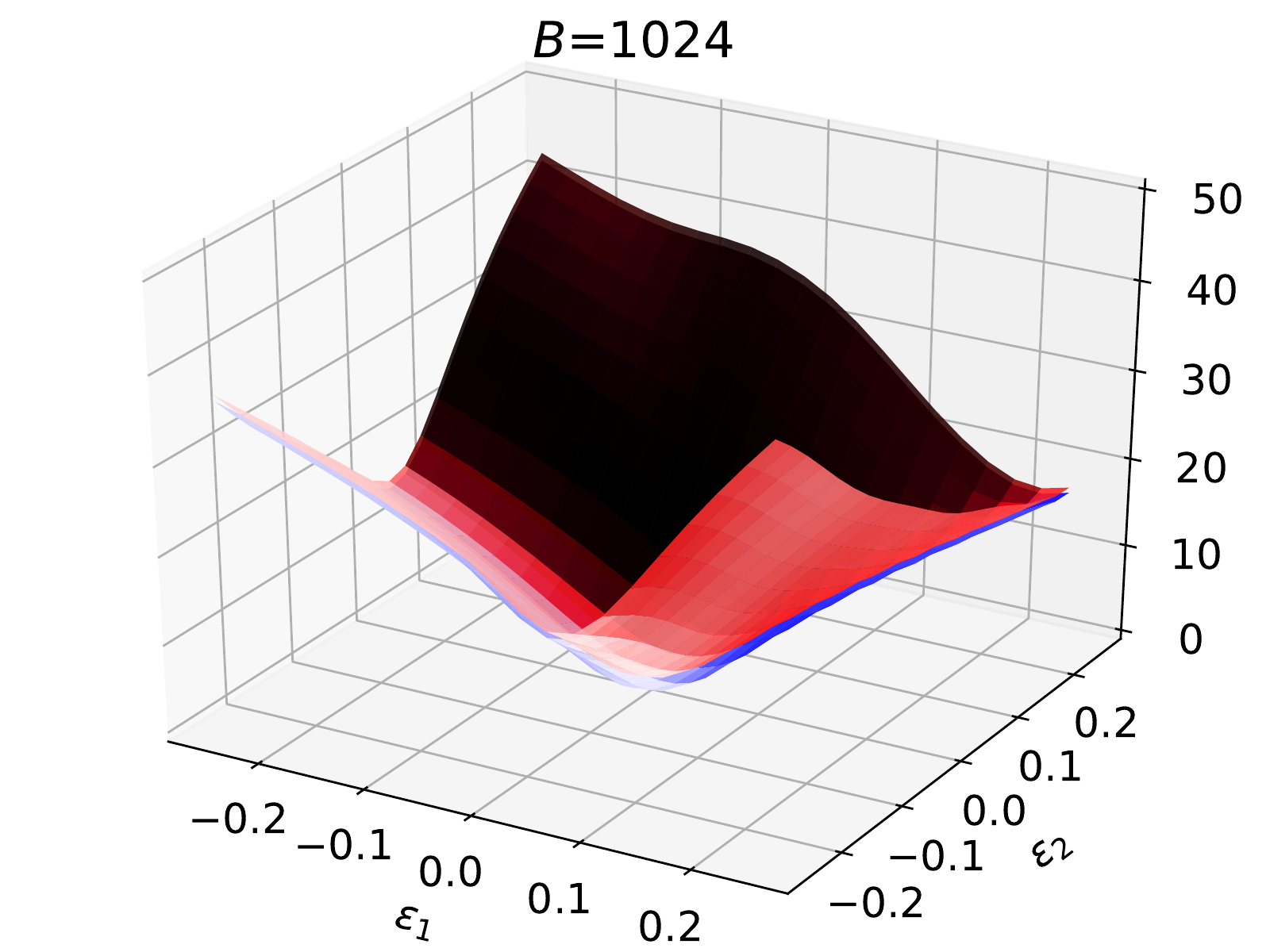}
  \includegraphics[width=.245\textwidth]{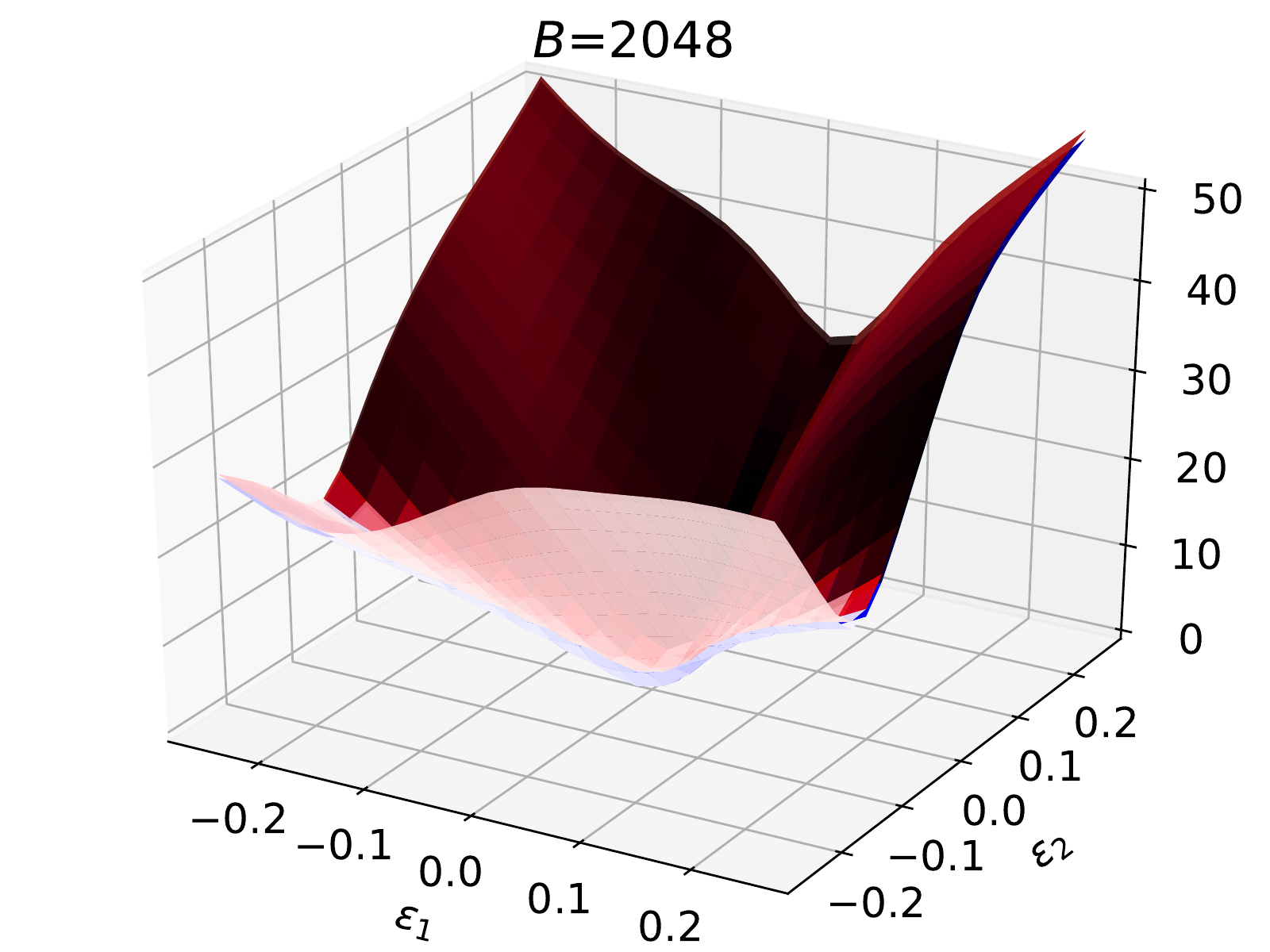}
\end{center}
\caption{
  The landscape of the loss is shown when the C1 model parameters are changed along the first two dominant eigenvectors of the Hessian with the perturbation magnitude $\epsilon_1$ and $\epsilon_2$.
}
\label{f:surface_largebatch_qalex_app}
\end{figure*}

This is reflected in the visual figures.
We have also added a 3D plot, where we perturb the parameters of C1 model along both the first and second eigenvectors as shown in~\fref{f:surface_largebatch_qalex_app}.
The visual results are in line
with the numbers shown for the Hessian spectrum (see $\lambda_1^\theta$) in~\tref{t:qalex_train_batch}, and~\tref{t:lenet_train_batch}.
For instance, note the value of $\lambda_1^\theta$ for the training and test loss for $B=256,\ 2048$ in Table~\ref{t:qalex_train_batch} and compare the corresponding results in~\fref{f:surface_largebatch_qalex_app}.

A recent argument has been that saddle-points in high dimension plague optimization for neural networks~\cite{dauphin2014identifying,ge2015escaping}.
We have computed the dominant eigenvalue of the Hessian along with the total gradient during training and report it in~\fref{fig:qalex_h_logger}.
As we can see, large batch size progressively gets attracted to areas with larger spectrum, but it clearly does not get stuck in saddle points~\cite{lee2017first}.

% -----------------------------------------
\begin{table*}[!htbp]
\centering
\caption{Result on CIFAR-10 dataset using C1, C2 network. We show the Hessian spectrum
of different batch training models, and the corresponding performances on
adversarial dataset generated by training/testing dataset (testing result is given in parenthese).
}
\label{t:qalex_train_batch}
\small
\begin{tabular}{cc|c|c|c|c|c|c}
      & Batch     &   Acc.        & $\lambda_1^\theta$      & $\lambda_1^\x$ & $\|\nabla_\x \J\|$    &Acc $\eps = 0.02$ 	&Acc $\eps = 0.01$ 	        \\%&Acc $\eps = 0.005$ \\
\toprule                                                                          
\Gc	  &  16  &   100 \  (77.68)   &    0.64 \ (32.78)        &   2.69   \ (200.7) & 0.05  \ (20.41)     &    48.07 \ (30.38)    &    72.67 \ (42.70)    \\%&   88.97 \ (52.67)\\
\Ga	  &  32  &   100 \  (76.77)   &    0.97 \ (45.28)        &   3.43   \ (234.5) & 0.05  \ (23.55)     &    49.04 \ (31.23)    &    72.63 \ (43.30)    \\%&   88.67 \ (52.91)\\
\Gc	  &  64  &   100 \  (77.32)   &    0.77 \ (48.06)        &   3.14   \ (195.0) & 0.04  \ (21.47)     &    50.40 \ (32.59)    &    73.85 \ (44.76)    \\%&   89.46 \ (54.14)\\
\Ga	  & 128  &   100 \  (78.84)   &    1.33 \ (137.5)        &   1.41   \ (128.1) & 0.02  \ (13.98)     &    33.15 \ (25.2 )    &    57.69 \ (39.09)    \\%&   80.62 \ (50.63)\\
\Gc	  & 256  &   100 \  (78.54)   &    3.34 \ (338.3)        &   1.51   \ (132.4) & 0.02  \ (14.08)     &    25.33 \ (19.99)    &    50.10 \ (34.94)    \\%&   74.89 \ (47.38)\\
\Ga	  & 512  &   100 \  (79.25)   &   16.88 \ (885.6)        &   1.97   \ (100.0) & 0.04  \ (10.42)     &    14.17 \ (12.94)    &    28.54 \ (25.08)    \\%&   60.63 \ (43.41)\\
\Gc	  &1024  &   100 \  (78.50)   &   51.67 \ (2372 )        &   3.11   \ (146.9) & 0.05  \ (13.33)     &     8.80 \ (8.40 )    &    23.99 \ (21.57)    \\%&   43.13 \ (35.80)\\
\Ga\mrowrot{-8}{C1 Cifar-10}
   	  &2048  &   100 \  (77.31)   &  80.18  \ (3769 )        &   5.18   \ (240.2) & 0.06  \ (18.08)     &     4.14 \ (3.77 )    &    17.42 \ (16.31)    \\%&   32.50 \ (29.40)\\
\midrule
\Gc	  & 256  &   100 \ (79.20)    &  0.62  \ (28 )           &   12.10  \ (704.0) & 0.10  \ (41.95)     &  0.57  \ (0.38)       & 0.73  \ (0.47)        \\%& 0.88  \ (0.56)     \\
\Ga	  & 512  &   100 \ (80.44)    &  0.75  \ (57 )           &    4.82  \ (425.2) & 0.03  \ (26.14)     &  0.34  \ (0.25)       & 0.54  \ (0.38)        \\%& 0.76  \ (0.50)     \\
\Gc	  &1024  &   100 \ (79.61)    &  2.36  \ (142)           &   0.523  \ (229.9) & 0.04  \ (17.16)     &  0.27  \ (0.22)       & 0.46  \ (0.35)        \\%& 0.69  \ (0.46)     \\
\Ga\mrowrot{-4}{C2 Cifar-10}
   	  &2048  &   100 \ (78.99)    &  4.30  \ (307)           &   0.145  \ (260.0) & 0.50  \ (17.94)     &  0.18  \ (0.16)       & 0.33  \ (0.28)        \\%& 0.55  \ (0.41)     \\
\bottomrule
\end{tabular}
\end{table*}
% -----------------------------------------

\section{Large Batch, Adversarial Attack and Robust training}
\label{sec:large_batch_adv}

We first give a brief overview of adversarial attack and robust training and then
present results connecting these with large batch size training.

\subsection{Robust Optimization and Adversarial Attack}\label{sec:advattack}

Here we focus on white-box adversarial attack, and in particular the optimization-based approach both for the attack and defense.
Suppose $\MS(\theta)$ is a learning model (the neural network architecture), and $(\x,y)$ are the input data and
the corresponding labels. The loss functional of the
network with parameter $\theta$ on $(\x,y)$ is denoted by $\J(\theta,\x,y)$.
For adversarial attack, we seek 
a perturbation $\Delta \x$ (with a bounded $L_\infty$ or $L_2$ norm) such that it maximizes $\J(\theta,\x,y)$: 

\begin{equation}
  \max_{\Delta \x\in \mathcal{U}} \J(\theta, \x+\Delta \x, y),
  \label{e:max}
\end{equation}

\noindent where $\mathcal{U}$ is an admissibility set for acceptable perturbation (typically restricting the magnitude of the perturbation).
A typical choice for this set is $\mathcal{U} = \B(\x, \eps)$, a ball of radius $\epsilon$ centered at $\x$.
A popular method for approximately computing $\Delta \x$, is Fast Gradient Sign Method~\cite{fgsm},
where the gradient of the loss functional is computed w.r.t. inputs, and the perturbation is set to:

\begin{equation}
  \Delta \x = \eps \sign(\frac{\partial J(\x,\theta)}{\partial \x}).
\end{equation}

This is not the only attack method possible. Other approaches include an iterative FGSM method (FGSM-10)\cite{fgsm10}
or using other norms such as $L_2$ norm instead of $L_\infty$ (We denote the
$L_2$ method by $L_2Grad$ in our results).  Here we also use a second-order
attack, where we use the Hessian w.r.t.
input to precondition the gradient direction with second order information; please see \tref{t:attack_def} in Appendix for details. 

One method to defend against such adversarial attacks, is to perform robust training~\cite{szegedy2013intriguing, madry2018towards}:
\begin{equation}
  \label{e:minmax}
  \min_{\theta}\max_{\Delta \x\in \mathcal{U}} \J(\theta, \x+\Delta \x, y).
\end{equation}

Solving this min-max optimization problem at each iteration requires first finding the worst
adversarial perturbation that maximizes the loss, and then updating the model
parameters $\theta$ for those cases.
Since adversarial examples have to be generated at every iteration,
it would not be feasible to find the exact perturbation that maximizes the
objective function. Instead, a popular method is to perform a single or
multiple gradient ascents to approximately compute $\Delta \x$.
After computing $\Delta\x$ at each iteration, a typical optimization step (variant of SGD) is performed to update $\theta$.

\noindent Next we show that solving the maximization part is actually a saddle-free problem \textit{almost everywhere}.
This property assures us that the Hessian w.r.t input does not have negative eigenvalue which allows us to use
CG for performing Newton solver for our 2nd order adversarial perturbation tests in~\secref{sec:result_adv_training}.
\footnote{This results might also be helpful for finding better optimization strategies for GANS.}

\subsection{Adversarial perturbation: A saddle-free problem}
Recall that our loss functional is $\J(\theta; \x,y)$. We make following assumptions for the model to help show our theoretical result,

\begin{assumption}\label{ass:jtwicediff}
We
assume the model's activation functions are strictly ReLu activation, and all
layers are either convolution or fully connected. Here, Batch Normalization layers are accepted.
Note that even though the ReLu activation has discontinuity at origin, i.e. $x=0$, ReLu function is twice differentiable almost everywhere.
\end{assumption}

The following theorem shows that the problem of finding an adversarial perturbation that maximized $\mathcal J$, is a saddle-free optimization
problem, with a Positive-Semi-Definite (PSD) Hessian w.r.t. input almost everywhere.
For details on the proof please see Appendix.~\ref{sec:proof_of_thm}.

\begin{theorem}\label{thm:convixity_of_dnn} 
  With Assumption.~\ref{ass:jtwicediff}, for a DNN, its loss functional $\J(\tha,\x,y)$ is a saddle-free function w.r.t. input $\x$ almost everywhere, i.e.
\begin{equation*}
    \frac{\nabla^2 \J(\tha, \x, y)}{\nabla \x^2} \succeq 0.
\end{equation*}
\end{theorem}

\noindent From the proof of Theorem~\ref{thm:convixity_of_dnn}, we could immediately get the following proposition of DNNs:

\begin{proposition}\label{prop:1}
Based on Theorem~\ref{thm:convixity_of_dnn} with
Assumption~\ref{ass:jtwicediff}, if the input $\x\in\R^d$ and the number of the
output class is $c$, i.e. $y\in\{1,2,3\ldots,c\}$, then the Hessian of DNNs
w.r.t. to $\x$ is almost a rank $c$ matrix almost everywhere; see
Appendix~\ref{sec:proof_of_thm} for details.
\end{proposition}

\begin{table*}[!htbp]
\centering
\caption{
Result on CIFAR-100 dataset using CR network. We show the Hessian spectrum
of different batch training models, and the corresponding performances on
adversarial dataset generated by training/testing dataset (testing result is given in parenthese).
}
\label{t:resnet_train_batch}
\begin{tabular}{c|c|c|c|c}
  Batch     &   Acc.                & $\lambda_1^\theta$       & Acc $\eps = 0.02$  &Acc $\eps = 0.01$  \\
\toprule                                                                          
\Gd     64  & 99.98   \ (70.81)   & 0.022    \ (10.43)  & 61.54    \ (34.48)     & 78.57    \ (39.94)   \\
\Ga    128  & 99.97   \ (70.9 )   & 0.055    \ (26.50 ) & 58.15    \ (33.73)    & 77.41    \ (38.77)    \\
\Gd    256  & 99.98   \ (68.6 )   & 1.090    \ (148.29) & 39.96    \ (28.37)    & 66.12    \ (35.02)    \\
\Ga    512  & 99.98   \ (68.6 )   & 1.090    \ (148.29) & 40.48    \ (28.37)    & 66.09    \ (35.02)    \\
\bottomrule
\end{tabular}
\end{table*}

\subsection{Large Batch Training and Robustness}

Here, we test the robustness of the models trained with different
batches to an adversarial attack.  We use Fast Gradient Sign Method for all the
experiments (we did not see any difference with FGSM-10 attack).
The adversarial performance is measured by the fraction of correctly
classified adversarial inputs.  We report the performance for both the training
and test datasets for different values of $\epsilon=0.02,0.01$ 
($\epsilon$ is the metric for the adversarial perturbation magnitude in $L_\infty$ norm).
The
performance results for C1, and C2 models on CIFAR-10, CR model on CIFAR-100,
are reported in the last two columns of Tables~\ref{t:qalex_train_batch},and
~\ref{t:resnet_train_batch} (MNIST results are given in
appendix, Table~\ref{t:lenet_train_batch}).  The interesting observation is that for
all the cases, large batches are considerably more prone to adversarial attacks
as compared to small batches.  This means that not only the model design
affects the robustness of the model, but also the hyper-parameters used during
optimization, and in particular the properties of the point that the model has
converged to.

\begin{table*}[h]
\centering
\caption{Accuracy of different models across different adversarial samples of MNIST, which are obtained by perturbing the original model $\M_{ORI}$}
\label{t:mnist_adv_accuracy}
\small
\begin{tabular}{c|c|c|c|c|c|c|c}
               & $\D_{clean}$   & $\D_{FGSM}$    & $\D_{FGSM10}$  & $\D_{L_2GRAD}$ & $\D_{FHSM}$    & $\D_{L_2HESS}$ & MEAN of Adv              \\ 
\toprule
\Ga $\M_{ORI}$     & 99.32           & 60.37           & 77.27           & 14.32           & 82.04           & 33.21        & 53.44       \\ 
\Gd $\M_{FGSM}$    & 99.49           & 96.18           & 97.44           & 63.46           & 97.56           & 83.33        & 87.59       \\ 
\Ga $\M_{FGSM10}$  & \textbf{99.5}   & 96.52           & \textbf{97.63}  & 66.15           & 97.66           & 84.64        & 88.52       \\ 
\Gd $\M_{L_2GRAD}$ & 98.91           & \textbf{96.88}  & 97.39           & \textbf{86.23}  & \textbf{97.66}  & \textbf{92.56} & \textbf{94.14}       \\ 
\Ga $\M_{FHSM}$    & 99.45           & 94.41           & 96.48           & 52.67           & 96.89           & 77.58        & 83.60       \\ 
\Gd $\M_{L_2HESS}$ & 98.72           & 95.02           & 96.49           & 77.18           & 97.43           & 90.33        & 91.29       \\ 
\bottomrule
\end{tabular}
\end{table*}

% ---------------------------------------------%
% ---------------------------------------------%
\begin{figure*}[tbp]
  \centering
\includegraphics[width=.32\textwidth]{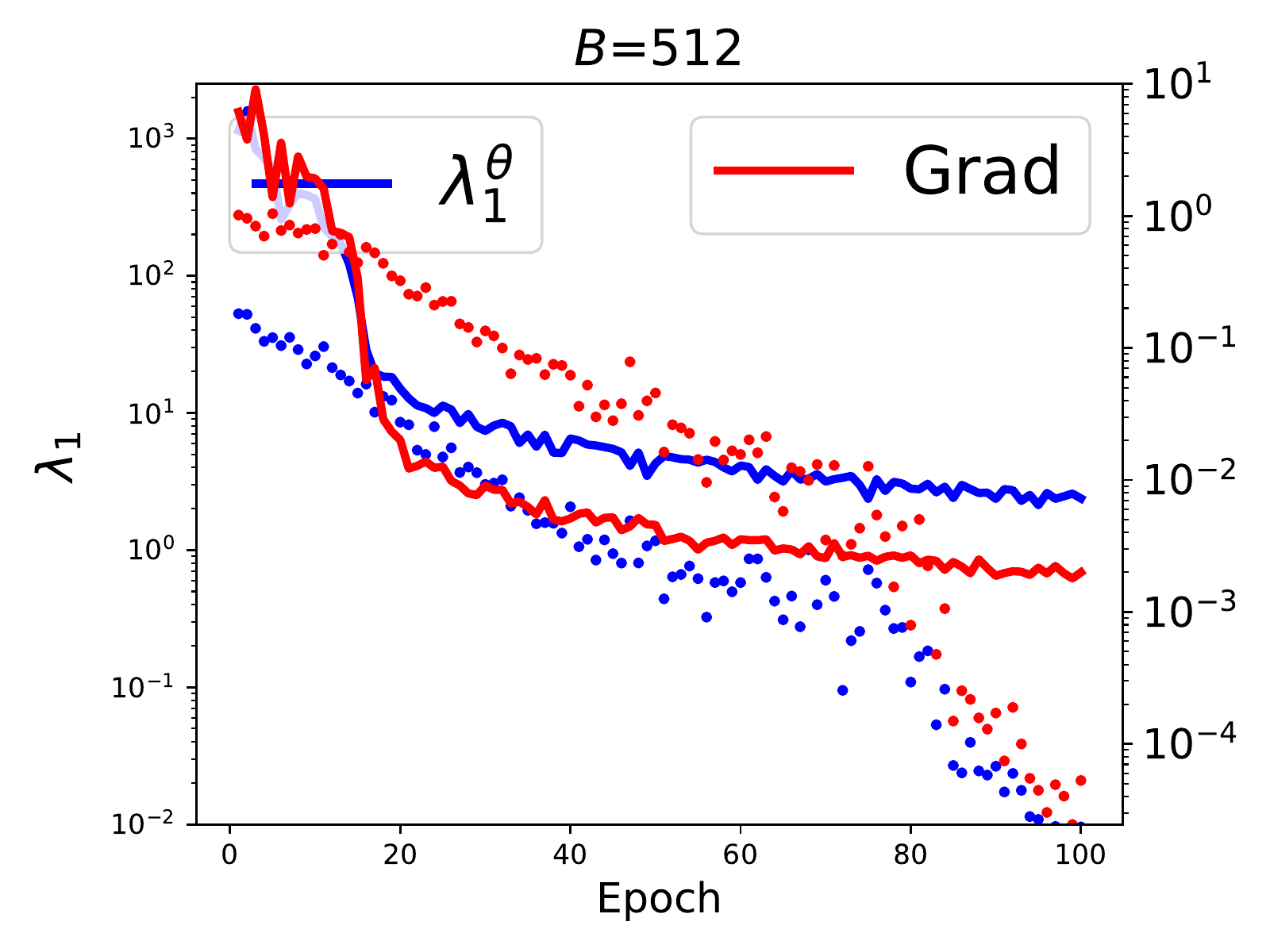}
\includegraphics[width=.32\textwidth]{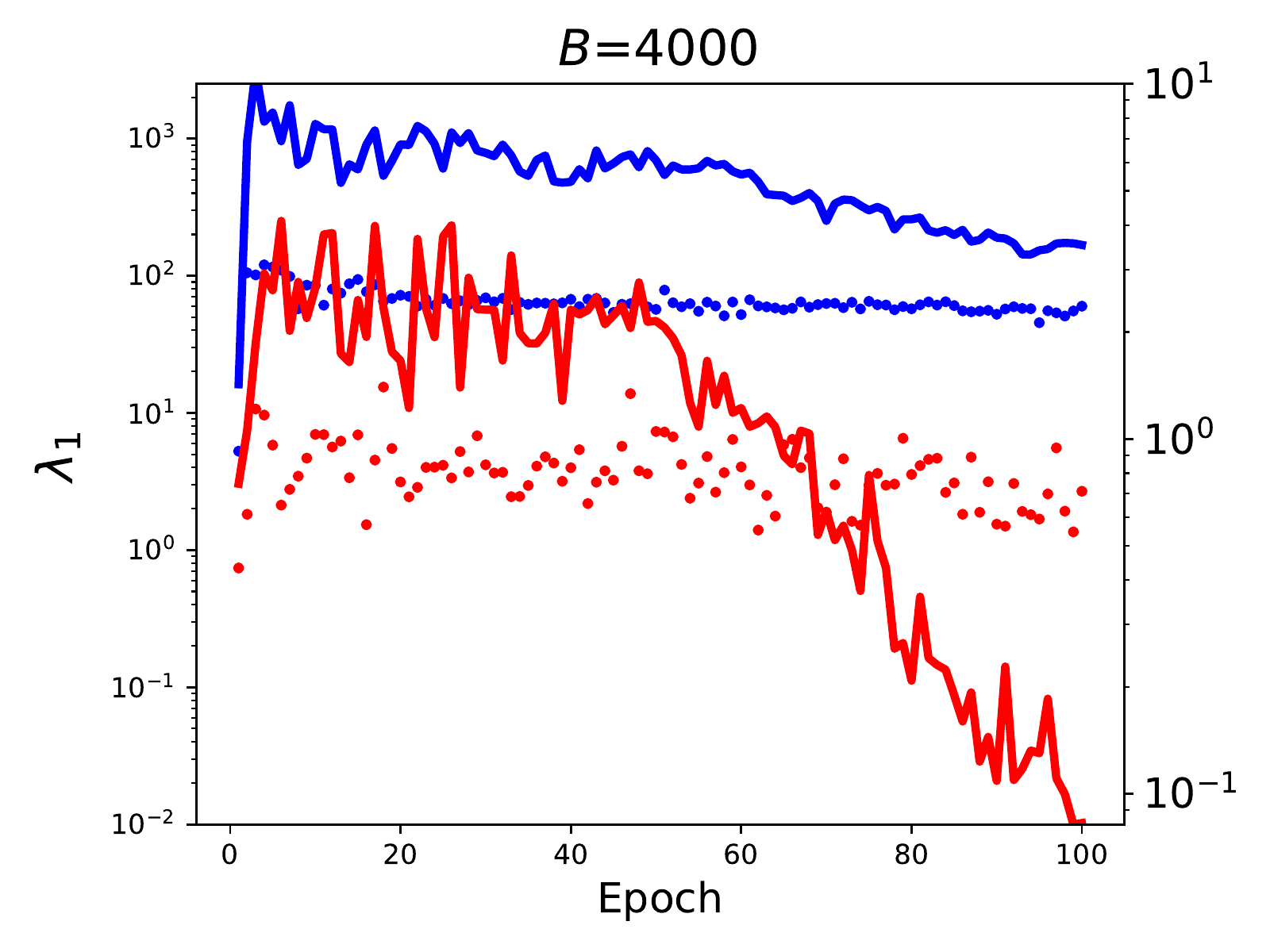}
\includegraphics[width=.32\textwidth]{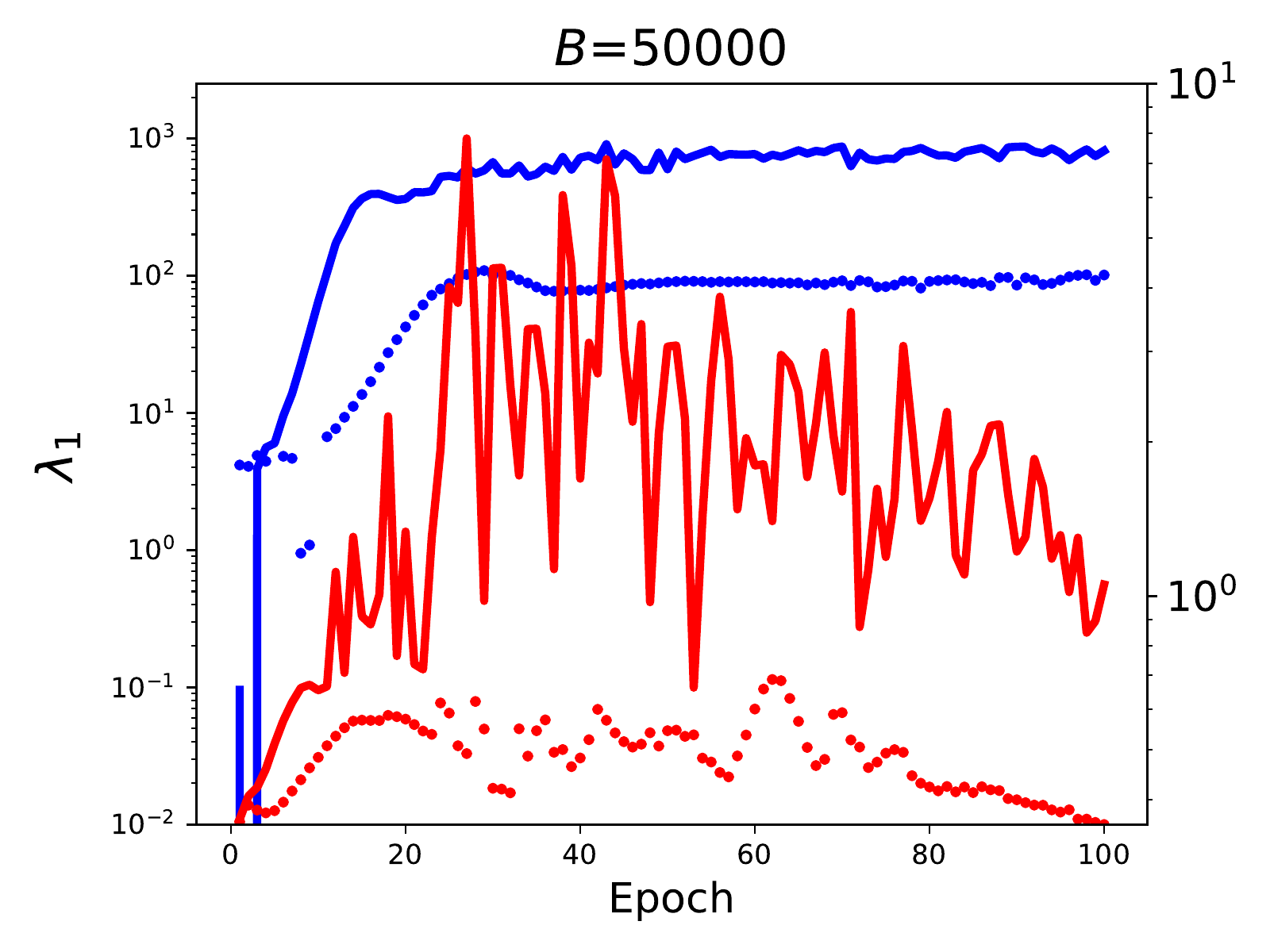}
\caption{
  Changes in the dominant eigenvalue of the Hessian w.r.t weights and the total gradient is shown for
  different epochs during training. Note the increase in $\lambda_1^\theta$ (blue curve) for large batch
  v.s. small batch. In particular, note that the values for total gradient along with the Hessian spectrum show
  that large batch does not get ``stuck'' in
  saddle points, but areas in the optimization landscape that have high curvature. More results are shown in~\fref{fig:qalex_h_logger_appendix}.
  The dotted points show the corresponding results when using robust optimization, which makes 
  the solver stay in areas with smaller spectrum.
}
\label{fig:qalex_h_logger}
\end{figure*}
% ---------------------------------------------%
% ---------------------------------------------%

From this result, there seems to be a strong correlation between the spectrum of the Hessian w.r.t. $\theta$ and
how robust the model is. However, we want to emphasize that in general there is no correlation between
the Hessian w.r.t. weights and the robustness of the model w.r.t. the input. For instance, consider a two variable function
$\mathcal{J}(\tha,\x)$ (we treat $\theta$ and $\x$ as two single variables),
for which the Hessian spectrum of
$\theta$ has no correlation to robustness of $\mathcal{J}$ w.r.t. $\x$.
This can be easily demonstrated for 
a least squares problem, $L=\|\theta\x-\y\|_2^2$. It is
not hard to see the Hessian of $\theta$ and $\x$ are, $\x\x^T$ and
$\theta\theta^T$, respectively. 
Therefore, in general we cannot link the Hessian spectrum w.r.t. weights
to robustness of the network. However, the numerical results for all the neural
networks show that models that have higher Hessian spectrum w.r.t. $\theta$ are
also more prone to adversarial attacks. 
A potential explanation for this would be to look at how the
gradient and Hessian w.r.t. input (i.e.\ $\x$) would change for different batch
sizes. We have computed the dominant eigenvalue of this Hessian using power
iteration for each individual input sample for both training and testing
datasets. Furthermore, we have computed the norm of the gradient w.r.t. $\x$
for these datasets as well.  These two metrics are reported in $\lambda_1^\x$,
and $\|\nabla_x\mathcal{J}\|$; see \tref{t:qalex_train_batch} for details. The results on all of our experiments show that
these two metrics actually do not correlate with the adversarial accuracy. For
instance, consider C1 model with $B=512$. It has both smaller gradient and
smaller Hessian eigenvalue w.r.t. $\x$ as compared to $B=32$, but it performs 
acidly worse under adversarial attack.  One possible reason for this could
be that the decision boundaries for large batches are less stable, such that
with small adversarial perturbation the model gets fooled.

% ---------------------------------------------%
% ---------------------------------------------%
\begin{table*}[tbp]
\centering
\caption{Accuracy of different models across different samples of CIFAR-10, which are obtained by perturbing the original model $\M_{ORI}$}
\label{t:cifar_adv_accuracy}
\small
\begin{tabular}{c|c|c|c|c|c|c|c}
                   & $\D_{clean}$     & $\D_{FGSM}$      & $\D_{FGSM10}$    & $\D_{L_2GRAD}$   & $\D_{FHSM}$    & $\D_{L_2HESS}$ & MEAN of Adv    \\ 
\toprule
\centering
\Ga $\M_{ORI}$     & \textbf{79.46} & 15.25            & 4.46             & 12.37            & 29.64          & 22.93          & 16.93          \\
\Gd $\M_{FGSM}$    & 71.82            & 63.05            & 63.44            & 57.68            & \textbf{66.04} & 62.36          & 62.51          \\
\Ga $\M_{FGSM10}$  & 71.14            & \textbf{63.32} & \textbf{63.88} & \textbf{58.25} & 65.95          & \textbf{62.70} & \textbf{62.82} \\
\Gd $\M_{L_2GRAD}$ & 63.52            & 59.33            & 59.73            & 57.35            & 60.44          & 58.98          & 59.16          \\
\Ga $\M_{FHSM}$    & 74.34            & 47.65            & 43.95            & 38.45            & 62.75          & 55.77          & 49.71          \\
\Gd $\M_{L_2HESS}$ & 71.59            & 50.05            & 46.66            & 42.95            & 62.87          & 58.42          & 52.19          \\ 
\bottomrule
\end{tabular}
\end{table*}
% ---------------------------------------------%
% ---------------------------------------------%

% ---------------------------------------------%
% ---------------------------------------------%
\begin{figure}[h]
  \centering
\includegraphics[width=.32\textwidth]{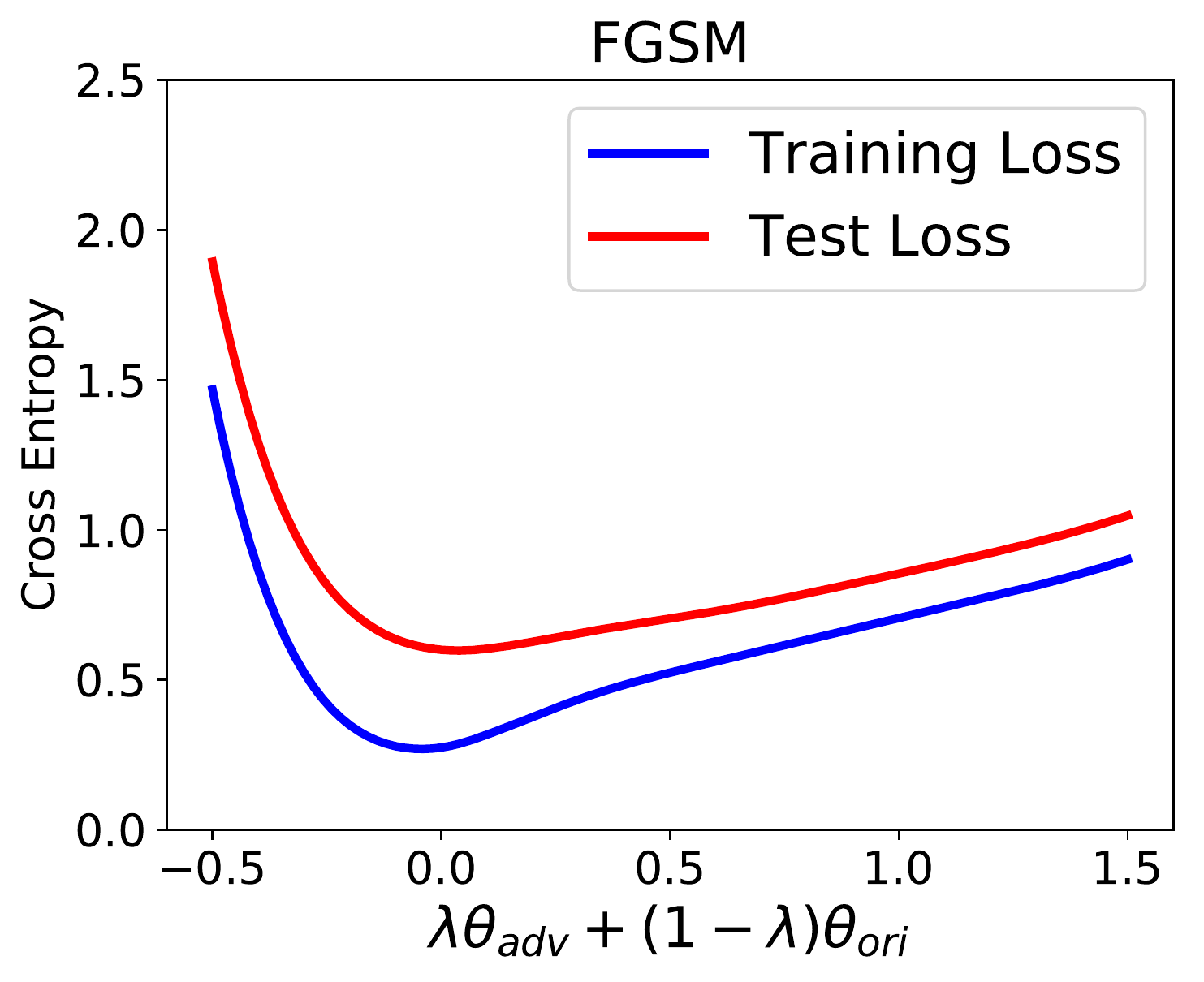}
\includegraphics[width=.32\textwidth]{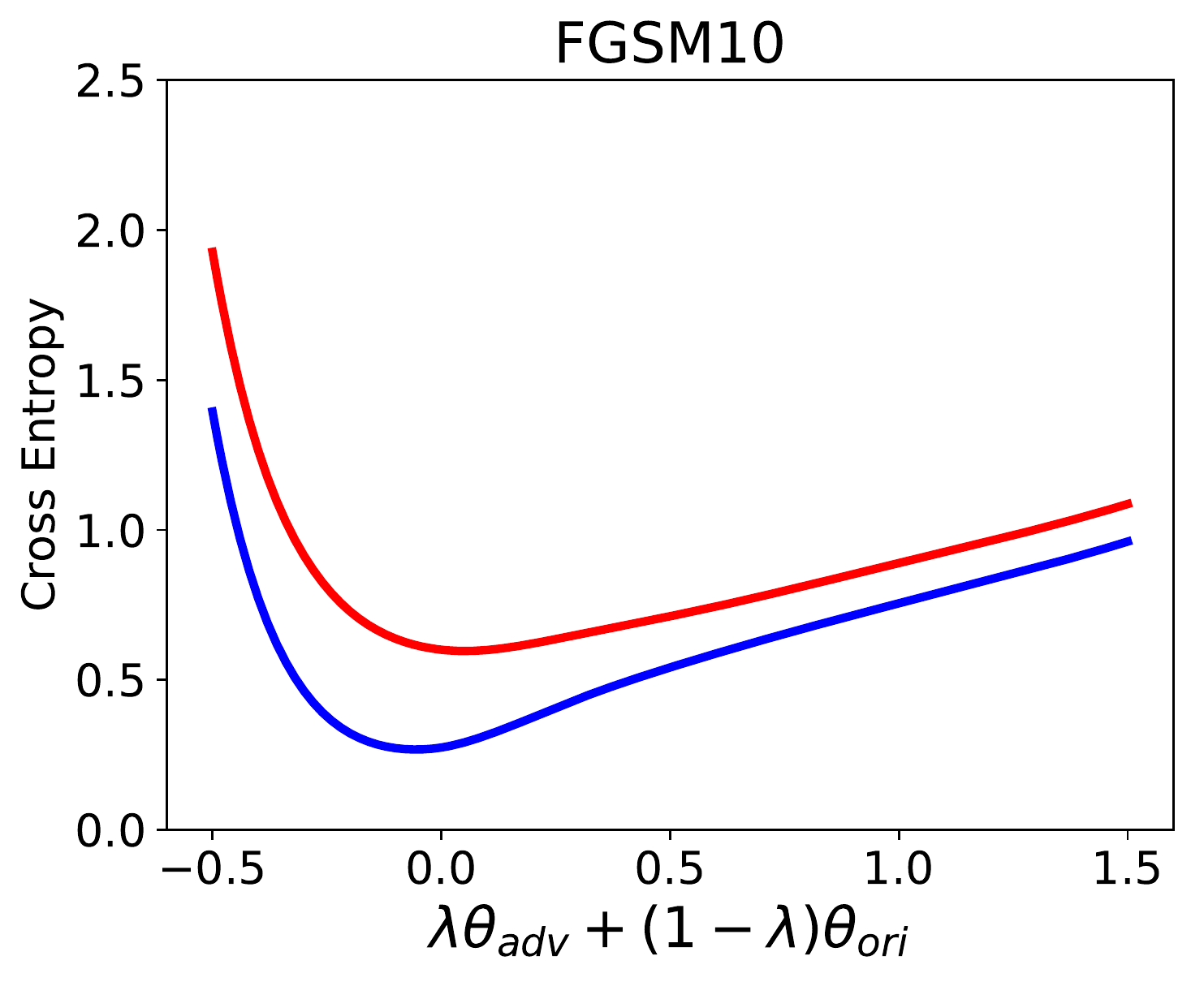}
\includegraphics[width=.32\textwidth]{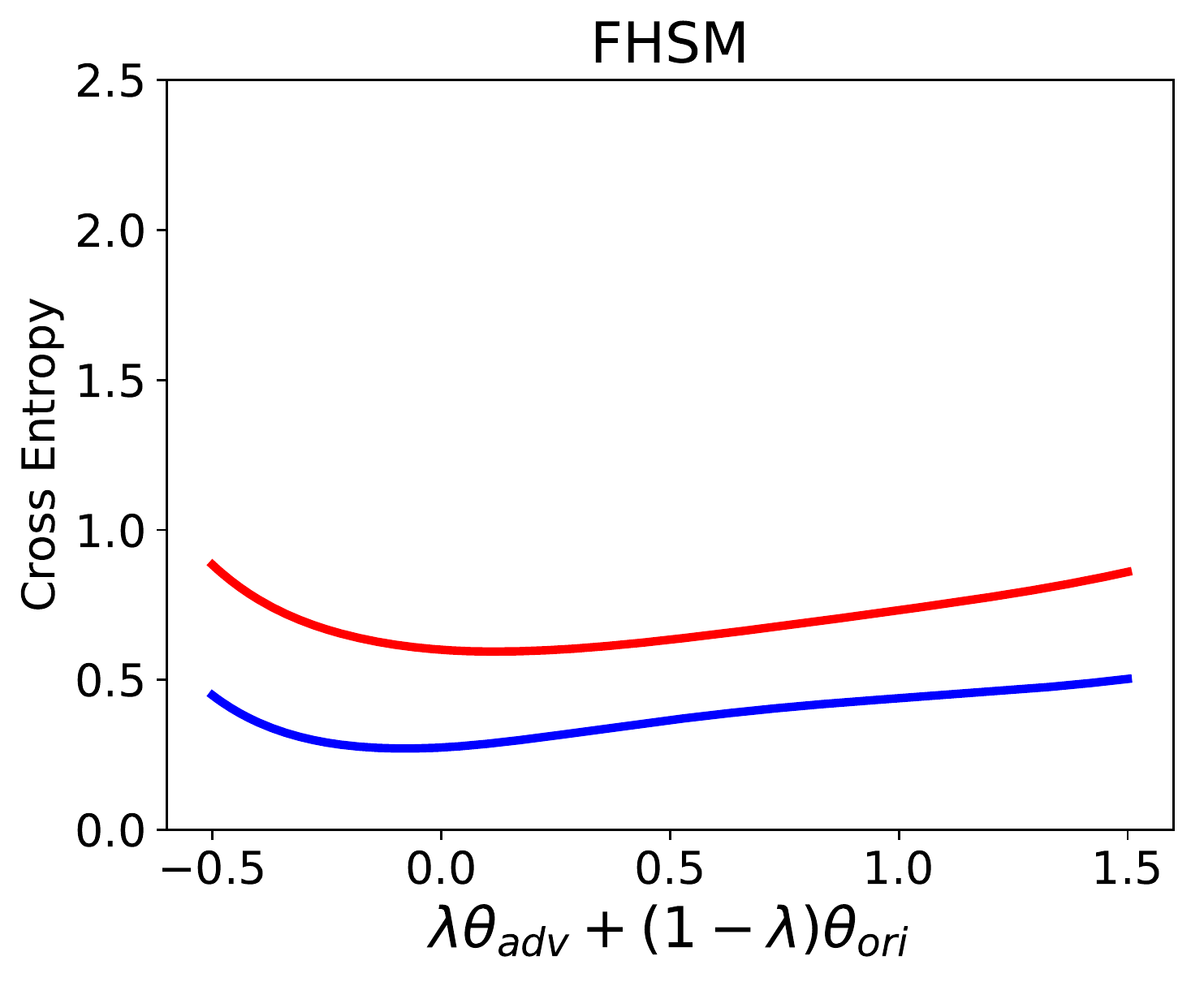}
\caption{1-D Parametric plot for C3 model on CIFAR-10. We interpolate between parameters of $\M_{ORI}$ and $\M_{ADV}$, and
  compute the cross entropy loss on the y-axis. 
}
\label{fig:cifar_adv_interpolation}
\end{figure}
% ---------------------------------------------%
% ---------------------------------------------%

% ---------------------------------------------%
% ---------------------------------------------%
\begin{figure*}[tbp]
  \centering
\includegraphics[width=.33\textwidth]{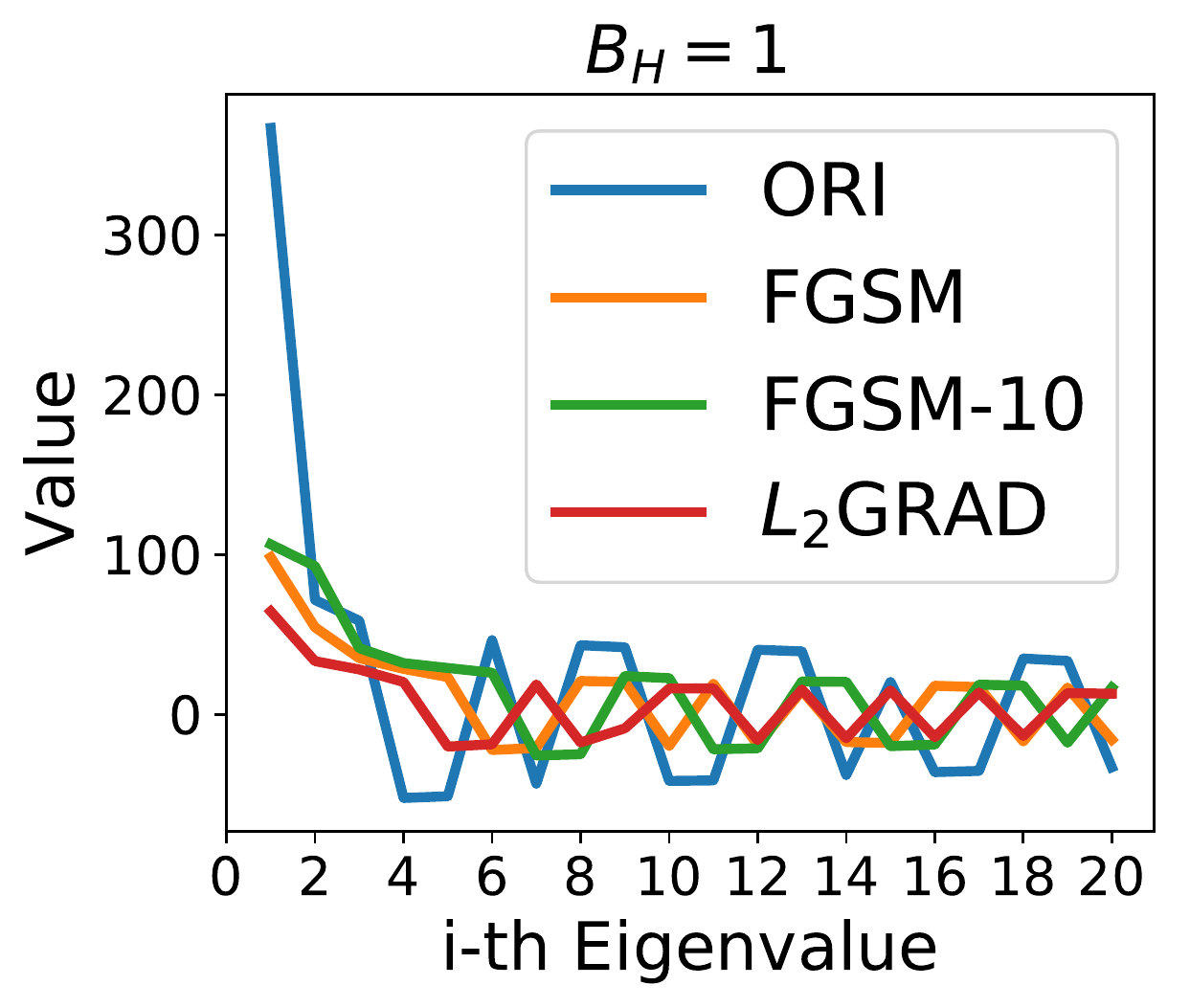}
\includegraphics[width=.32\textwidth]{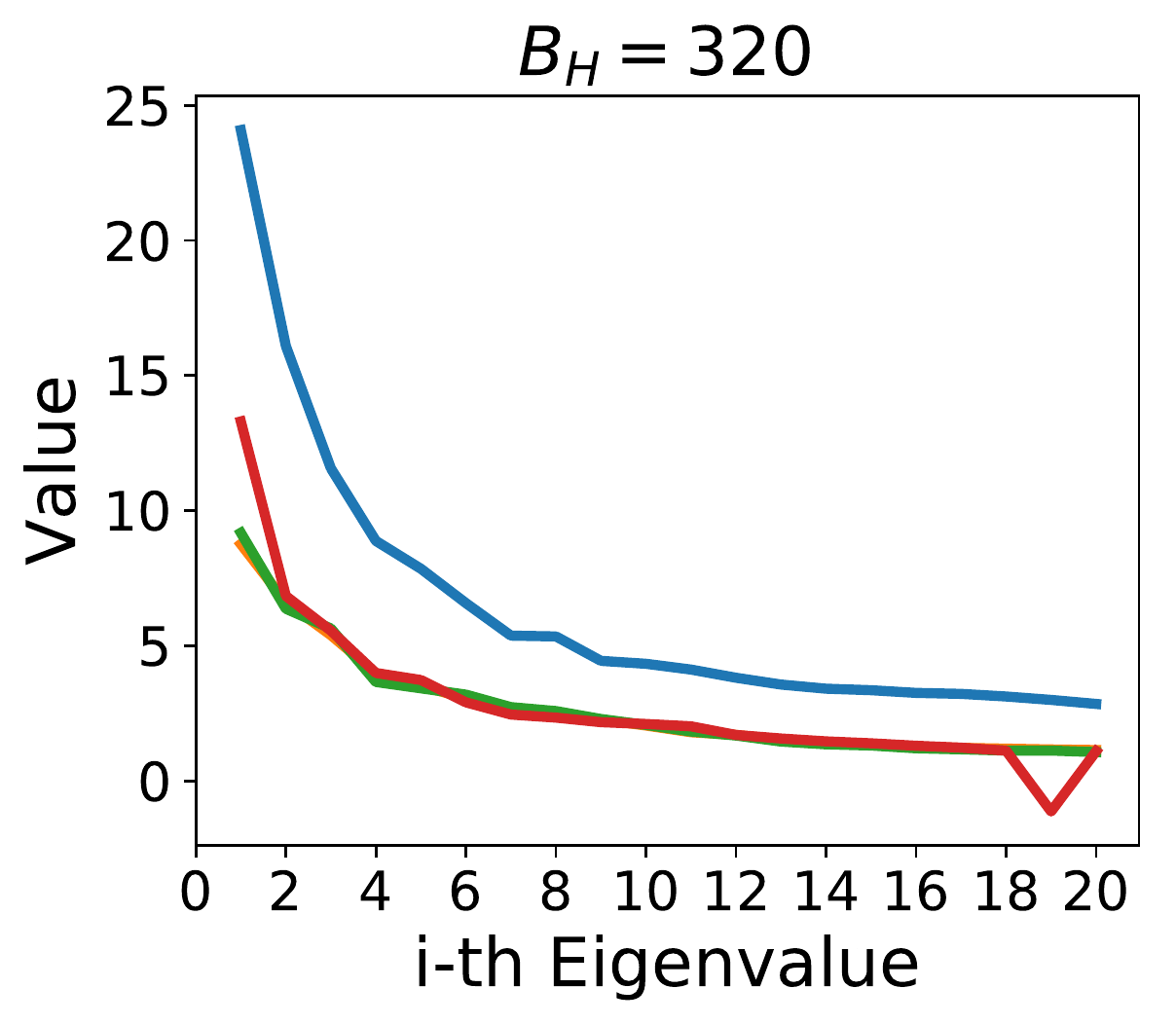}
\includegraphics[width=.32\textwidth]{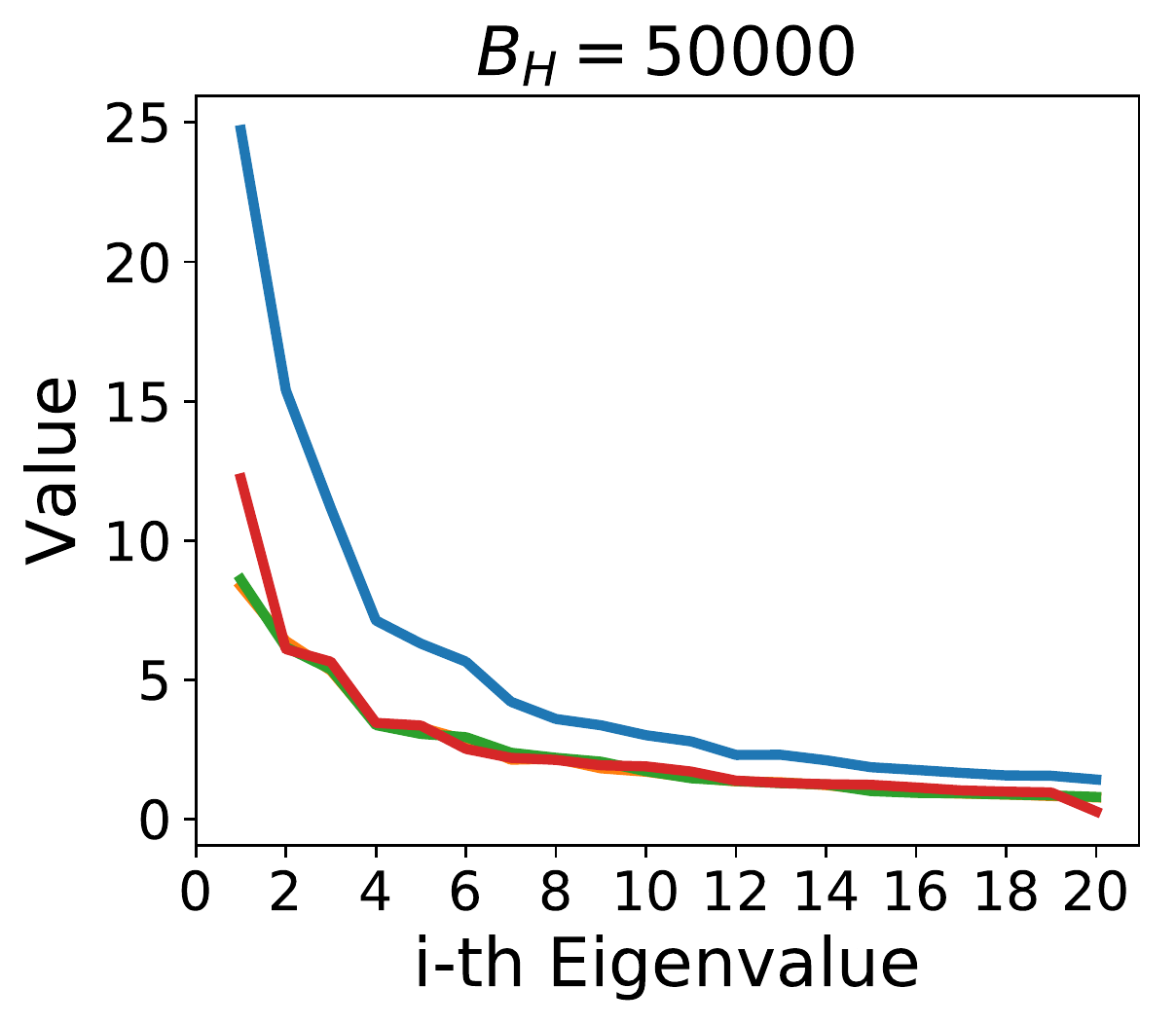}
\caption{
  Spectrum of the sub-sampled Hessian of the loss functional w.r.t. weights.
  The results are computed for different batch sizes, which are randomly chosen, of $B=1,~320,~50000$ of C1. 
}
\label{fig:cifar_adv_topeig}
\end{figure*}
% ---------------------------------------------%
% ---------------------------------------------%
\subsection{Adversarial Training and Hessian Spectrum}\label{sec:result_adv_training}
In this part, we study how the Hessian spectrum and the landscape of the loss functional change after adversarial training is performed. Here, we fix the batch size (and all other
optimization hyper-parameters) and use five different adversarial training methods as described in~\secref{sec:advattack}.

For the sake of clarity let us denote $\D$ to be the test dataset which can be the
original clean test dataset or one created by using an adversarial method.  For
instance, we denote $\D_{FGSM}$ to be the adversarial dataset generated by FGSM,
and $\D_{clean}$ to be the original clean test dataset. 

\textbf{Setup:} For the MNIST experiments, we train a standard LeNet on MNIST
dataset~\cite{mnist} (using M1 network). For the original training, we set the learning rate to
0.01 and momentum to 0.9, and decay the learning rate by half after every 5
epochs, for a total of 100 epochs. Then we perform an additional five epochs of adversarial training
with a learning rate of $0.01$. The perturbation
magnitude, $\epsilon$, is set to $0.1$ for $L_\infty$ attack and $2.8$ for $L_2$ attack.
We also present results for C3 model~\cite{cifarmodel} on CIFAR-10, using the same
hyper-parameters, except that the training is performed for 100 epochs.
Afterwards, adversarial training is performed for a subsequent $10$ epochs
with a learning rate of $0.01$ and momentum of $0.9$ (the learning rate is decayed by half after five epochs).
Furthermore, the adversarial
perturbation magnitude is set to $\epsilon=0.02$ for $L_\infty$ attack and $1.2$ for $L_2$ attack\cite{robust}. 

The results are shown in~\tref{t:mnist_adv_accuracy}, \ref{t:cifar_adv_accuracy}.
We can see that after adversarial training the model becomes more robust to these attacks. 
Note that the accuracy of different
adversarial attacks varies, which is expected since the various strengths of
different attack method. In addition, all adversarial training methods improve
the robustness on adversarial dataset, though they lose some accuracy on
$\D_{clean}$, which is consistent with the observations in~\cite{fgsm}.
As an example, consider the second row of~\tref{t:mnist_adv_accuracy} which shows the results
when FGSM is used for robust training. The
performance of this model when tested against the $L_2GRAD$ attack method is 63.46\% as
opposed to 14.32\% of the original model ($\M_{ORI}$). The rest of the rows show the results
for different algorithms. In \secref{subsec:conjecture_second_order}, we give further discussion about the performances 
of first order and second order attacks.

The main question here is how
the landscape of the loss functional is changed after these robust
optimizations are performed?  We first show a 1-D parametric interpolation
between the original model parameters $\theta$ and that of the robustified models, as
shown in~\fref{fig:cifar_adv_interpolation}~(see \fref{fig:cifar_adv_interpolation_app} for all cases) and~\ref{fig:mnist_adv_interpolation}.
Notice the robust models are at a point
that has smaller curvature as compared to the original model. To exactly quantify this, we
compute the spectrum of the Hessian as shown in~\fref{fig:cifar_adv_topeig}, and~\ref{f:mnist_adv_topeig}.
Besides the full Hessian spectrum, we also report the spectrum of sub-sampled Hessian. The latter
is computed by randomly selecting a subset of the training dataset. We denote the size of this
subset as $B_H$ to avoid confusion with the training batch size. In particular, we
report results for $B_H=1$ and $B_H=320$. 
There are several important observations here. First, notice that the spectrum of the robust models
is noticeably smaller than the original model. This means that the min-max problem of~\eref{e:minmax} favors
areas with lower curvature. Second, note that even though the total Hessian shows that we have converged
to a point with positive curvature (at least based on the top 20 eigenvalues), but that is not necessarily the case
when we look at individual samples (i.e.\ $B_H=1$). For a randomly selected batch of $B_H=1$, we see that we have actually converged to a point that has both positive and negative curvatures, with a non-zero gradient (meaning it is not a saddle point).
To the best of our knowledge this is a new finding, but one that is expected as SGD optimizes the expected loss instead of individual ones.

Now going back to~\fref{fig:qalex_h_logger}, we show how the spectrum changes during training when we use robust optimization.
We can clearly see that with robust optimization the solver is pushed to areas with smaller spectrum.
This is a very interesting finding and shows the possibility of using robust training as a systematic means to bias the
solver to avoid \textit{sharp} minimas. A preliminary result is shown in Table~\ref{t:robust_potential}, where we can see that
robust optimization performs better for large batch size training as opposed to the baseline, or when we use the method
proposed by~\cite{goyal2017accurate}. However, we emphasize that the goal is to perform analysis
to better understand the problems with large batch size training.
More extensive tests are needed before one could claim that robust optimization performs better
than other methods.

\section{Conclusion}\label{sec:conclusion}
In this work, we studied NNs through the lens of the Hessian operator.
In particular, we studied large batch size training and its connection with stability of the model
in the presence of white-box adversarial attacks. 
By computing the Hessian spectrum we provided
several points of evidence that show that
large batch size training tends to get attracted to areas with higher Hessian spectrum.
We reported the eigenvalues of the Hessian w.r.t. whole dataset,
and plotted the landscape of the loss when perturbed along the dominant eigenvector.
Visual results were in line with the numerical values for the spectrum.
Our empirical results show that adversarial attacks/training and large batches are closely related.
We provided several empirical results on multiple datasets that show large batch size training is more prone to adversarial attacks
(more results are provided in the supplementary material). This means that
not only is the model design important, but also that the optimization hyper-parameters can drastically affect a network's robustness. 
Furthermore, we observed that robust training is antithetical to large batch size training, in the sense
that it favors areas with
noticeably smaller Hessian spectrum w.r.t. $\theta$.

The results show that the robustness of the model does not (at least directly) correlate with the Hessian w.r.t. $\x$. We also found
that this Hessian is actually a PSD matrix, meaning that the problem of finding the adversarial perturbation is actually a saddle-free problem \textit{almost everywhere}
for cases that satisfy this criterion~\ref{ass:jtwicediff}.
Furthermore, we showed that even though the model may converge to an area with positive curvature when considering all
of the training dataset (i.e.\ total loss), if we look at individual samples then the Hessian can 
actually have significant negative eigenvalues. From an optimization viewpoint, this is due to the fact that SGD optimizes
the expected loss and not the individual per sample loss.

 \clearpage
\bibliographystyle{plainnat}
\bibliography{ref}
\newpage
\appendix
\section{Appendix}

\subsection{Proof of Theorem 1}\label{sec:proof_of_thm}

In this section, we give the proof of Theorem ~\ref{thm:convixity_of_dnn}. The
first thing we want to point out is that, although we prove the Hessians of these NNs
are positive semi-definite almost everywhere, these NNs are not convex w.r.t. inputs,
i.e., $\x$. The discontinuity of ReLU is the cause. (For instance, consider a
combination of two step functions in 1-D, e.g. $f(x) = 1_{x\ge1}+1_{x\ge2}$ is
not a convex function but has $0$ second derivative almost everywhere.) However, this
has an important implication, that the problem is saddle-free.

Before we go to the proof of Theorem ~\ref{thm:convixity_of_dnn}, let us prove the following lemma for cross-entropy loss with soft-max layer.

\begin{lemma}\label{lemma:1} Let us denote by $\s\in \R^d$ the input of the soft-max function, by $y\in\{1,2,\ldots,d\}$ the correct label of the inputs $\x$, by $g(\s)$ the soft-max function, and by $L(\s, y)$ the cross-entropy loss. Then we have 
\[
  \frac{\partial^2 L(\s,y)}{\partial \s^2} \succeq \0.
\]

\end{lemma}

\begin{proof} Let $s_d = \sum_{j=1}^d e^{\s_j}$, $\p_i=\frac{e^{\s_i}}{s_d}$, and then it follows that 
\[
  L(\s,y) = -\sum_{i=1}^d y_i\log\p_i  .
\]
Further, it is not hard to see that
\begin{align*}
\frac{\partial L(\s,y)}{\partial \s_j} 
&= -\sum_{i=1}^d y_i\frac{\partial \log\p_i}{\partial \s_j} \\
& = -y_j(1-\p_j) - \sum_{i\not=j} y_i\frac{\p_k\p_j}{\p_k} \\
& = \p_j - y_j.
\end{align*}
Then, the second order derivative of $L$ w.r.t. $\s_i\s_j$ is
\[
  \frac{\partial^2 L(\s,y)}{\partial \s_j^2} = \p_j(1-\p_j),~~~~~~~\text{and}~~~~~~~  \frac{\partial^2 L(\s,y)}{\partial \s_j\partial\s_i} = -\p_j\p_i.
\]
Since
\[
  \frac{\partial^2 L(\s,y)}{\partial \s_j^2} + \sum_{i\not=j} \frac{\partial^2 L(\s,y)}{\partial \s_j\partial\s_i} = 0,~~~\text{and}~~~\frac{\partial^2 L(\s,y)}{\partial \s_j^2} \geq 0,
\]
we have
\[
  \frac{\partial^2 L(\s,y)}{\partial \s^2} \succeq \0.
\]
\end{proof}

Now, let us give the proof of Theorem ~\ref{thm:convixity_of_dnn}:

Assume the input of the soft-max layer is $\s$ and the cross-entropy is $L(\s, y)$. Based on Chain Rule, it follows that
\begin{equation*}
\frac{\partial \J(\tha, \x, y)}{\partial \x} = \frac{\partial L}{\partial  \s}\frac{\partial \s}{\partial \x}.
\end{equation*}

From Assumption.~\ref{ass:jtwicediff} we know that all the layers before the soft-max are either linear or ReLU, which indicates
$\dfrac{\partial^2 \s}{\partial \x^2} = \0$ (a tensor) almost everywhere. Therefore, applying chain rule again for the above equation, 
\begin{align*}
    \frac{\partial^2 \J(\tha, \x, y)}{\partial \x^2} 
    &= (\frac{\partial \s}{\partial \x})^T\frac{\partial^2 L}{\partial  \s^2}\frac{\partial \s}{\partial \x} + \frac{\partial L}{\partial  \s}\frac{\partial^2 \s}{\partial \x^2}\\
    &= (\frac{\partial \s}{\partial \x})^T\frac{\partial^2 L}{\partial  \s^2}\frac{\partial \s}{\partial \x}.
\end{align*}
It is easy to see $\dfrac{\partial^2 \J(\tha, \x, y)}{\partial \x^2}\succeq 0$ almost everywhere since $\dfrac{\partial^2 L}{\partial  \s^2}\succeq 0$ from Lemma~\ref{lemma:1}. 

From above we could see that the Hessian of NNs w.r.t. $\x$ is at most a rank $c$ (the number of class) matrix, since the rank of the Hessian matrix
\[
  \frac{\partial^2 \J(\tha, \x, y)}{\partial \x^2} = (\frac{\partial \s}{\partial \x})^T\frac{\partial^2 L}{\partial  \s^2}\frac{\partial \s}{\partial \x}
\]
is dominated by the term $\frac{\partial^2 L}{\partial  \s^2}$, which is at most rank $c$.

\subsection{Attacks Mentioned in Paper}
In this section, we show the details about the attacks used in our paper. Please see \tref{t:attack_def} for details.
\begin{table}[!htbp]
\centering
\caption{The definition of all attacks used in the paper. Here $\g_x \triangleq  \dfrac{\partial \J(\x,\theta)}{\partial \x}$ and $\H_x\triangleq\dfrac{\partial^2 \J(\x,\theta)}{\partial \x^2}$.}
\label{t:attack_def}
\begin{tabular}{c|c}
\toprule
\Ga          & $\Delta \x$         \\                                                                                           
\Gd FGSM      & $\eps\cdot \sign(\g_x)$                                                       \\ 
\Ga FGSM-10    & $\eps\cdot \sign(\g_x)$ (iterate 10 times)                                    \\ 
\Gd $L_2$GRAD & $\eps\cdot \g_x/\|\g_x\|$       \\
\Ga FHSM      & $\eps\cdot \sign(\H_x^{-1}\g_x)$                                                   \\ 
\Gd  $L_2$HESS & $\eps\cdot \H_x^{-1}\g_x/\|\H_x^{-1}\g_x\|$ \\
\bottomrule
\end{tabular}
\end{table}

\subsection{Models Mentioned in Paper}
In this section, we give the details about the NNs used in our paper. For clarification, We omit the ReLu activation here. However, in practice, we implement ReLu regularity. Also, for all convolution layers, we add padding to make sure there is no dimension reduction. We denote Conv(a,a,b) as a convolution layer having b channels with a by a filters, MP(a,a) as a a by a max-pooling layer, FN(a) as a fully-connect layer with a output and SM(a) is the soft-max layer with a output. For our Conv(5,5,b) (Conv(3,3,b))layers, the stride is 2 (1). See \tref{t:model_def} for details of all models used in this paper.

\begin{table}[!htbp]
\centering
\caption{The definition of all models used in the paper.}
\label{t:model_def}
\begin{tabular}{c|c}
\toprule
\Ga    Name             & Structure                           \\
\Gd  C1 (for CIFAR-10)   & \begin{tabular}[c]{@{}c@{}}Conv(5,5,64) -- MP(3,3) -- BN--Conv(5,5,64)\\ --MP(3,3)--BN--FN(384)--FN(192)--SM(10)\end{tabular}                 \\
\Ga  C2 (for CIFAR-10)   & \begin{tabular}[c]{@{}c@{}}Conv(3,3,63)--BN--Conv(3,3,64)--BN--Conv(3,3,128)\\ --BN--Conv(3,3,128)--BN--FC(256)--FC(256)--SM(10)\end{tabular} \\
\Gd  C3 (for CIFAR-10)   & \begin{tabular}[c]{@{}c@{}}Conv(3,3,64)--Conv(3,3,64)--Conv(3,3,128)\\ --Conv(3,3,128)--FC(256)--FC(256)--SM(10)\end{tabular}                 \\
\Ga  M1 (for MNIST)      & Conv(5,5,20)--Conv(5,5,50)--FC(500)--SM(10)                                                                                                   \\
\Gd   CR (for CIFAR-100) & ResNet18 For CIFAR-100 \\
\bottomrule
\end{tabular}
\end{table}

\subsection{Discussion on Second Order Method} \label{subsec:conjecture_second_order}
Although second order adversarial attack looks well for MNIST (see \tref{t:mnist_adv_accuracy}), but for most our
experiments on CIFAR-10 (see \tref{t:cifar_adv_accuracy}), the second order methods are weaker than variations of the gradient
based methods. Also, notice that the robust models trained by second order method are
also more prone to attack on CIFAR-10, particularly $\M_{FHSM}$ and
$\M_{L_2HESS}$. We give two potential explanation here. %we do not what

First note that the Hessian w.r.t. input is a low rank matrix.
In fact, as mentioned above, the rank of the input Hessian for CIFAR-10 is at most ten; see Proposition~\ref{prop:1},
the matrix itself is $3K\times3K$. 
Even though we use inexact Newton method~\cite{eisenstat1996choosing} along
with Conjugate Gradient solver, but this low rank nature creates numerical problems.
Designing preconditioners for second-order attack is part of our future work. The second
point is that, as we saw in the previous section the input Hessian does not directly
correlate with how robust the network is. In fact, the most effective attack method would be 
to perturb the input towards the decision boundary, instead of just maximizing the loss. 

\subsection{More Numerical Result for \secref{sec:large_batch} and \ref{sec:large_batch_adv}}
In this section, we provide more numerical results for \secref{sec:large_batch} and \ref{sec:large_batch_adv}. All conclusions from the numerical results are consistency with those in \secref{sec:large_batch} and \ref{sec:large_batch_adv}.

% ---------------------------------------------%
% ---------------------------------------------%
\begin{table*}[h]
\centering
\caption{Result on MNIST dataset for M1 model (LeNet-5). We shows the Hessian
spectrum of different batch training models, and the corresponding performances
on adversarial dataset generated by training/testing dataset. The testing results are shown in parenthesis. We report the adversarial
accuracy of three different magnitudes of attack. The interesting observation
is that the $\lambda_1^\theta$ is increasing while the adversarial accuracy is
decreasing for fixed $\eps$. Meanwhile, we do not know if there is a
relationship between $\lambda_1^\theta$ and Clean accuracy or not. Also, we cannot see the relation between $\lambda_1^\x$, $\|\nabla_\x\J(\theta, \x,y)\|$ and the adversarial accuracy. }
\label{t:lenet_train_batch}
\small
\begin{tabular}{c|c|c|c|c|c|c}
  Batch     &   Acc           & $\lambda_1^\theta$  & $\lambda_1^\x$    & $\|\partial_\x\J(\theta, \x,y)\|$  & Acc $\eps = 0.2$  & Acc $\eps = 0.1$  \\
\toprule     
\Gd    64   &  100 \ (99.21)  & 0.49  \ (2.96 )    &   0.07   \ (0.41) & 0.007  \ (0.10)                  &  0.53  \ (0.53)  & 0.85  \ (0.85)        \\
\Ga    128  &  100 \ (99.18)  & 1.44  \ (8.10 )    &   0.10   \ (0.51) & 0.009  \ (0.12)                  &  0.50  \ (0.51)  & 0.83  \ (0.83)        \\                                  
\Gd    256  &  100 \ (99.04)  & 2.71  \ (13.54)    &   0.09   \ (0.50) & 0.008  \ (0.12)                  &  0.45  \ (0.46)  & 0.81  \ (0.82)        \\
\Ga    512  &  100 \ (99.04)  & 5.84  \ (26.35)    &   0.12   \ (0.52) & 0.010  \ (0.13)                  &  0.42  \ (0.42)  & 0.79  \ (0.80)        \\
\Gd   1024  &  100 \ (99.05)  & 21.24  \ (36.96)    &   0.25   \ (0.42) & 0.032  \ (0.11)                  &  0.32  \ (0.33)  & 0.73  \ (0.74)       \\
\Ga   2048  &  100 \ (98.99)  & 44.30  \ (49.36)    &   0.36   \ (0.39) & 0.075  \ (0.11)                  &  0.19  \ (0.19)  & 0.72  \ (0.73)       \\
\bottomrule
\end{tabular}
\end{table*}

% ---------------------------------------------%
% ---------------------------------------------%
\begin{figure*}[!htbp]
\begin{center}
  \includegraphics[width=.32\textwidth]{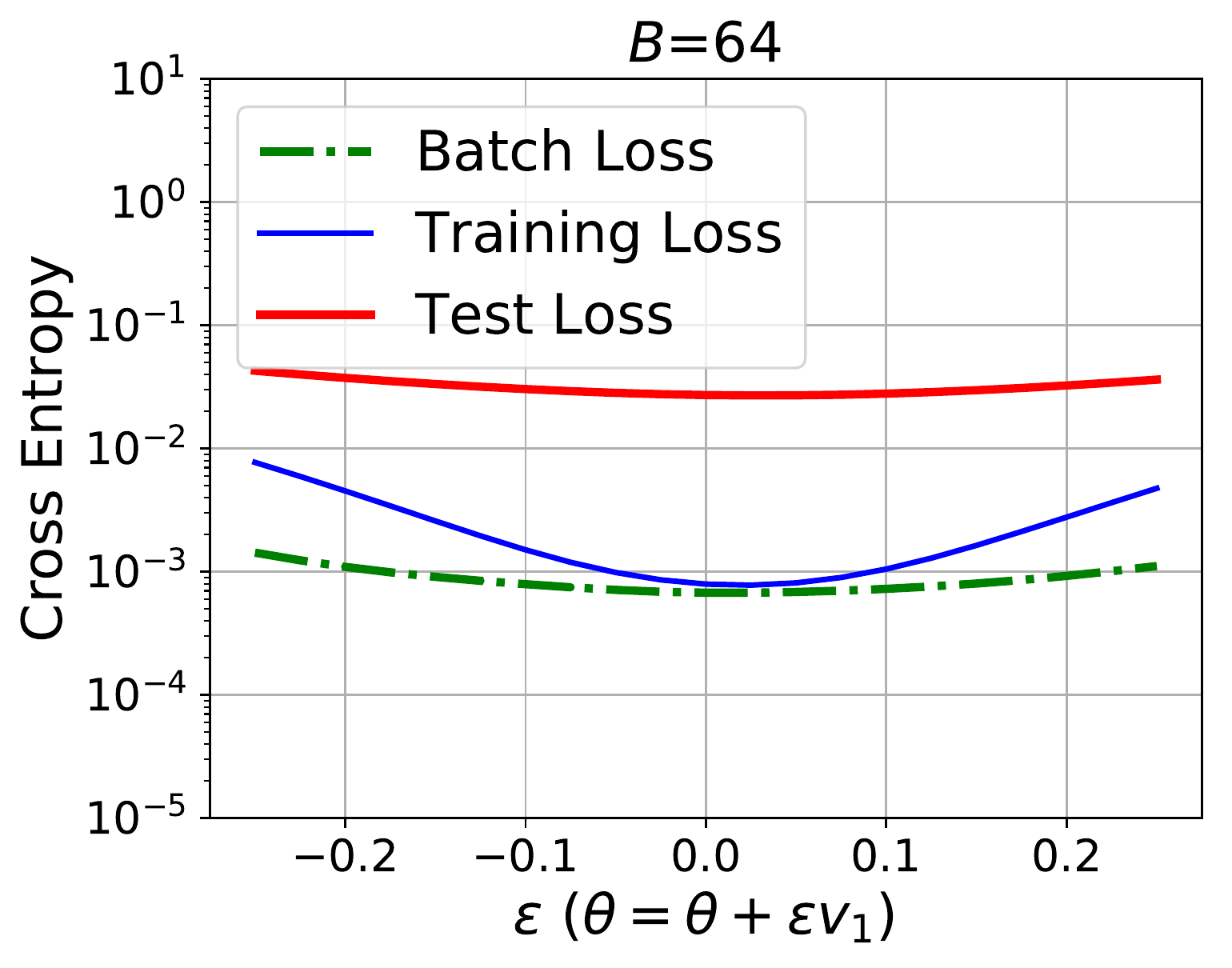}
  \includegraphics[width=.32\textwidth]{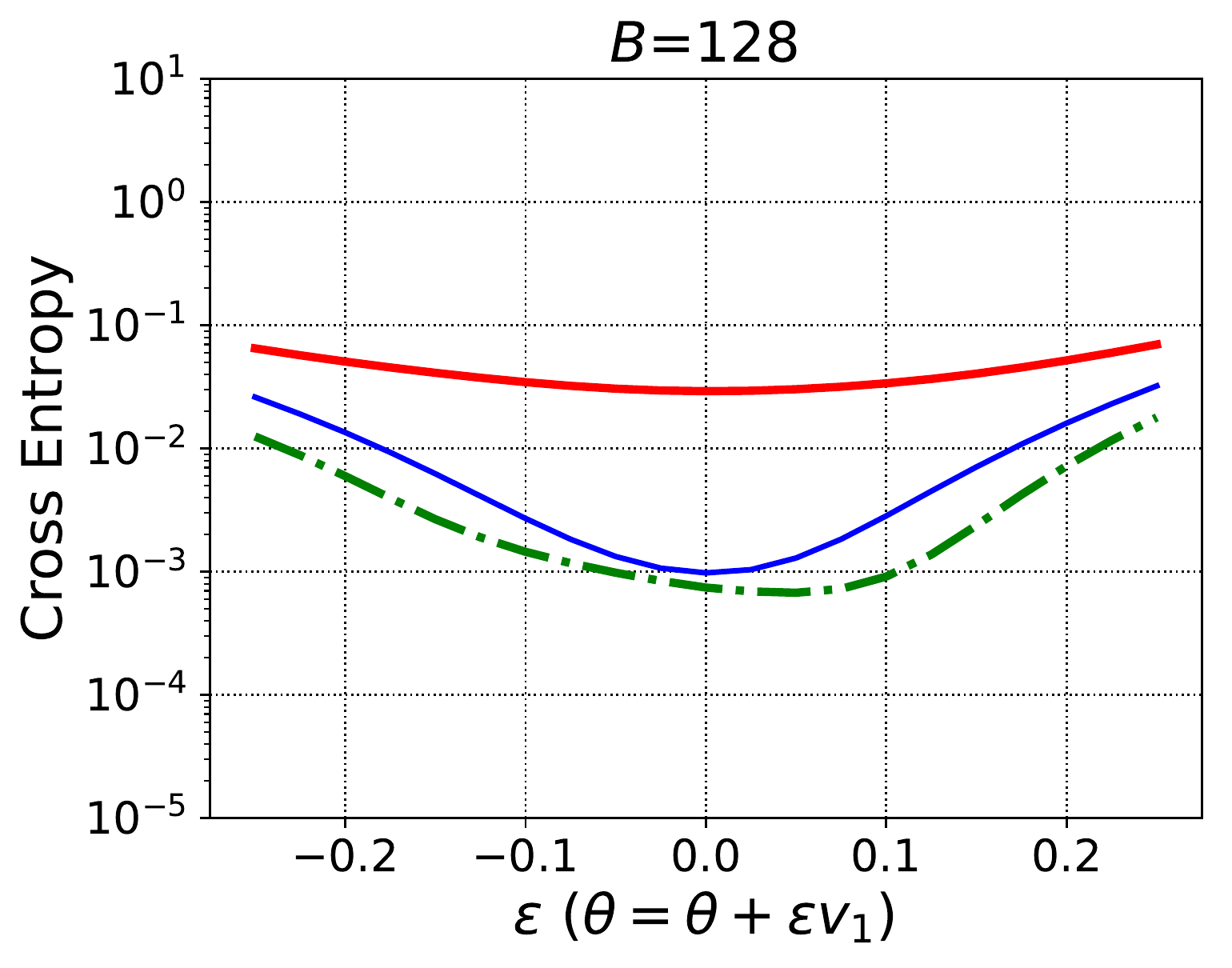}
  \includegraphics[width=.32\textwidth]{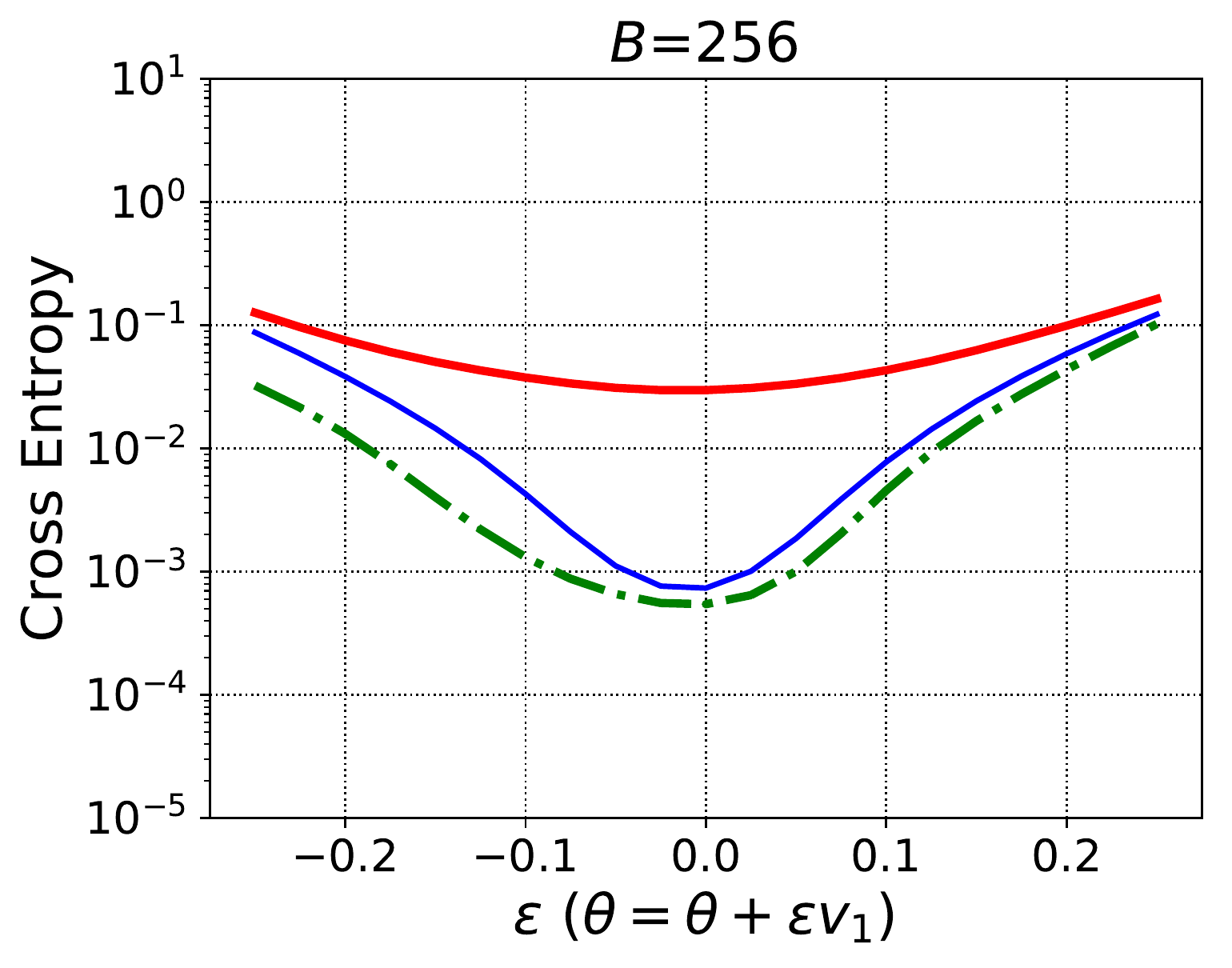}\\
  \includegraphics[width=.32\textwidth]{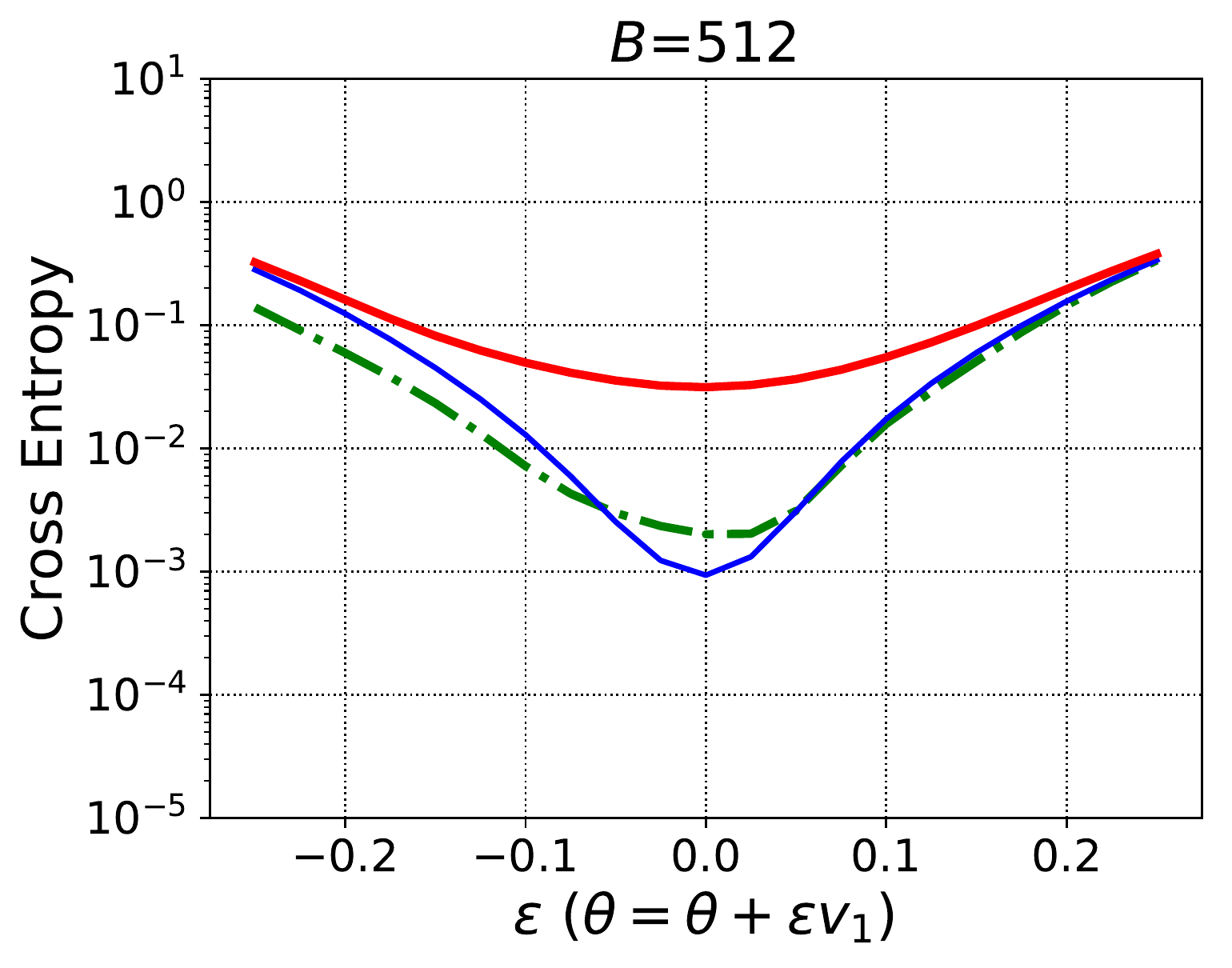}
  \includegraphics[width=.32\textwidth]{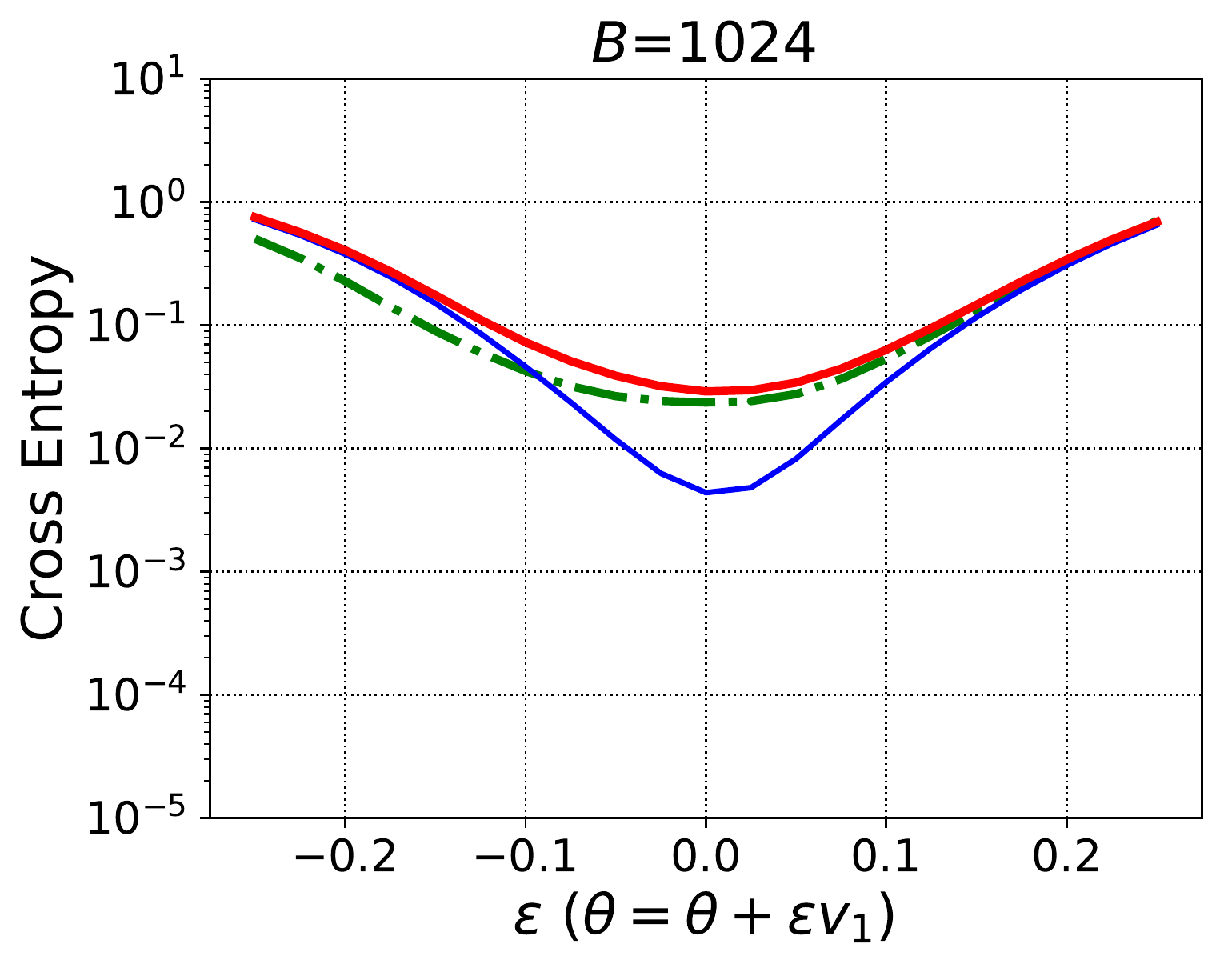}
  \includegraphics[width=.32\textwidth]{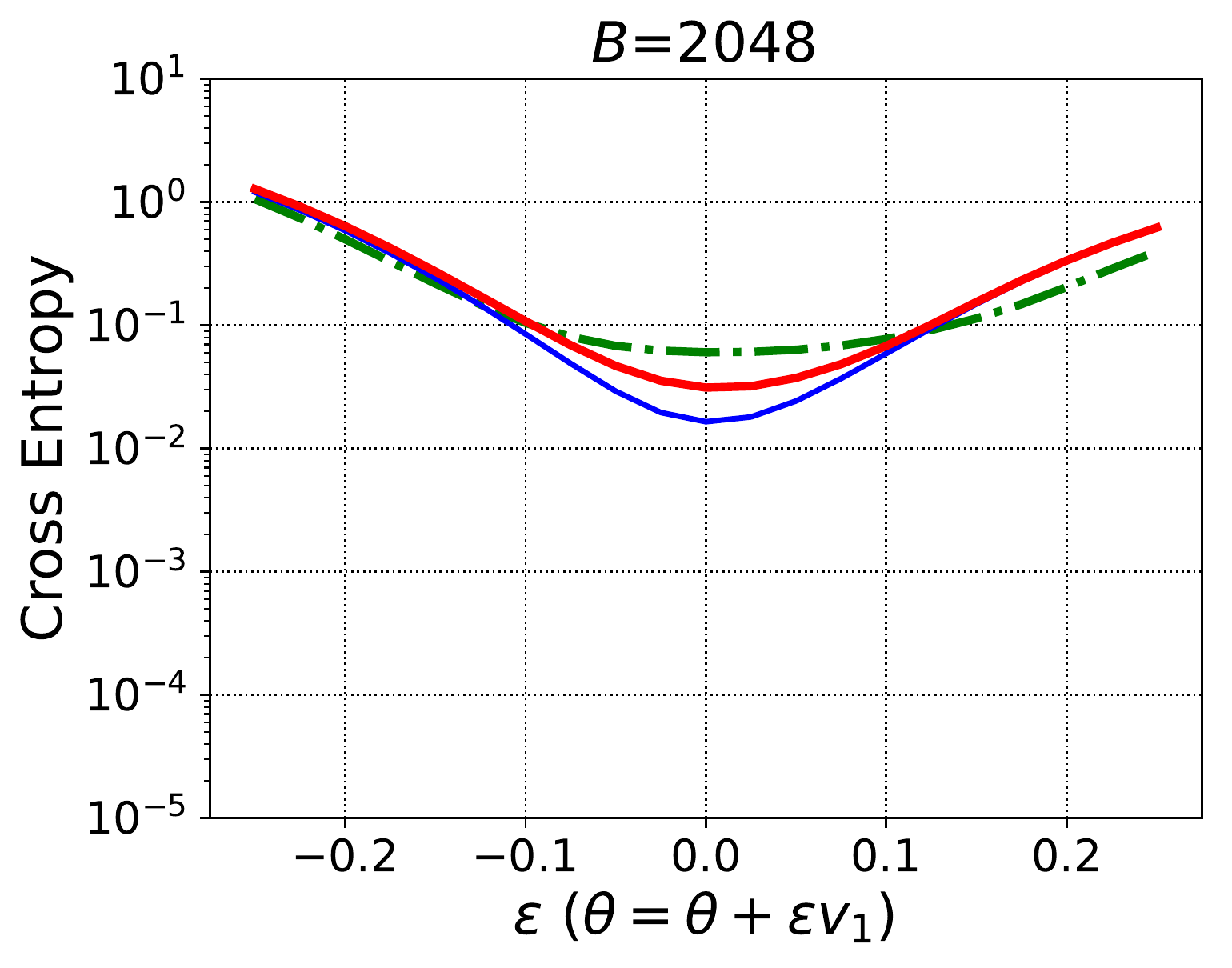}\\
\end{center}
\caption{
  The landscape of the loss functional is shown along the dominant eigenvector of the Hessian on MNIST for M1. Note that the $y-axis$ is in logarithm scale.
  Here $\epsilon$ is a scalar that perturbs the model parameters along the dominant eigenvector denoted by $v_1$. The green line is the loss for a randomly batch with batch-size 320 on MNIST. The blue and red line are the training and test loss, respectively. From the figure we could see that the curvature of test loss is much larger than training.
}
\label{f:landscape_largebatch_lenet}
\end{figure*}

% --------------MNIST-----------------------------%
% ---------------------------------------------%
\begin{figure*}[!htbp]
\label{fig:mnist_adv_eigenland}
\begin{center}
\includegraphics[width=.31\textwidth]{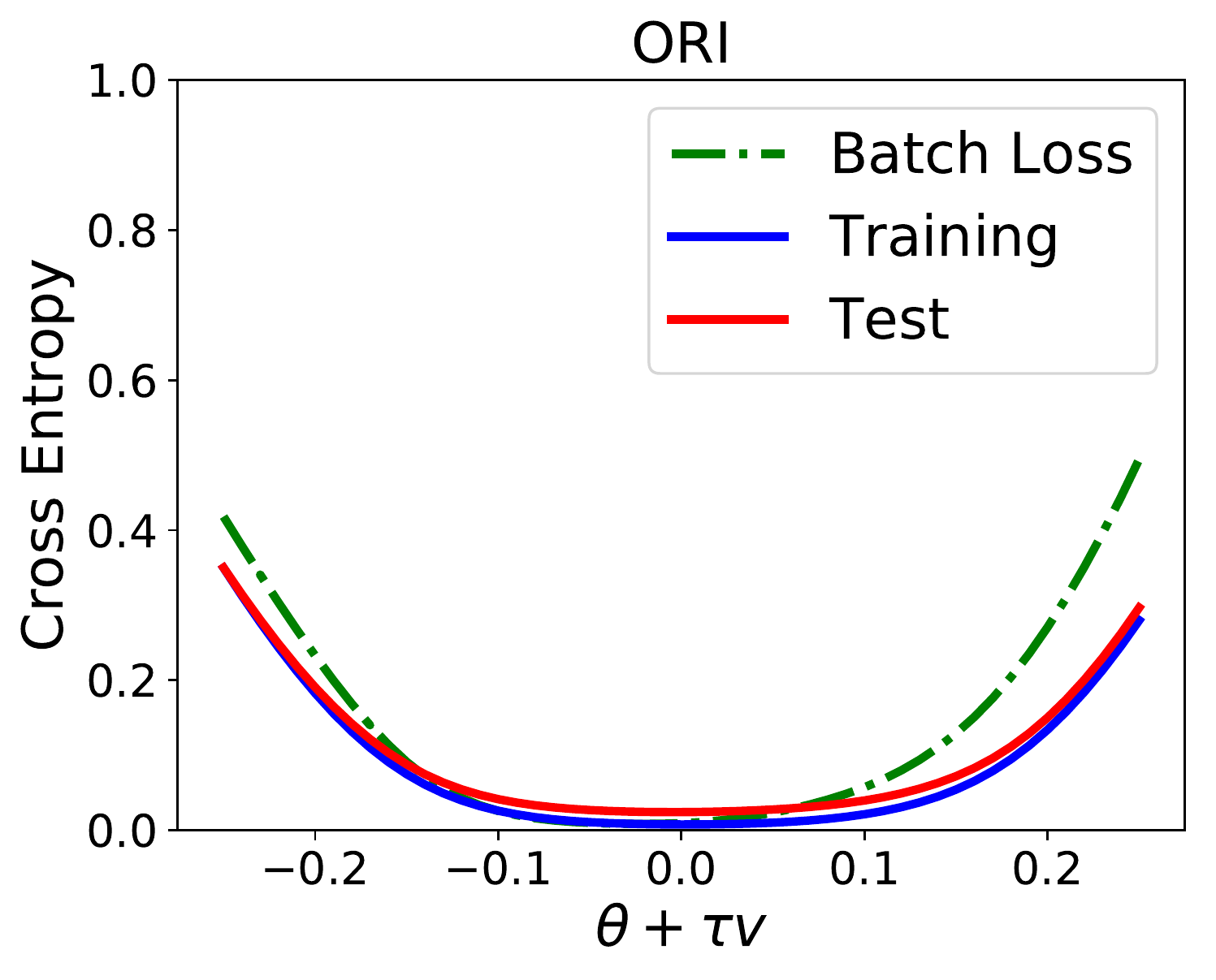}
\includegraphics[width=.31\textwidth]{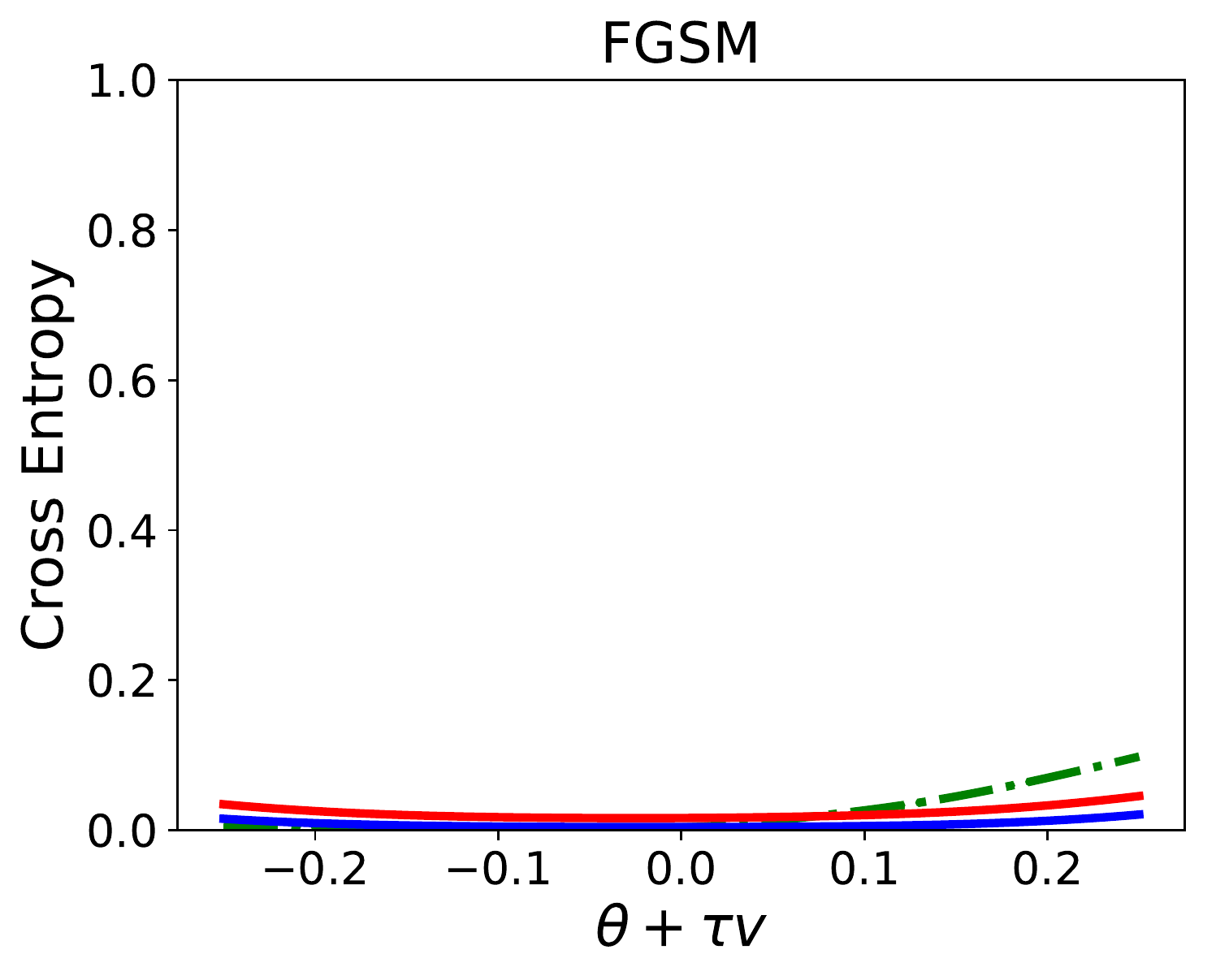}
\includegraphics[width=.31\textwidth]{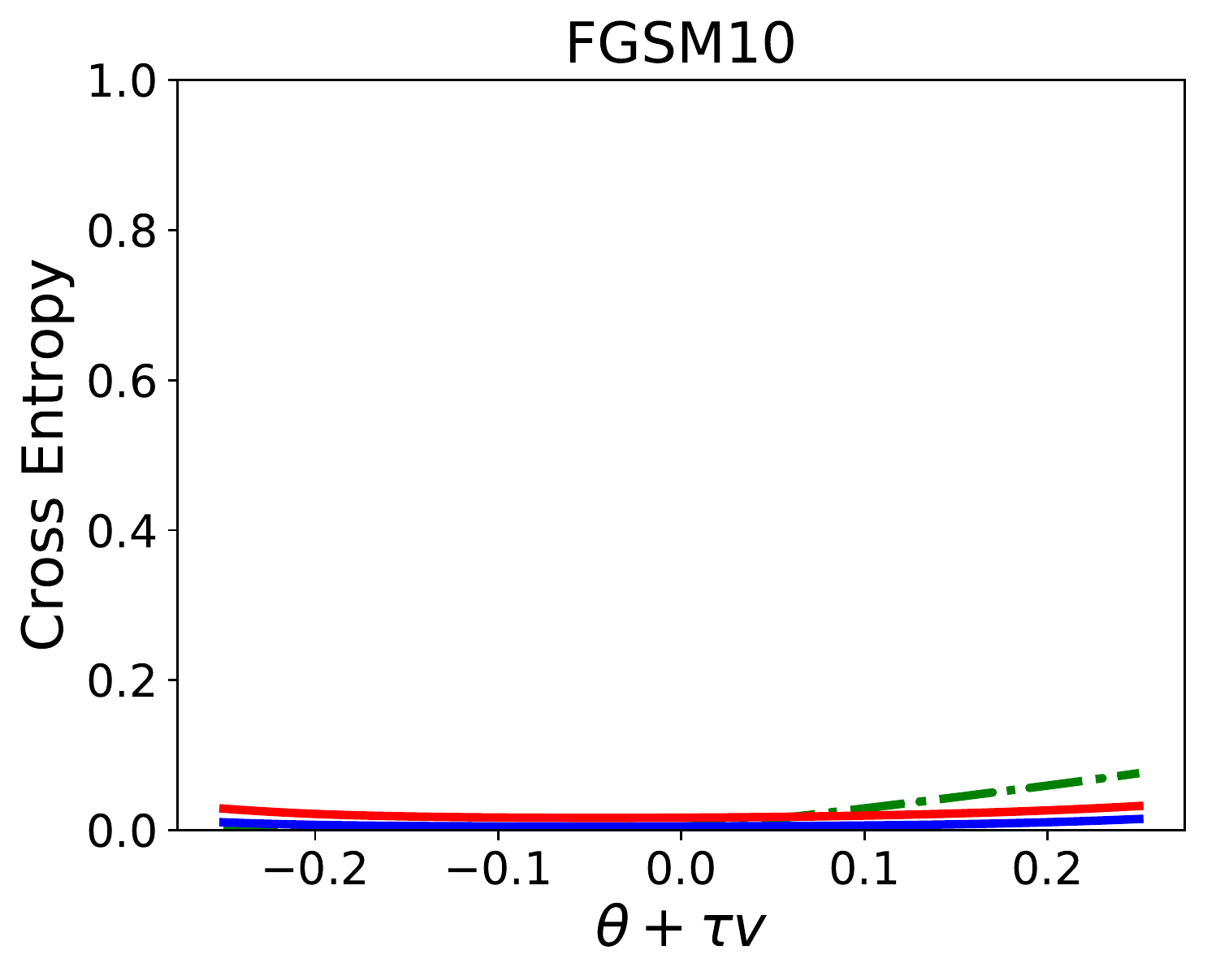}\\
\includegraphics[width=.31\textwidth]{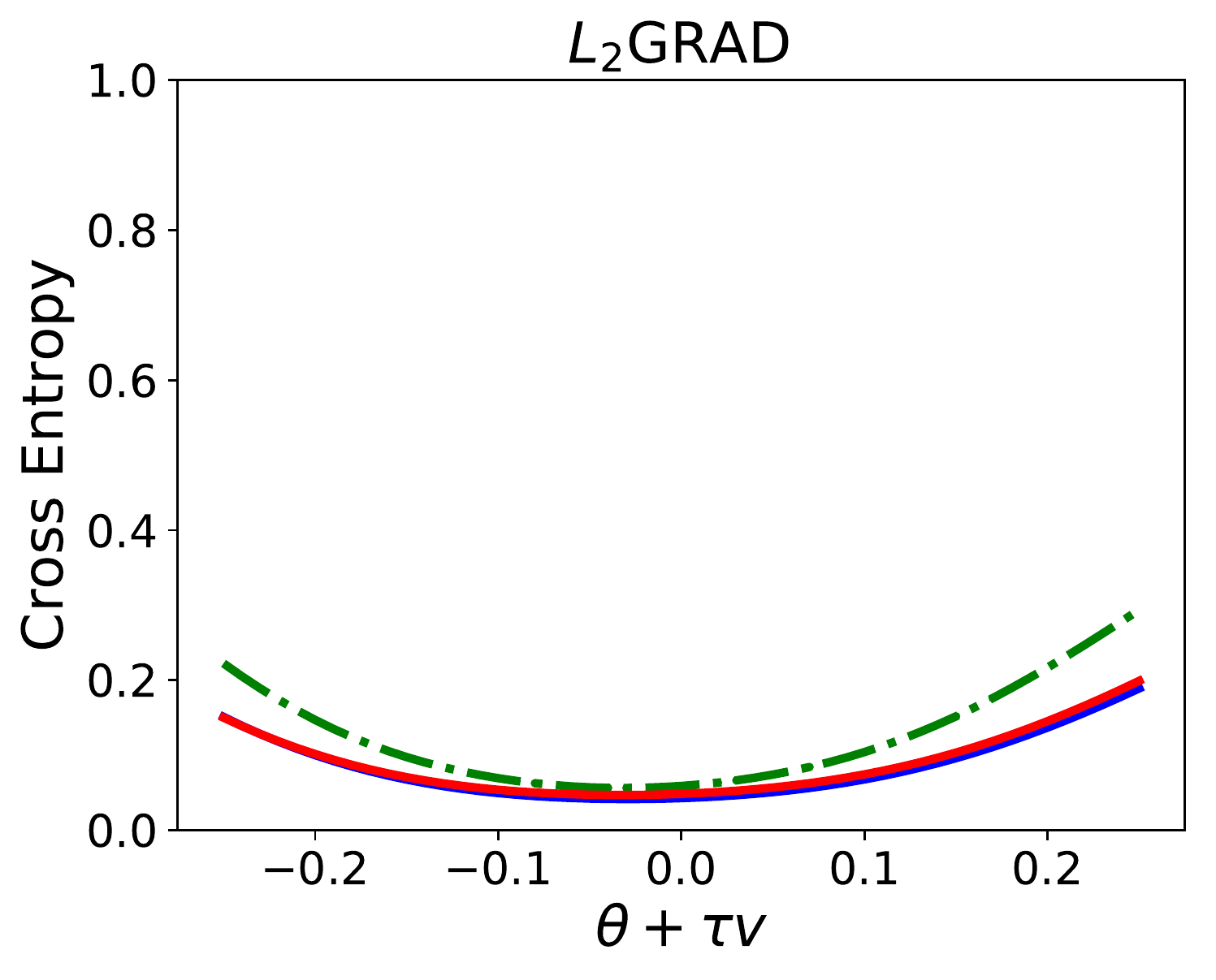}
\includegraphics[width=.31\textwidth]{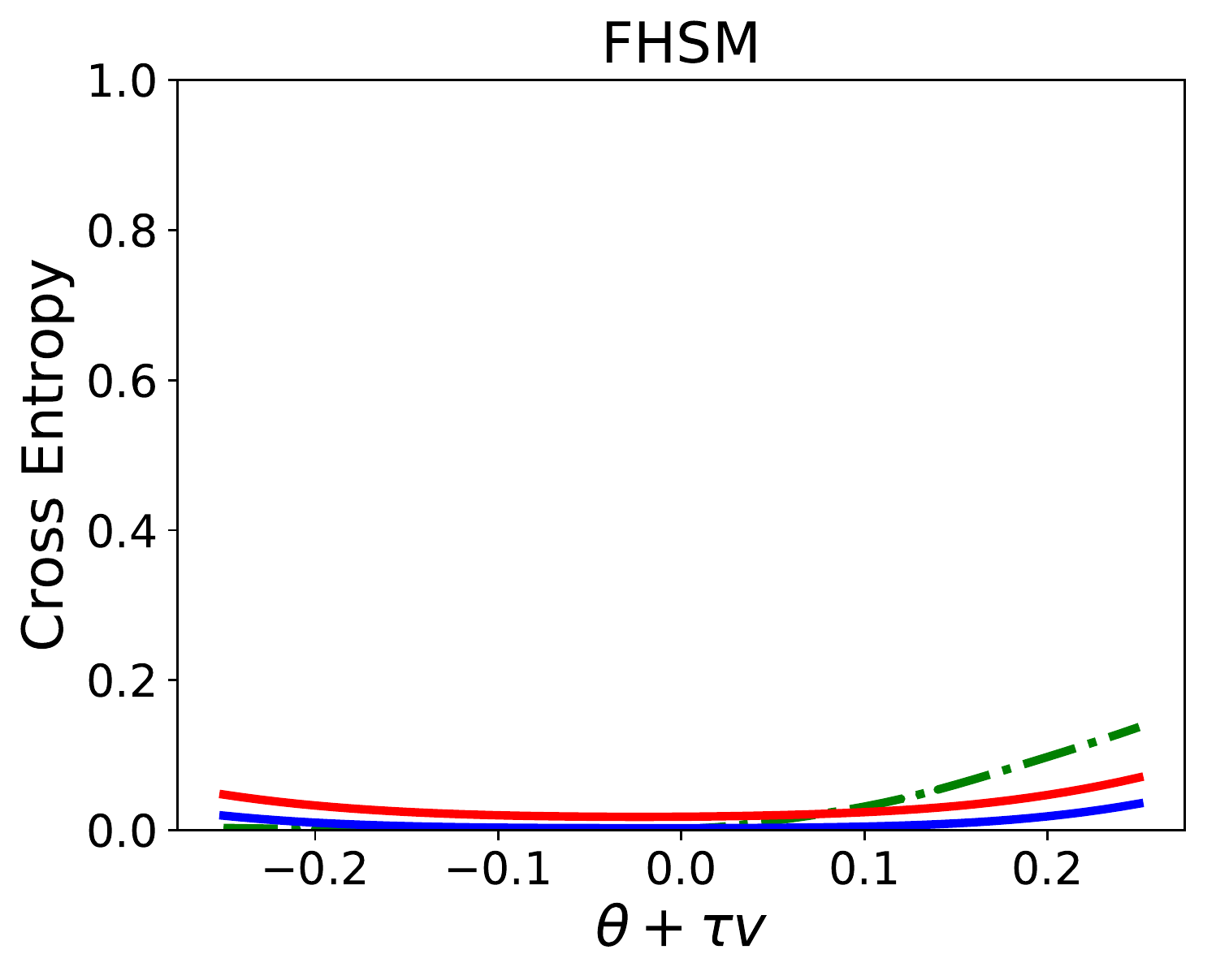}
\includegraphics[width=.31\textwidth]{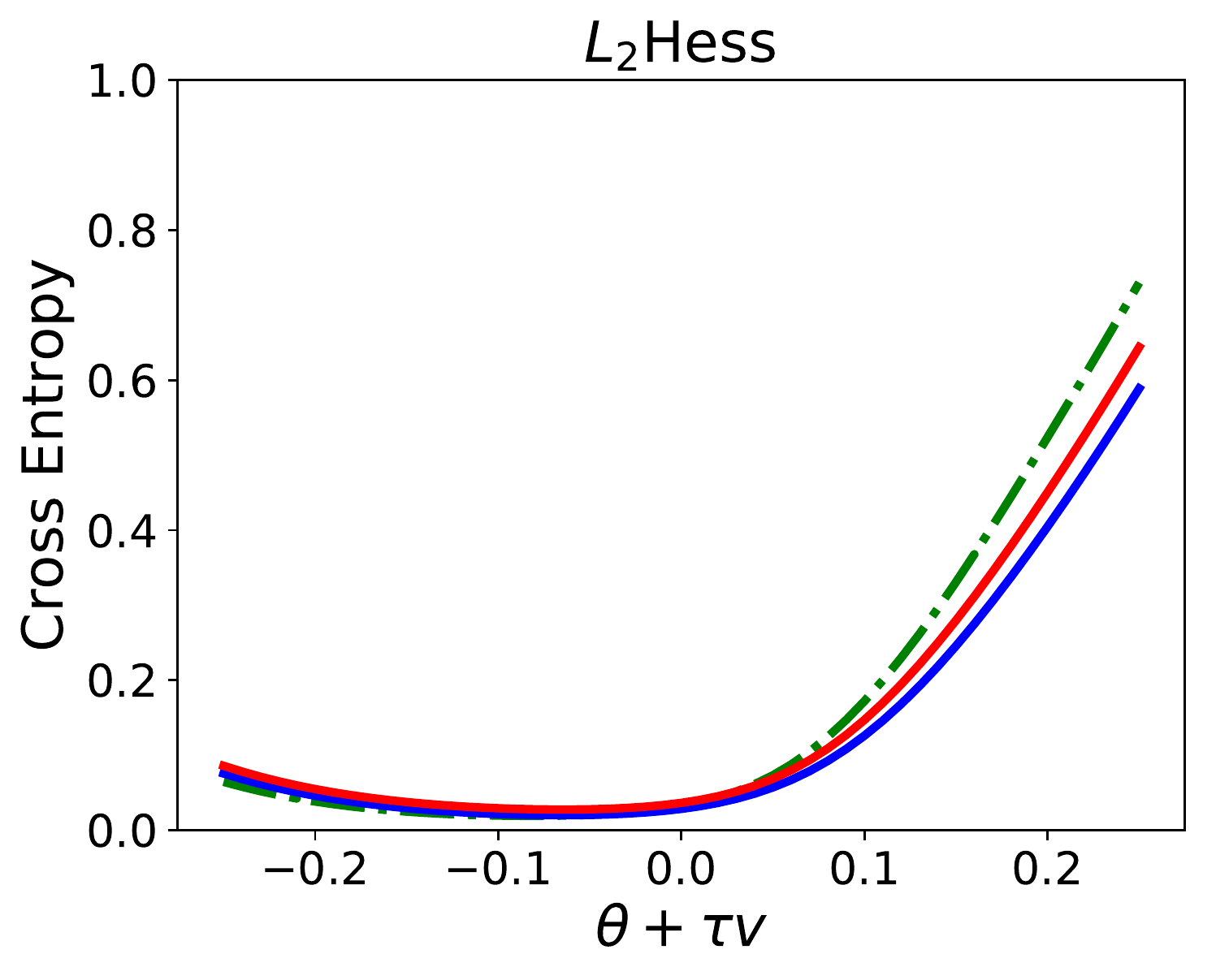}
\end{center}
\caption{
  We show the landscape of the test and training objective functional along the first eigenvector of the sub-sampled Hessian with $B=320$, i.e. 320 samples from training dataset, on MNIST for M1.
  We plot both the batch loss as well as the total training and test loss. One can see that visually the results show that the robust models
  converge to a region with smaller curvature.
}
\label{f:mnist_eig}
\end{figure*}
% ---------------------------------------------%
% ---------------------------------------------%

% ------------CIFAR-------------------------------%
% ---------------------------------------------%
\begin{figure*}[!htbp]
\label{fig:cifar_adv_eigenland}
\begin{center}
\includegraphics[width=.32\textwidth]{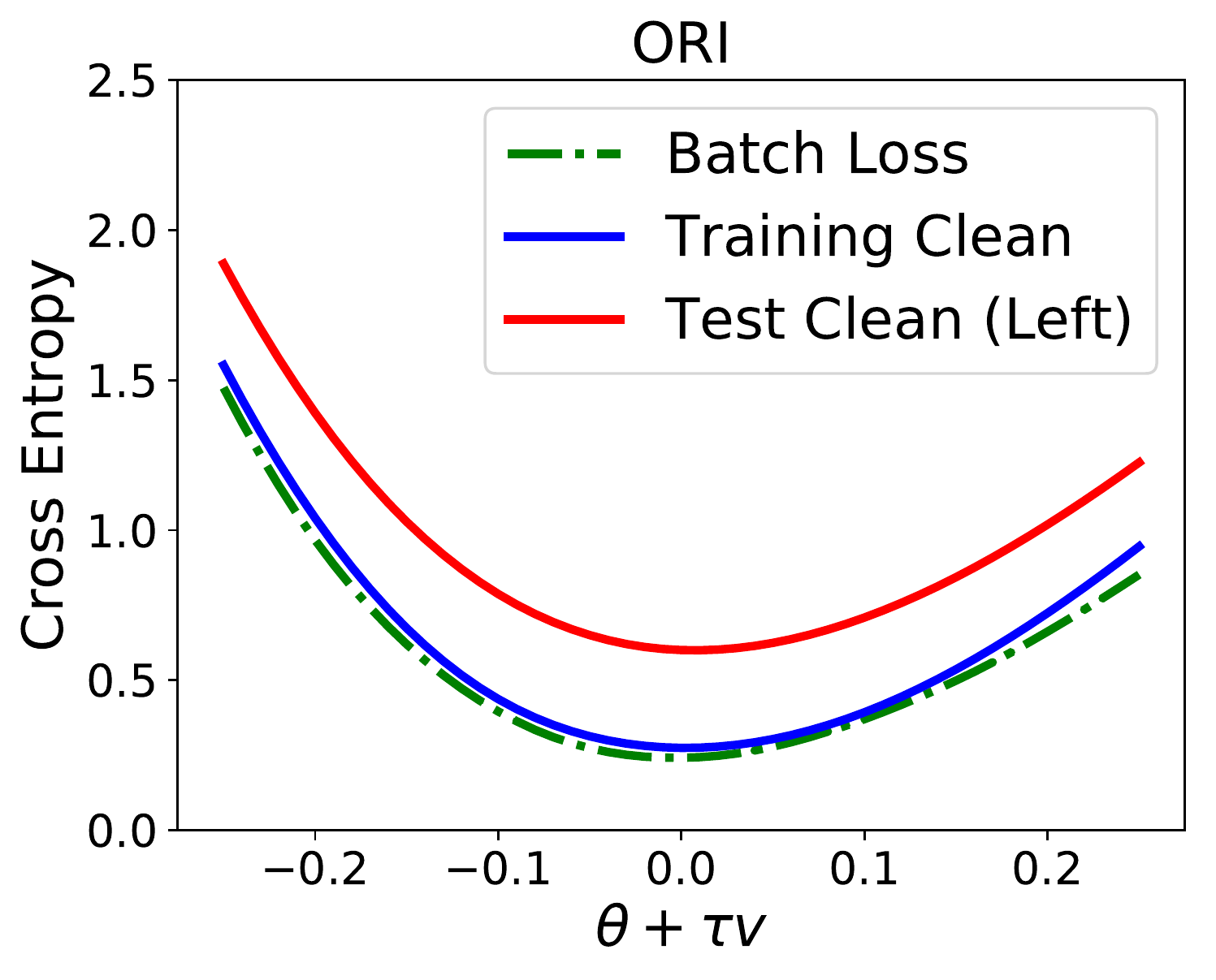}
\includegraphics[width=.32\textwidth]{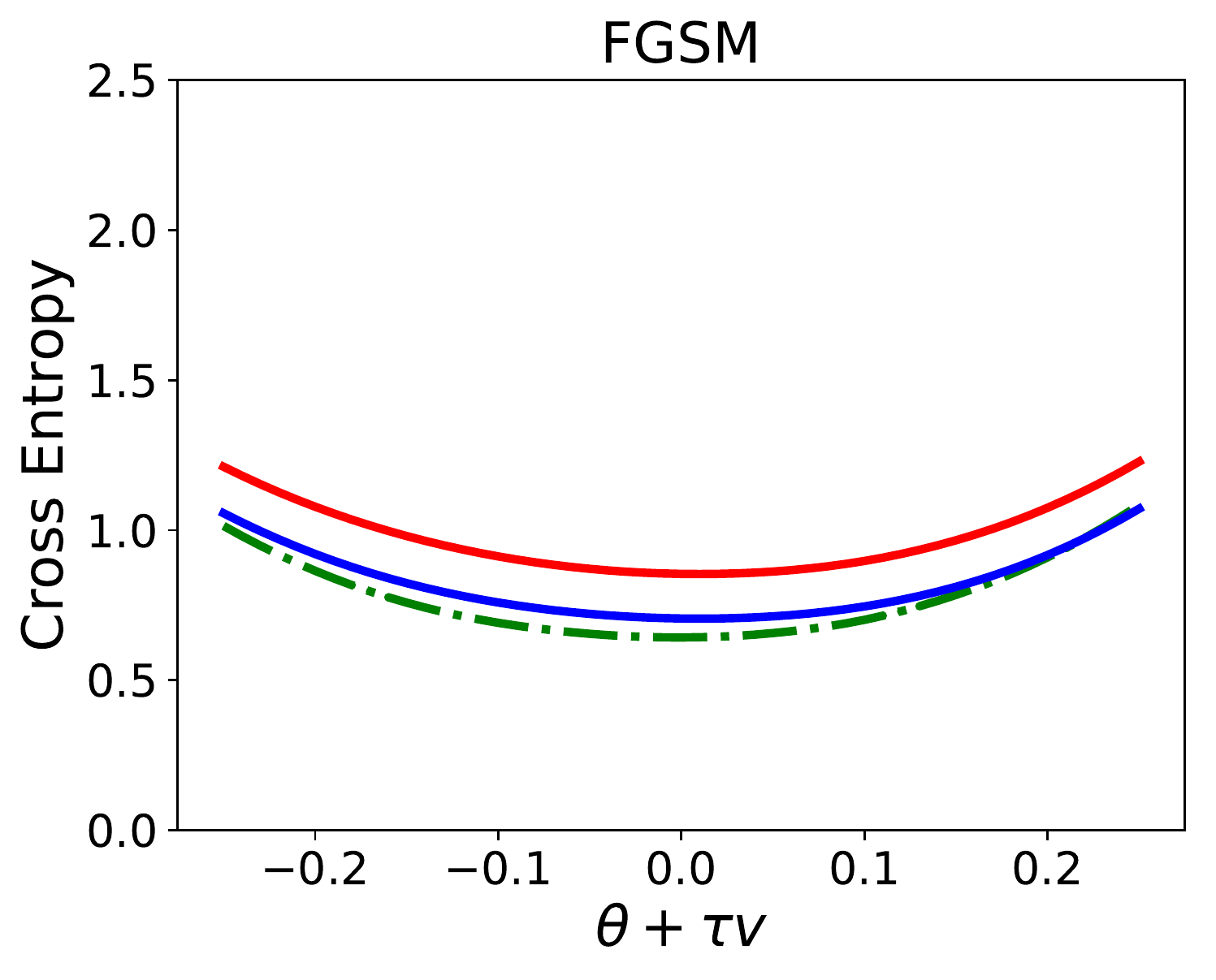}
\includegraphics[width=.32\textwidth]{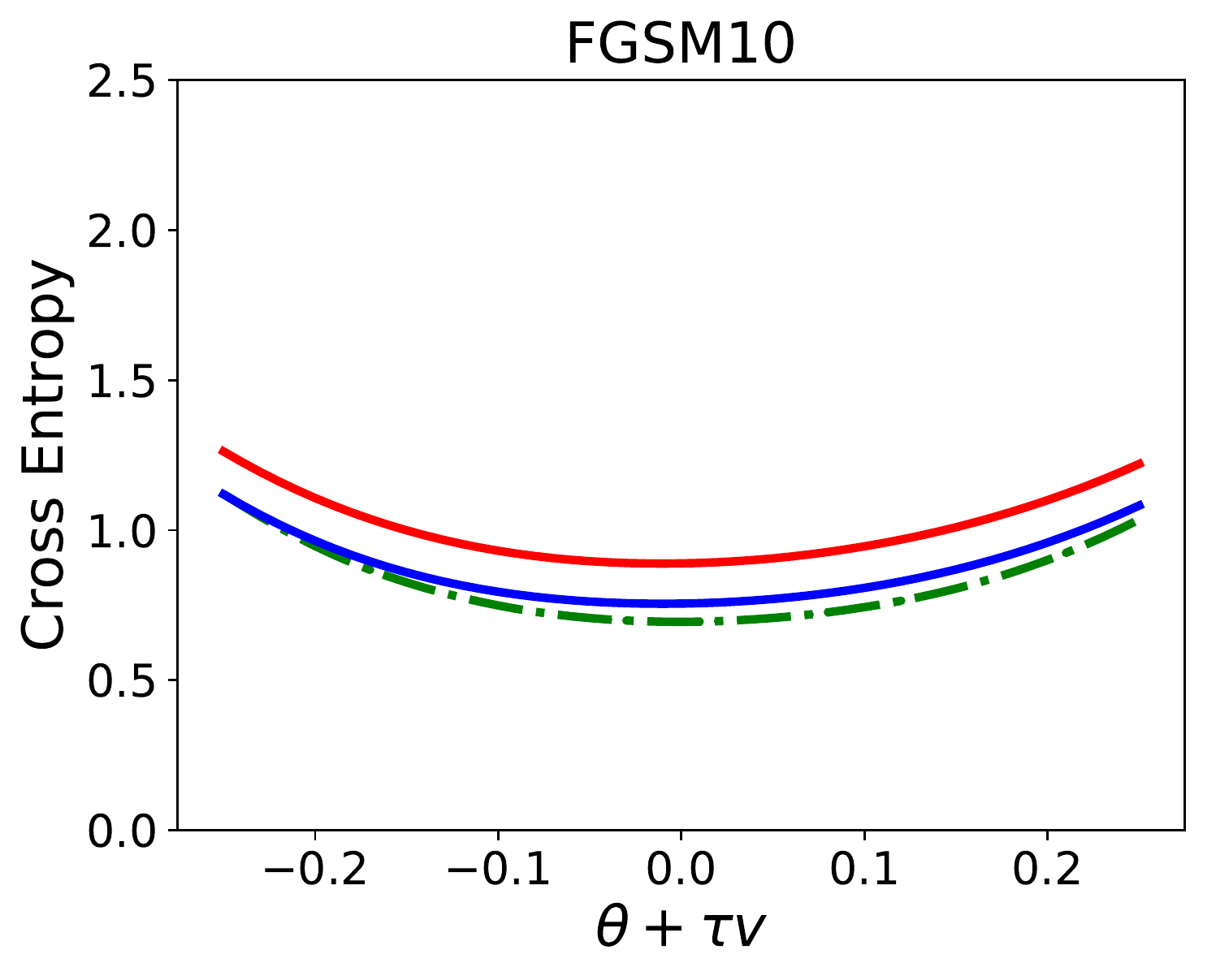}\\
\includegraphics[width=.32\textwidth]{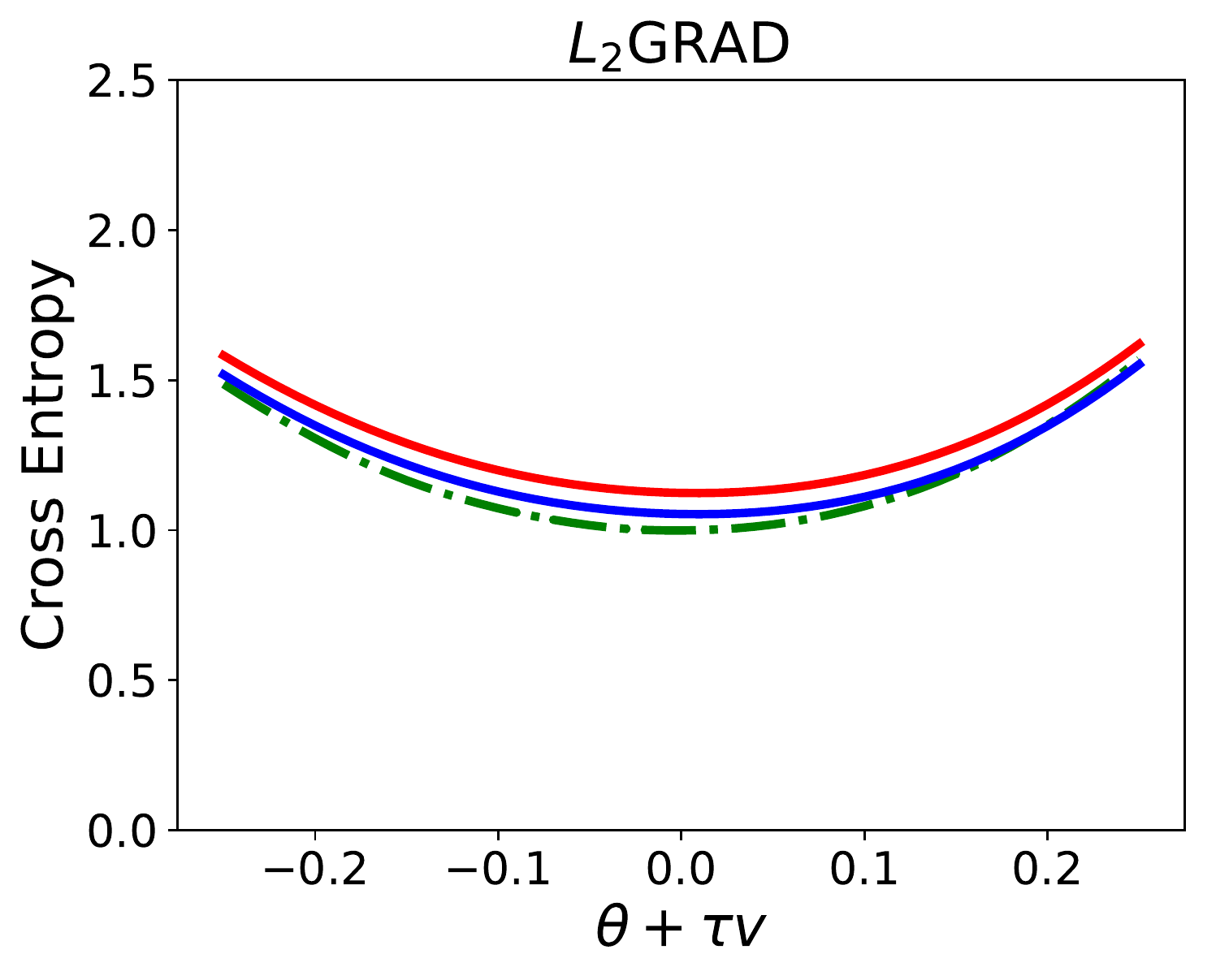}
\includegraphics[width=.32\textwidth]{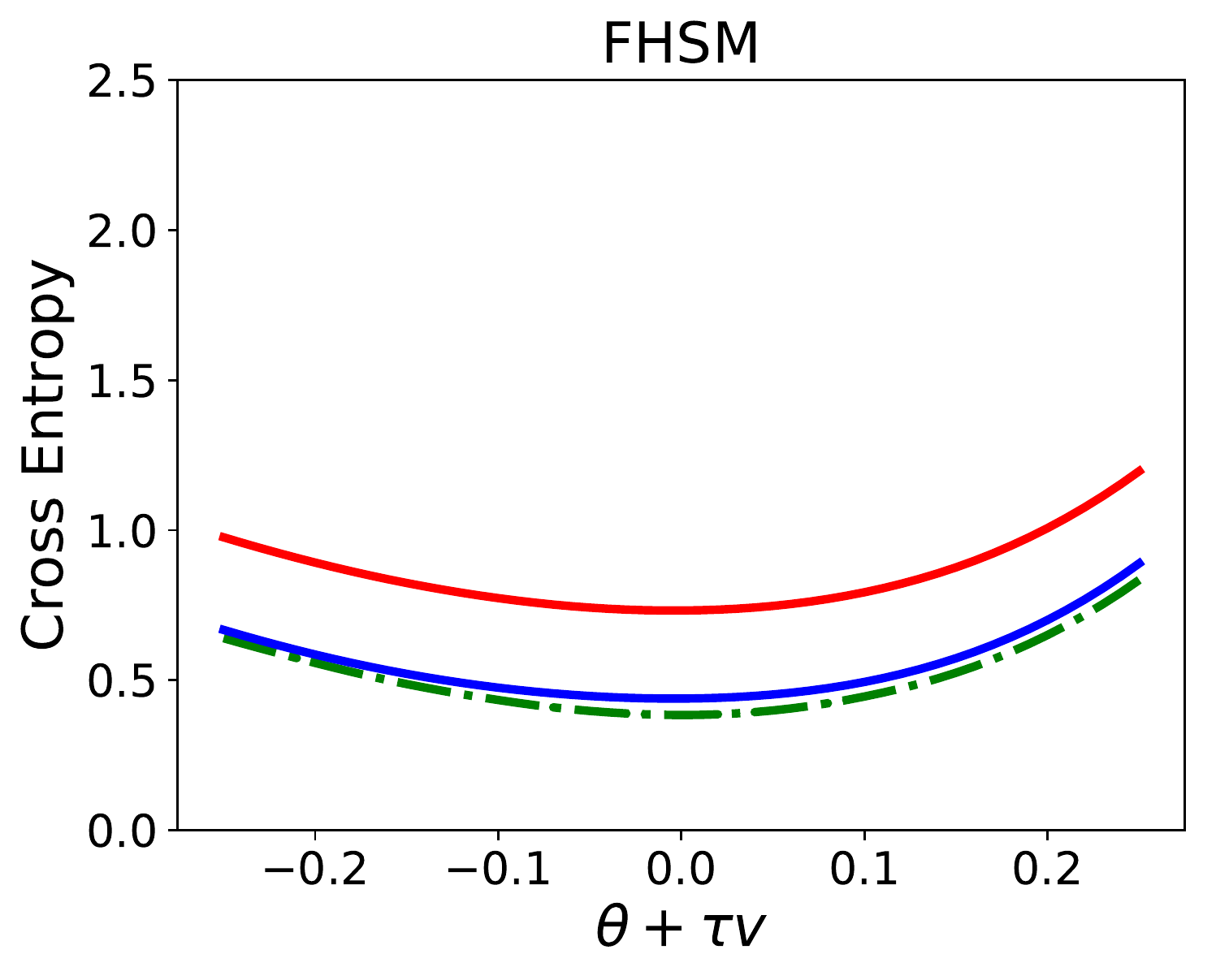}
\includegraphics[width=.32\textwidth]{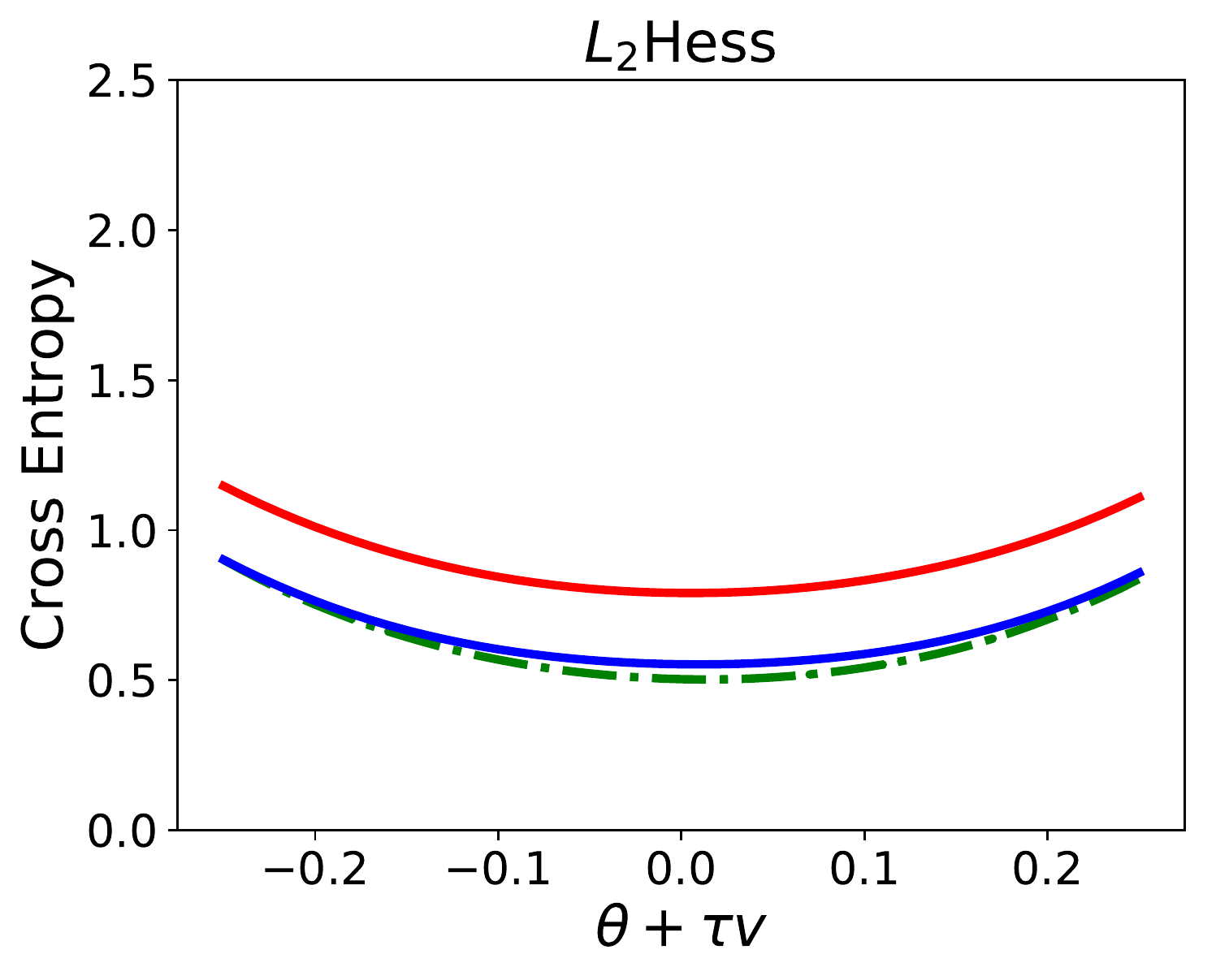}
\end{center}
\caption{
  We show the landscape of the test and training objective functional along the first eigenvector of the sub-sampled Hessian with $B=320$, i.e. 320 samples from training, on CIFAR-10 for C3.
  We plot both the batch loss as well as the total training and test loss. One can see that visually the results show that the curvature of robust models is smaller.
}
\end{figure*}

% -----------------MNIST--------------------------
\begin{figure*}[h]
\begin{center}
\includegraphics[width=.32\textwidth]{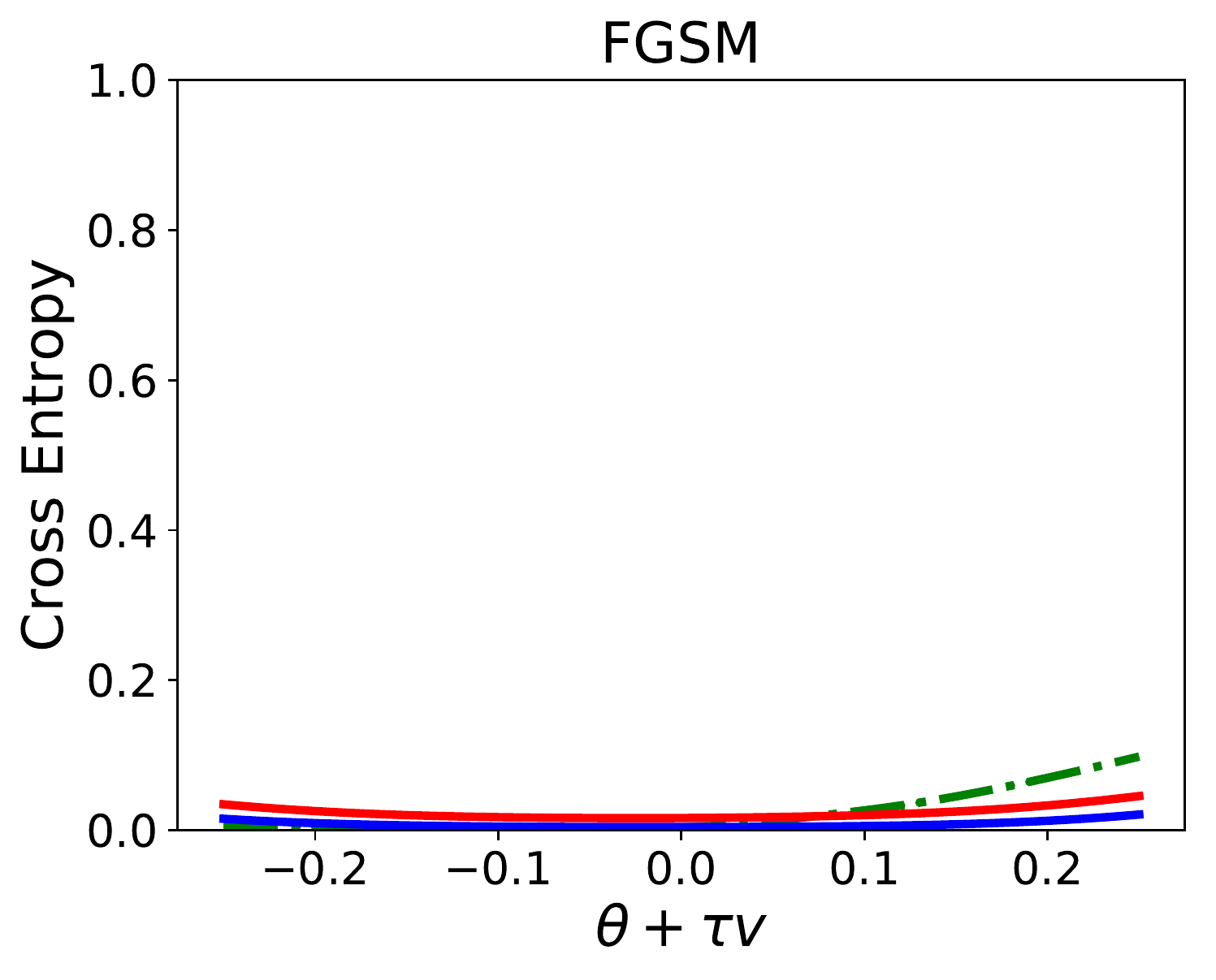} 
\includegraphics[width=.32\textwidth]{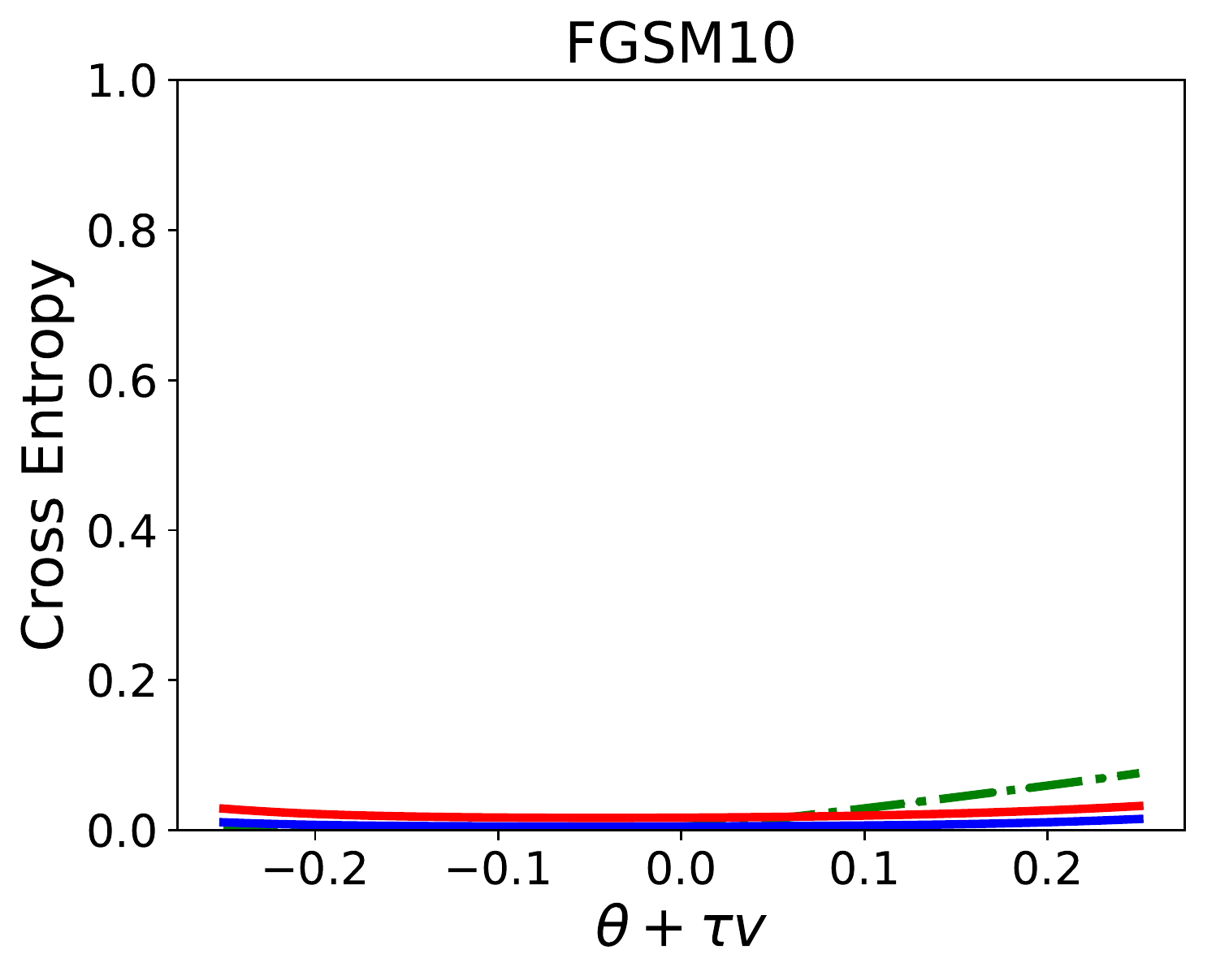}
\includegraphics[width=.32\textwidth]{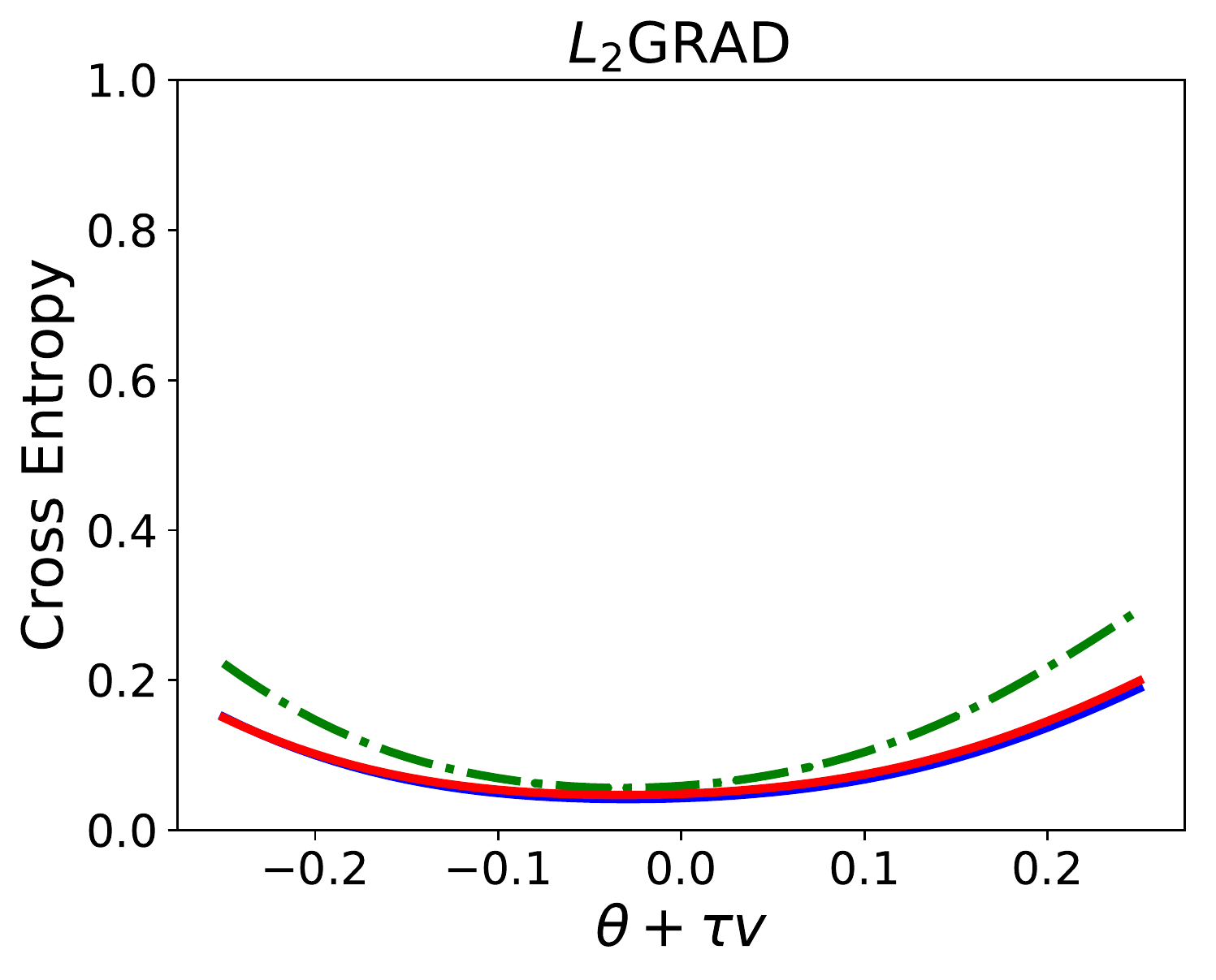}\\
\includegraphics[width=.32\textwidth]{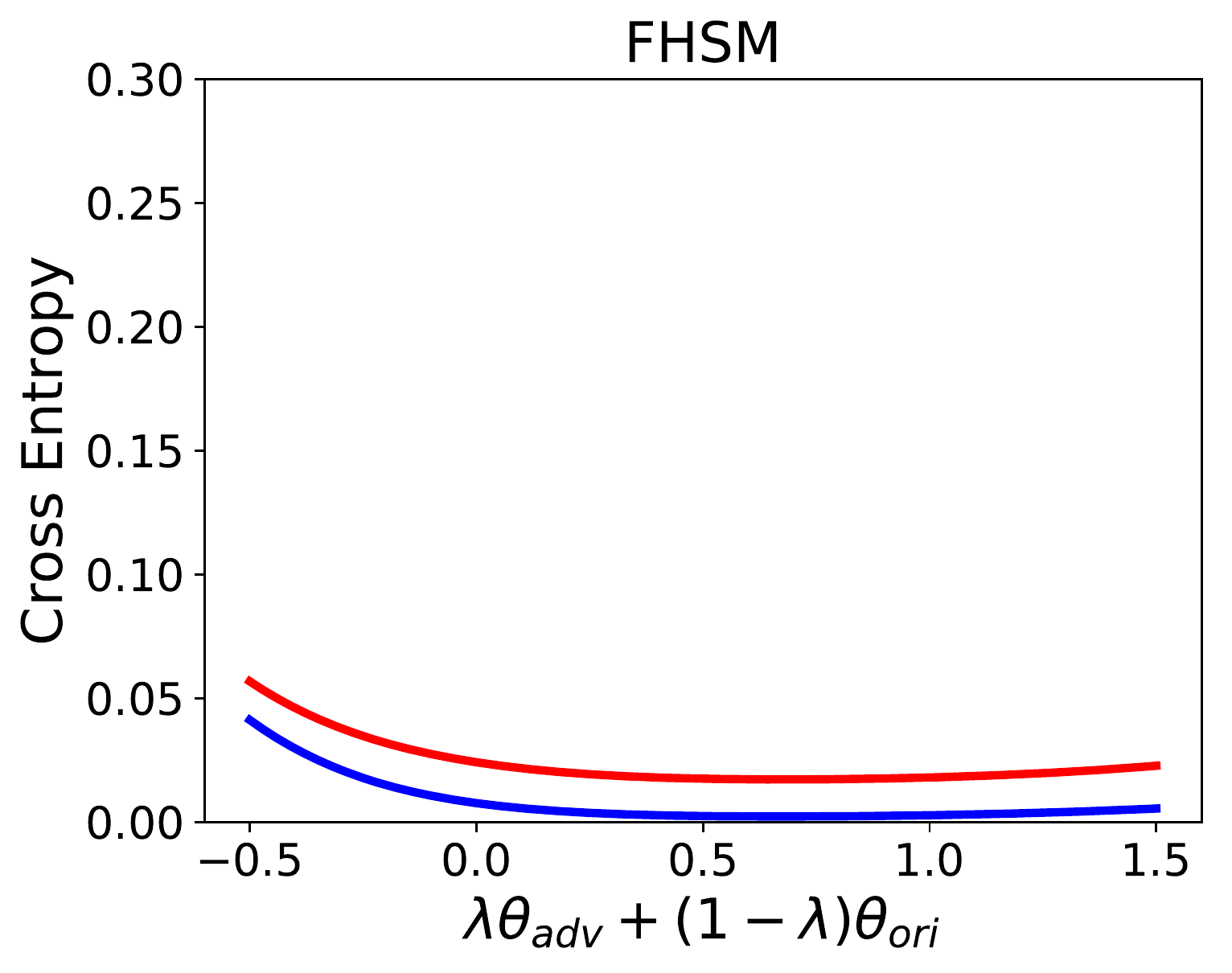}
\includegraphics[width=.32\textwidth]{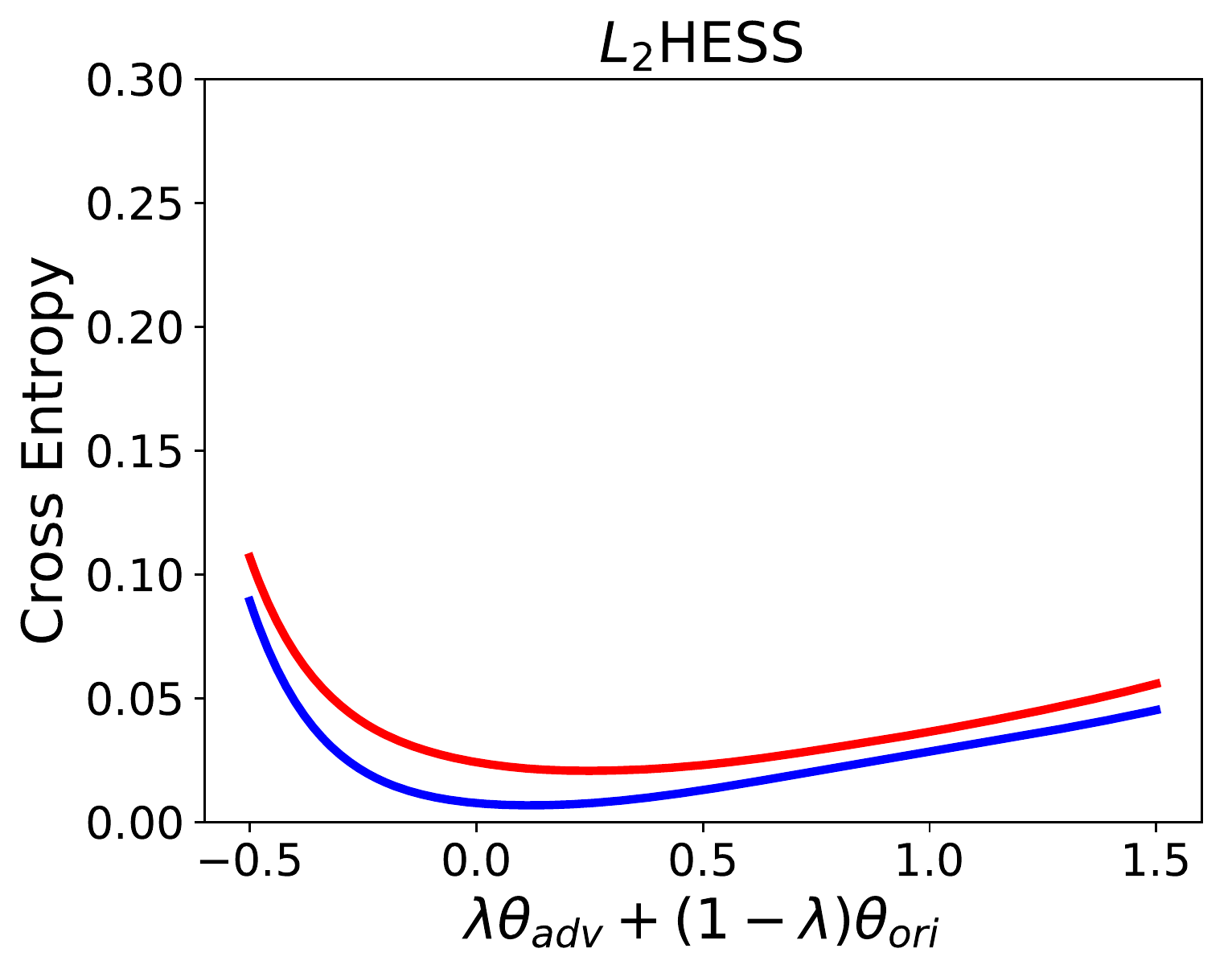}
\end{center}
\caption{1-D Parametric Plot on MNIST for M1 of $\M_{ORI}$ and adversarial models.
  Here we are showing how the landscape of the total loss functional changes when we interpolate from the original model ($\lambda=0$)
  to the robust model ($\lambda=1$). For all cases the robust model ends up at a point that has relatively smaller curvature
  compared to the original network. 
}
\label{fig:mnist_adv_interpolation}
\end{figure*}

% ---------------------------------------------%
% ---------------------------------------------%

\begin{figure*}[h]
\begin{center}
\includegraphics[width=.32\textwidth]{figures/fgsm.pdf}
\includegraphics[width=.32\textwidth]{figures/fgsm10.pdf}
\includegraphics[width=.32\textwidth]{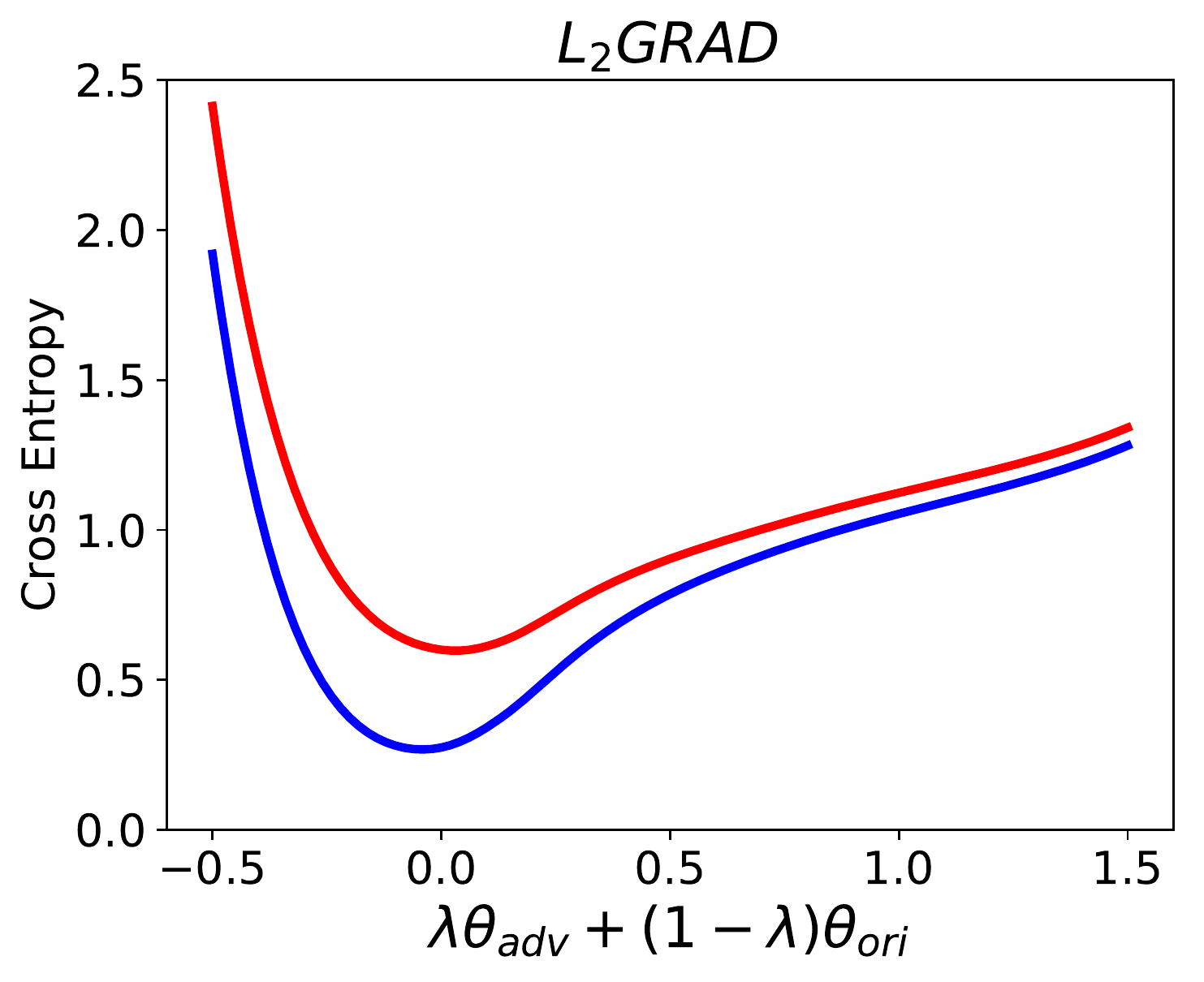}
\includegraphics[width=.32\textwidth]{figures/fasthessian.pdf}
\includegraphics[width=.32\textwidth]{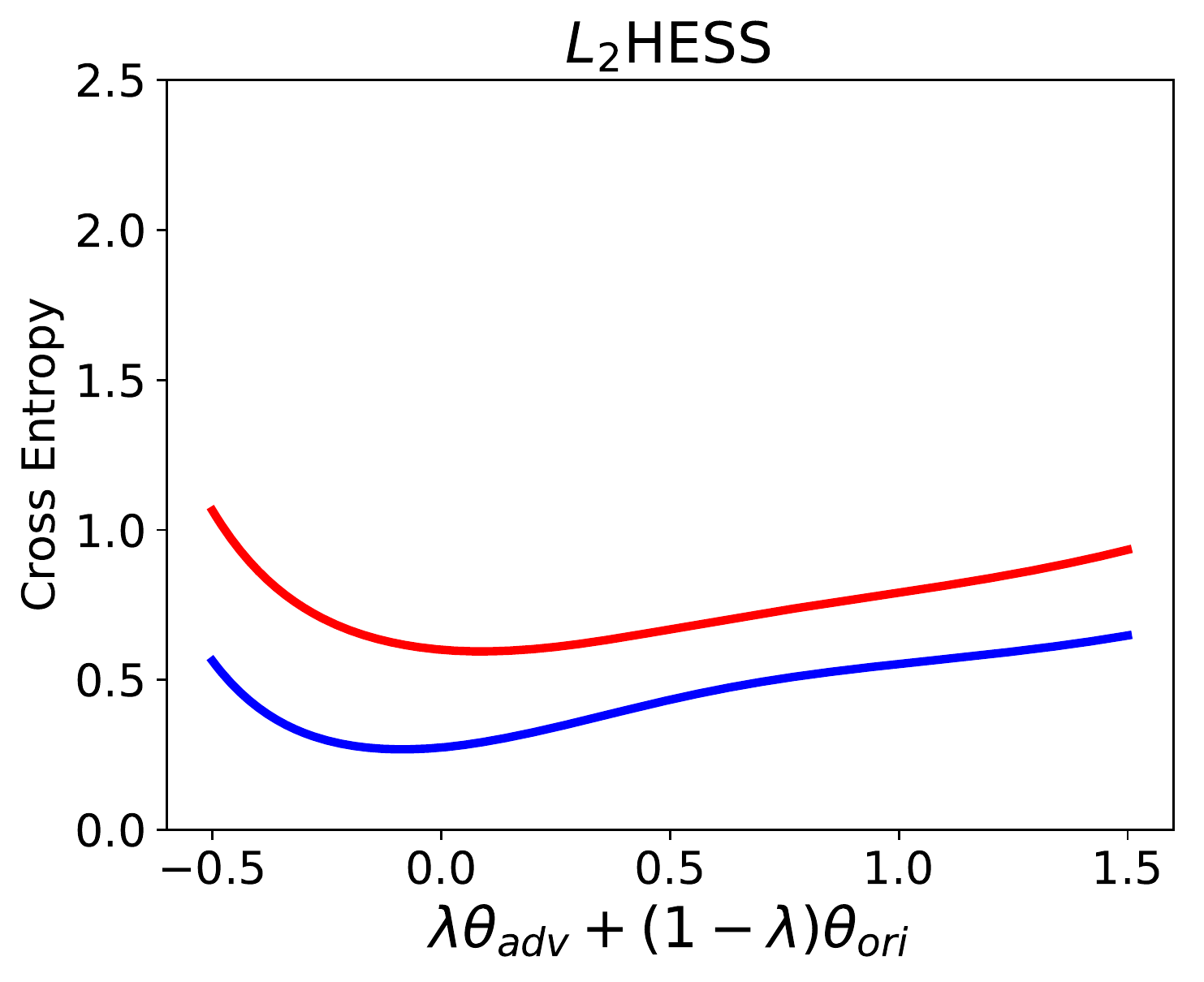}
\end{center}
\caption{1-D Parametric Plot on CIFAR-10 of $\M_{ORI}$ and adversarial models, i.e. total loss functional changes interpolating from the original model ($\lambda=0$)
  to the robust model ($\lambda=1$). For all cases the robust model ends up at a point that has relatively smaller curvature
  compared to the original network. 
}
\label{fig:cifar_adv_interpolation_app}
\end{figure*}

% --------------------------------------------------------------------------------------
% ------------------MNIST------------------------%
% ---------------------------------------------%
\begin{figure*}[h]
\begin{center}
\includegraphics[width=.48\textwidth]{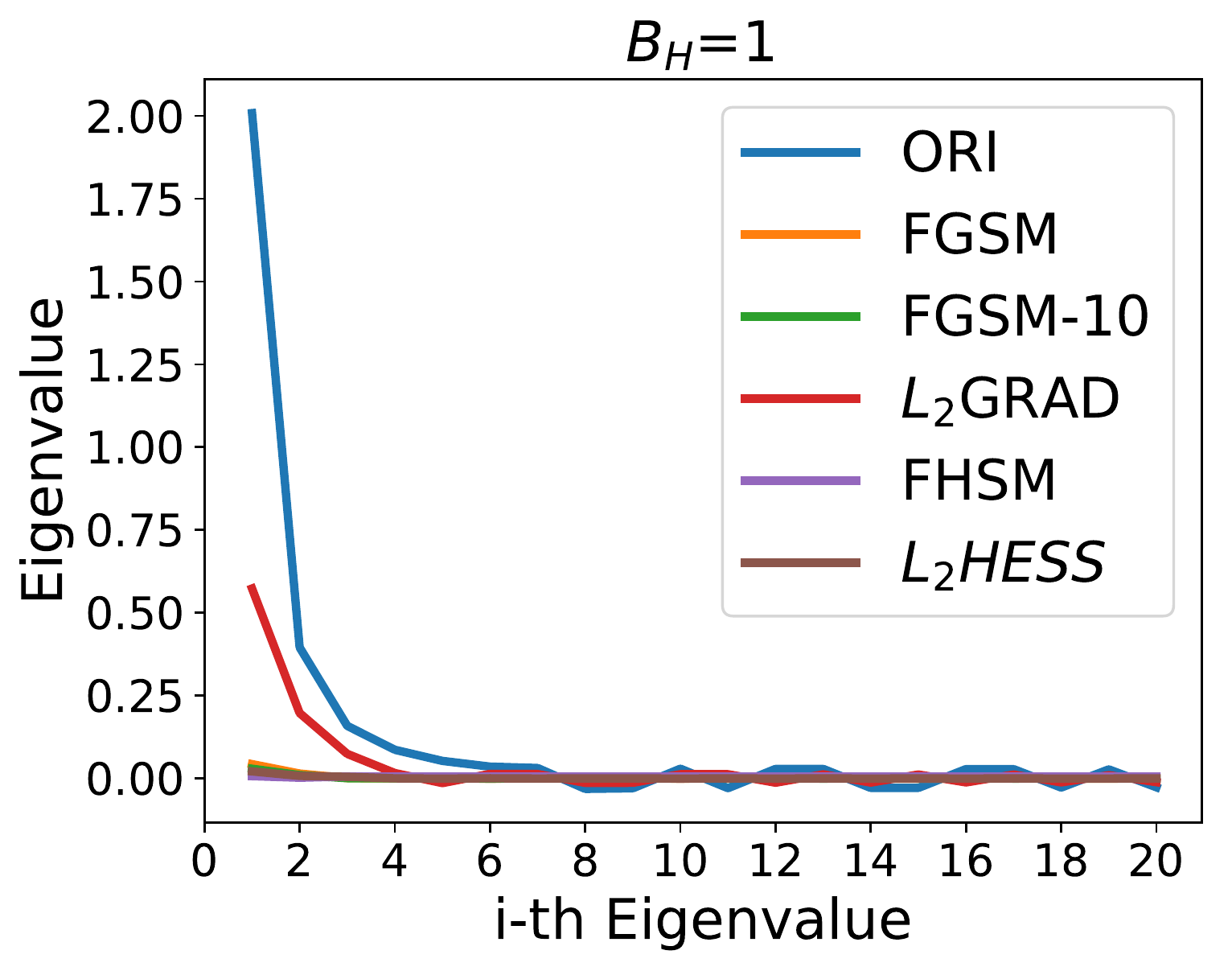}
\includegraphics[width=.48\textwidth]{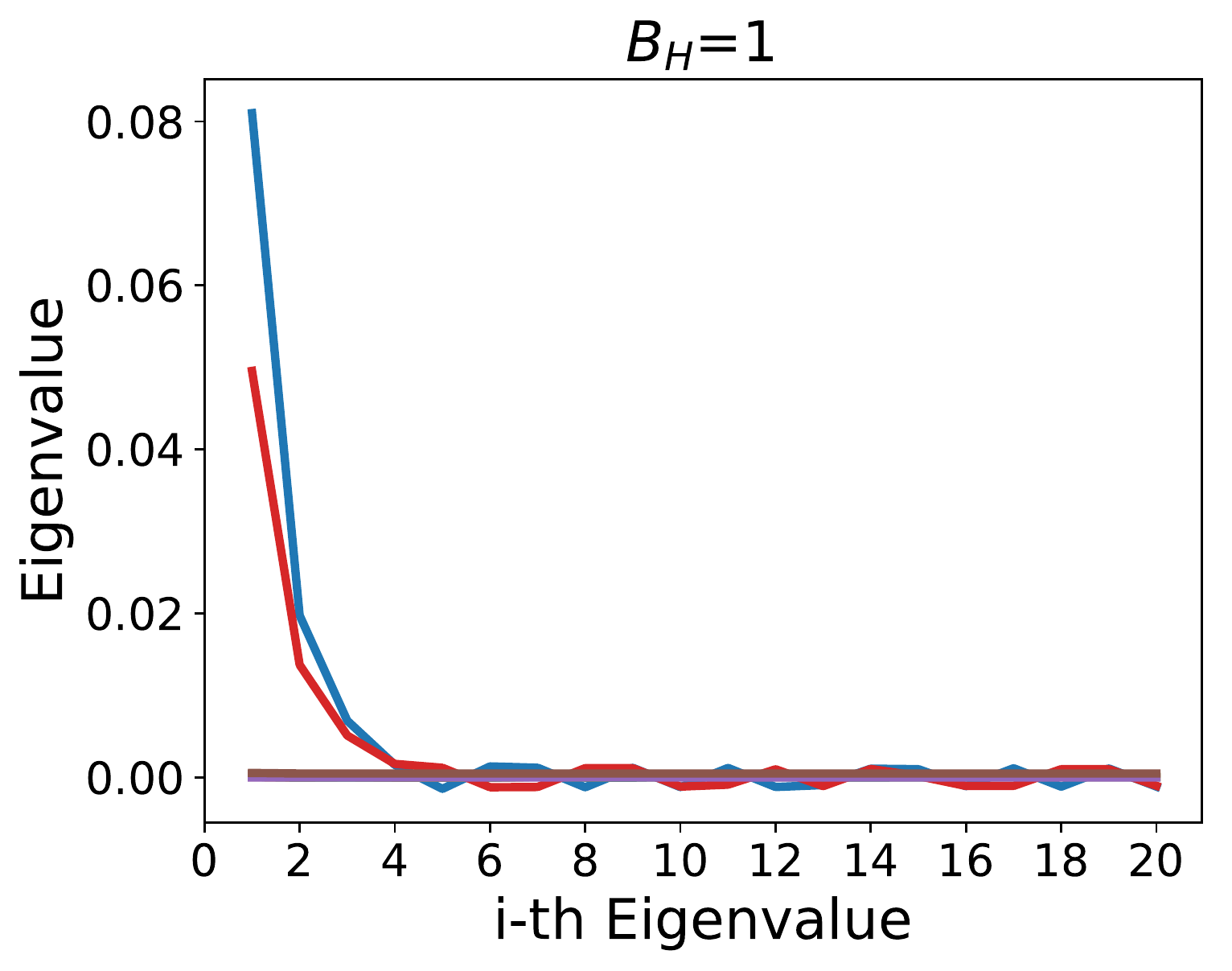}\\
\includegraphics[width=.48\textwidth]{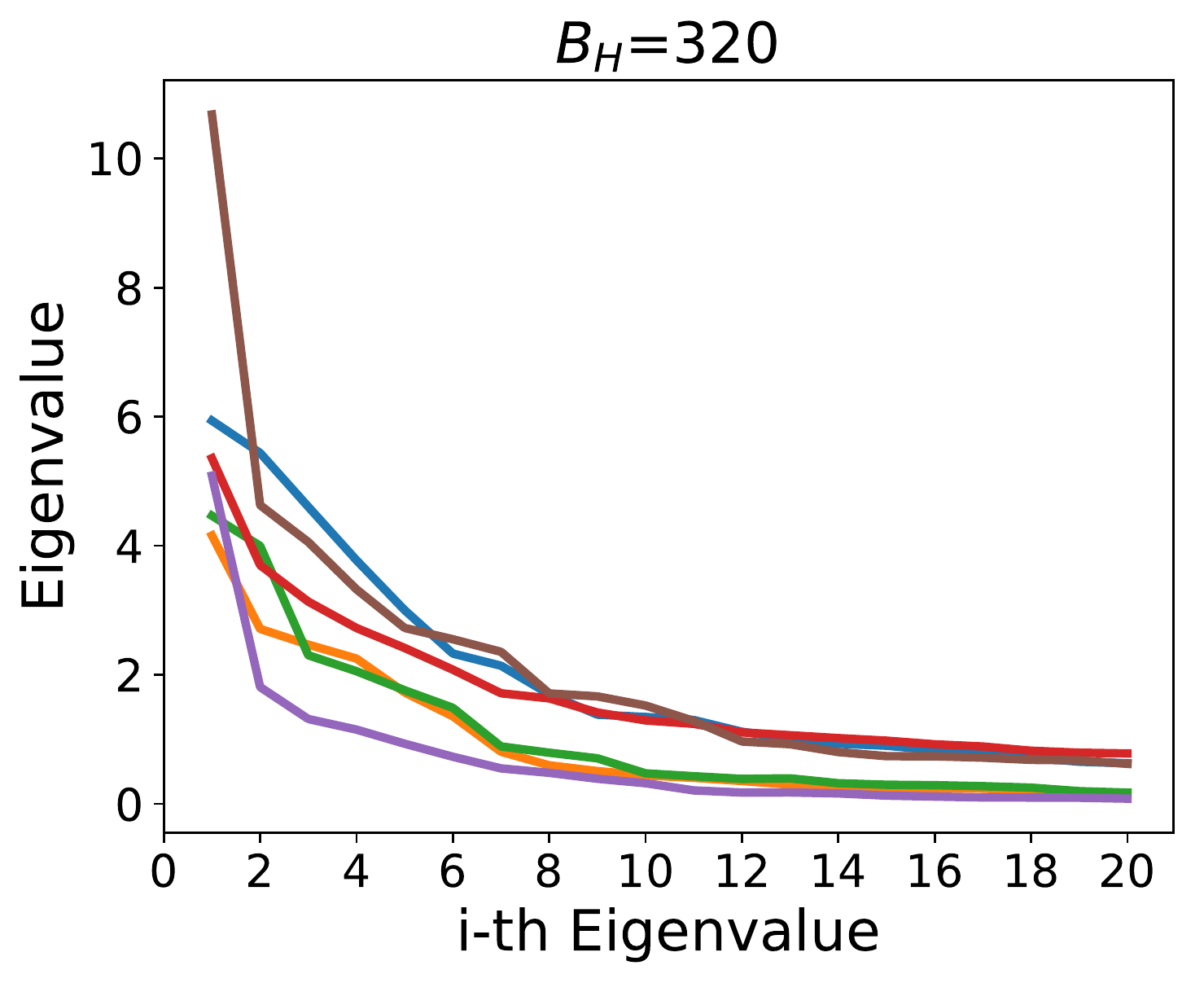}
\includegraphics[width=.48\textwidth]{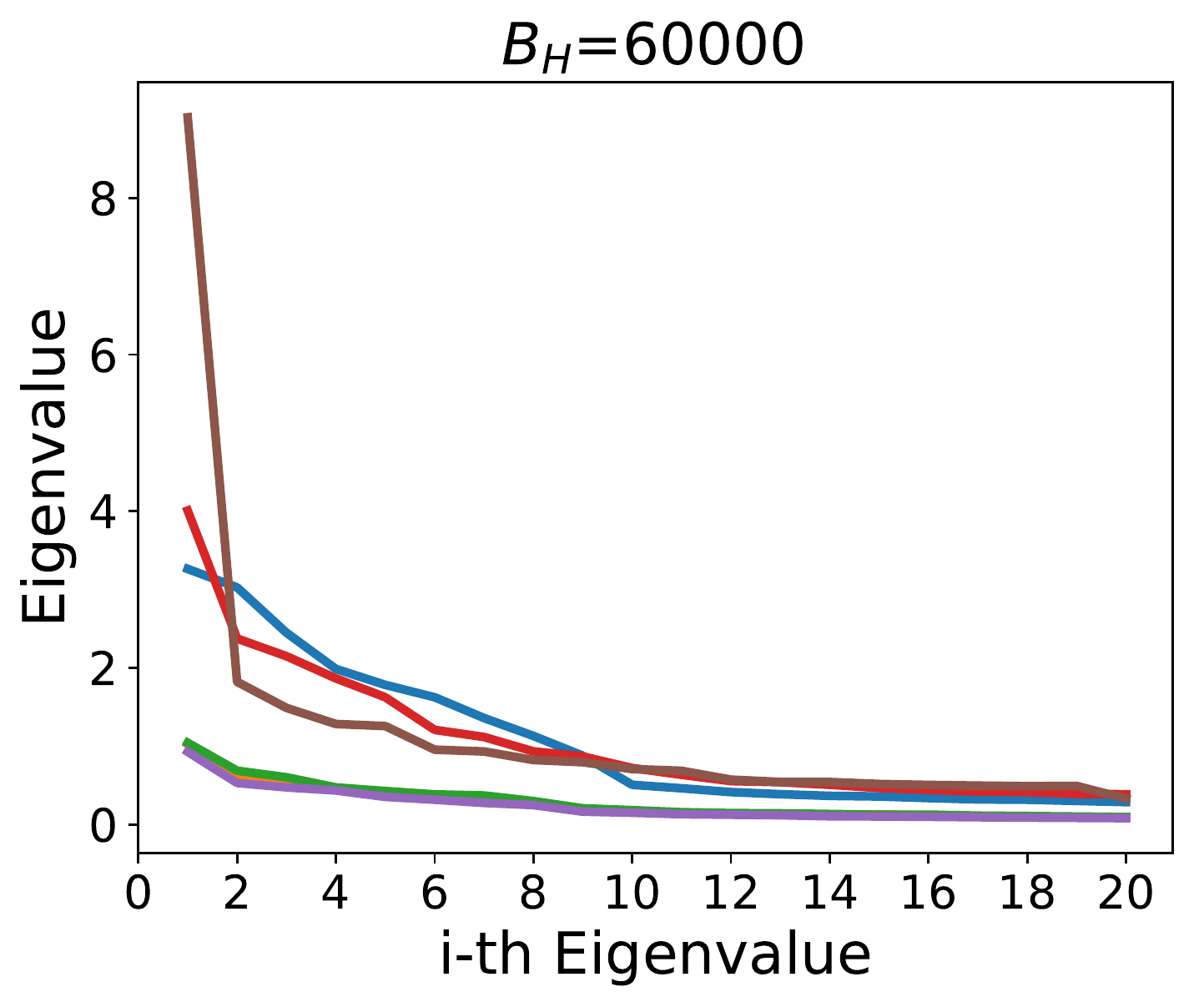}
\end{center}
\caption{
  Spectrum of the sub-sampled Hessian of the loss functional w.r.t. the model parameters computed by power iteration on MNIST of M1.
  The results are computed for different batch sizes of $B=1$, $B=320$, and $B=60000$. 
  We report two cases for the single batch experiment, which is drawn randomly from the clean training data. 
  The results show that the sub-sampled Hessian spectrum decreases for robust models.
  An interesting observation is that for the MNIST dataset, the original model has actually converged to a saddle point, even
  though it has a good generalization error.
  Also notice that the results for $B=320$ and $B=60,000$ are relatively close, which hints that the curvature for the full Hessian
  should also be smaller for the robust methods. This is demonstrated in~\fref{f:mnist_eig}.
}
\label{f:mnist_adv_topeig}
\end{figure*}
% ---------------------------------------------%
% ---------------------------------------------%
\begin{figure*}[tbp]
\begin{center}
  \includegraphics[width=.45\textwidth]{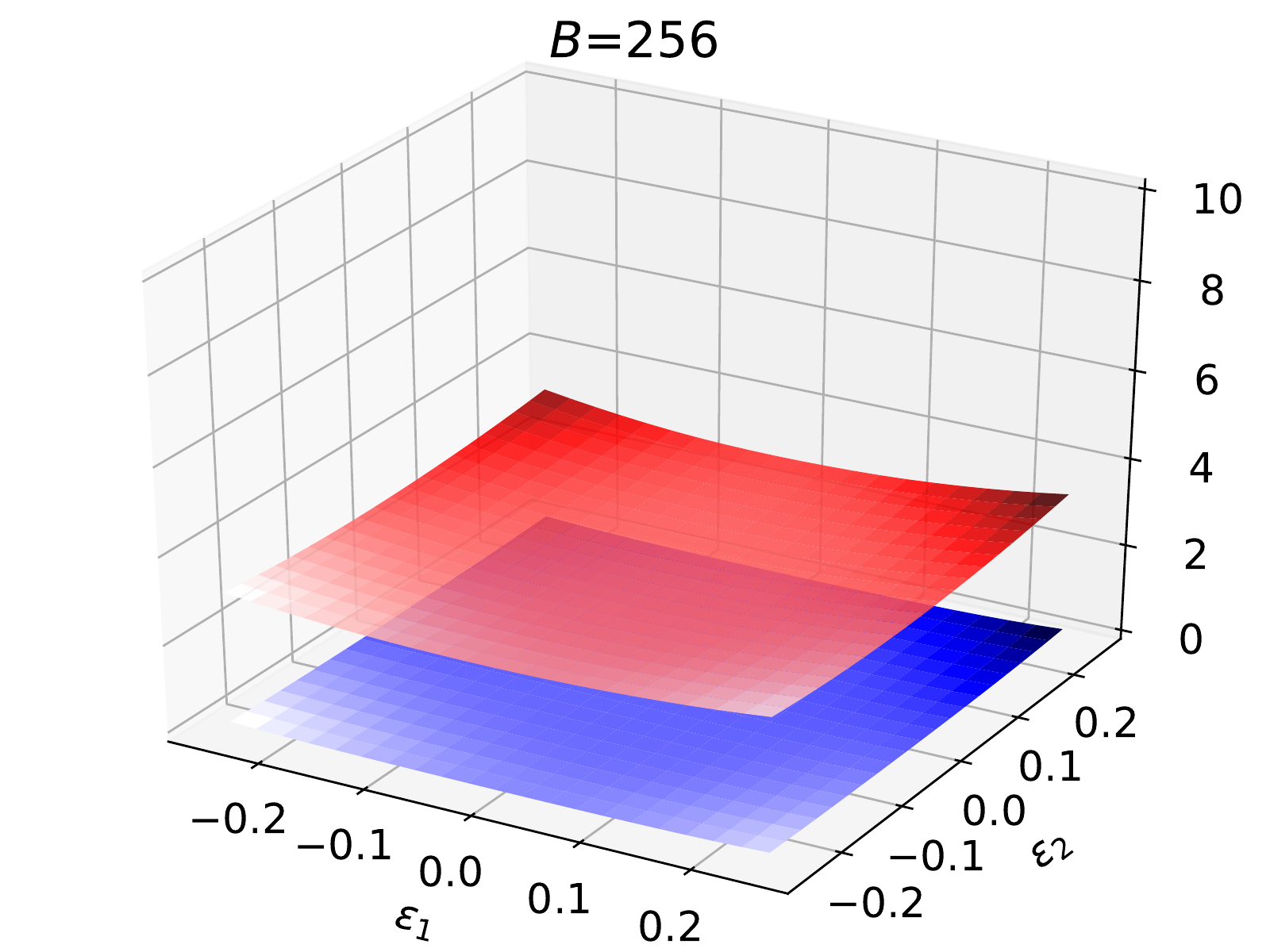}
  \includegraphics[width=.45\textwidth]{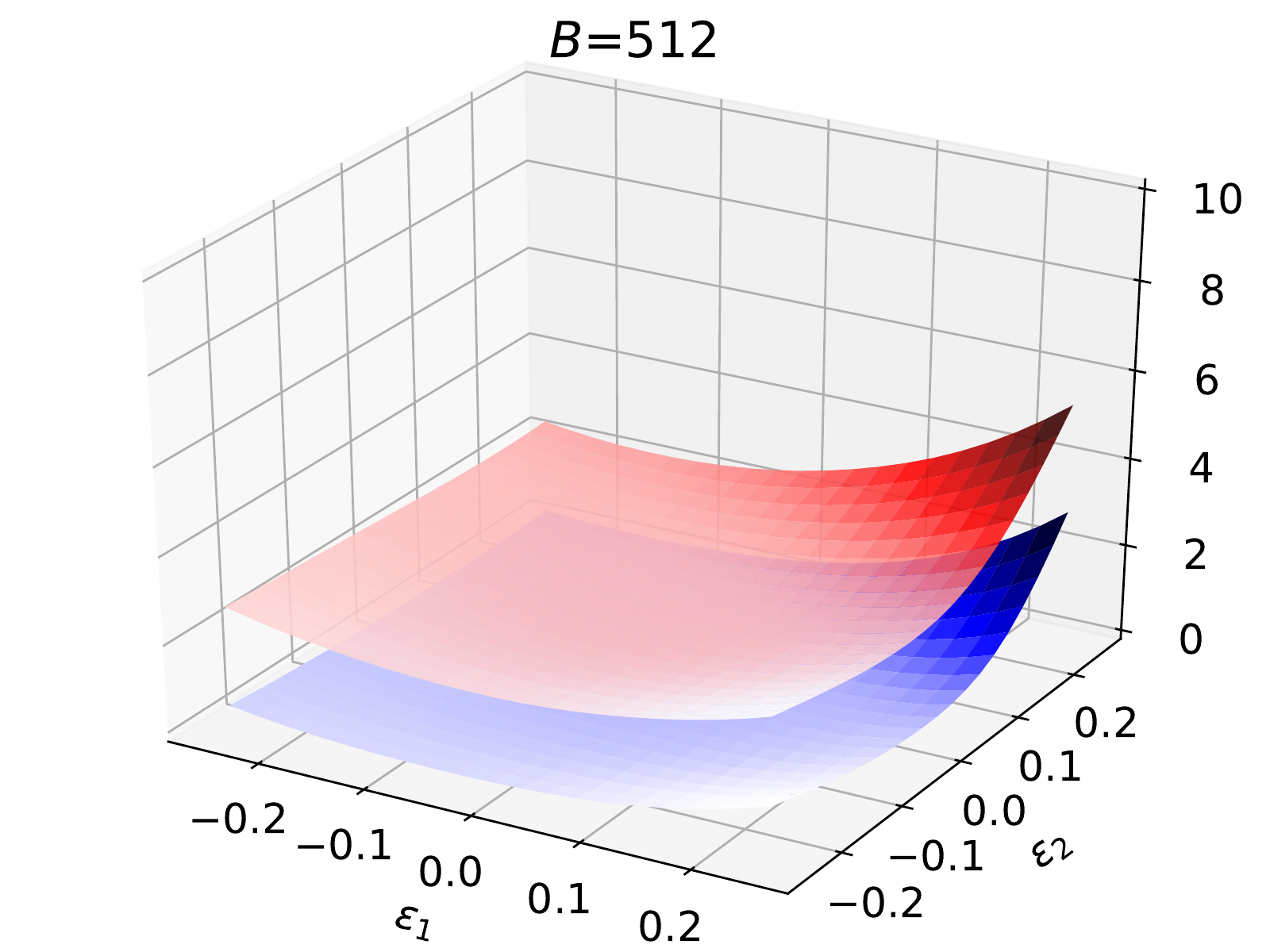}\\
  \includegraphics[width=.45\textwidth]{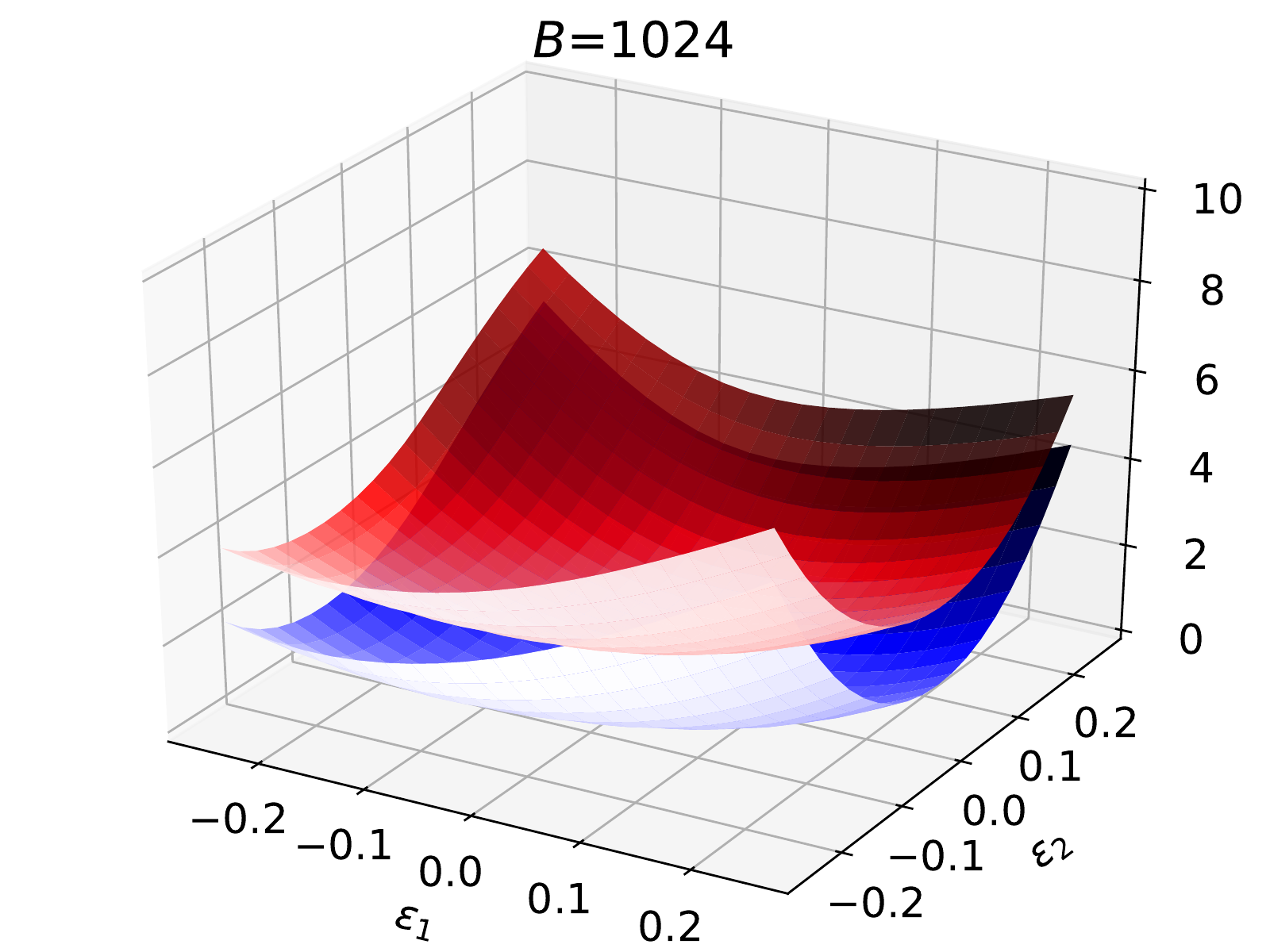}
  \includegraphics[width=.45\textwidth]{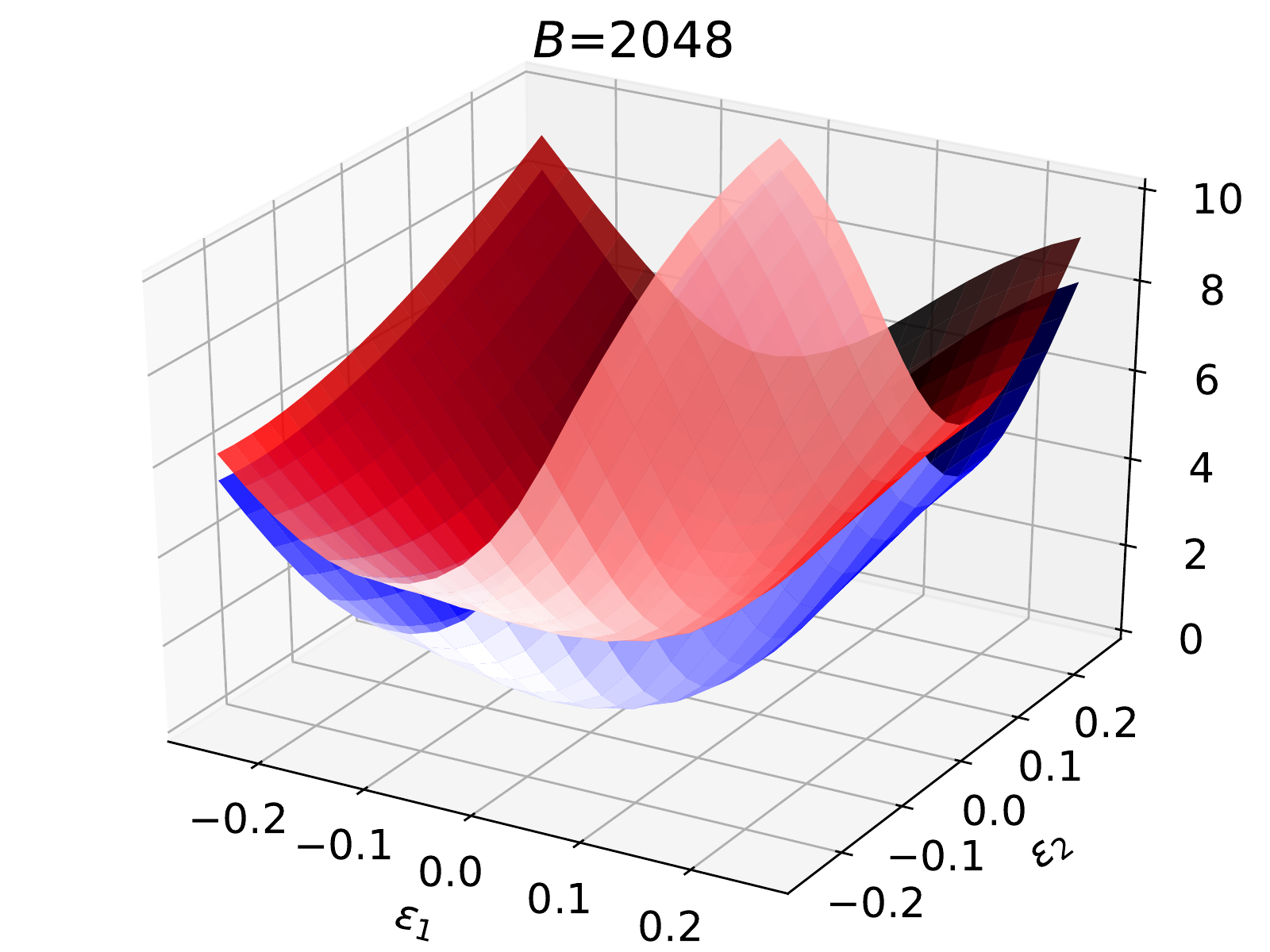}
\end{center}
\caption{
  The landscape of the loss functional is shown when the C2 model parameters are changed along the first two  dominant eigenvectors of the Hessian.
  Here $\epsilon_1,\ \epsilon_2$ are scalars that perturbs the model parameters along the first and second dominant eigenvectors.
}
\label{f:surface_largebatch_anet_app}
\end{figure*}
% ---------------------------------------------%
% ---------------------------------------------%
\begin{figure*}[tbp]
\begin{center}
  \includegraphics[width=.31\textwidth]{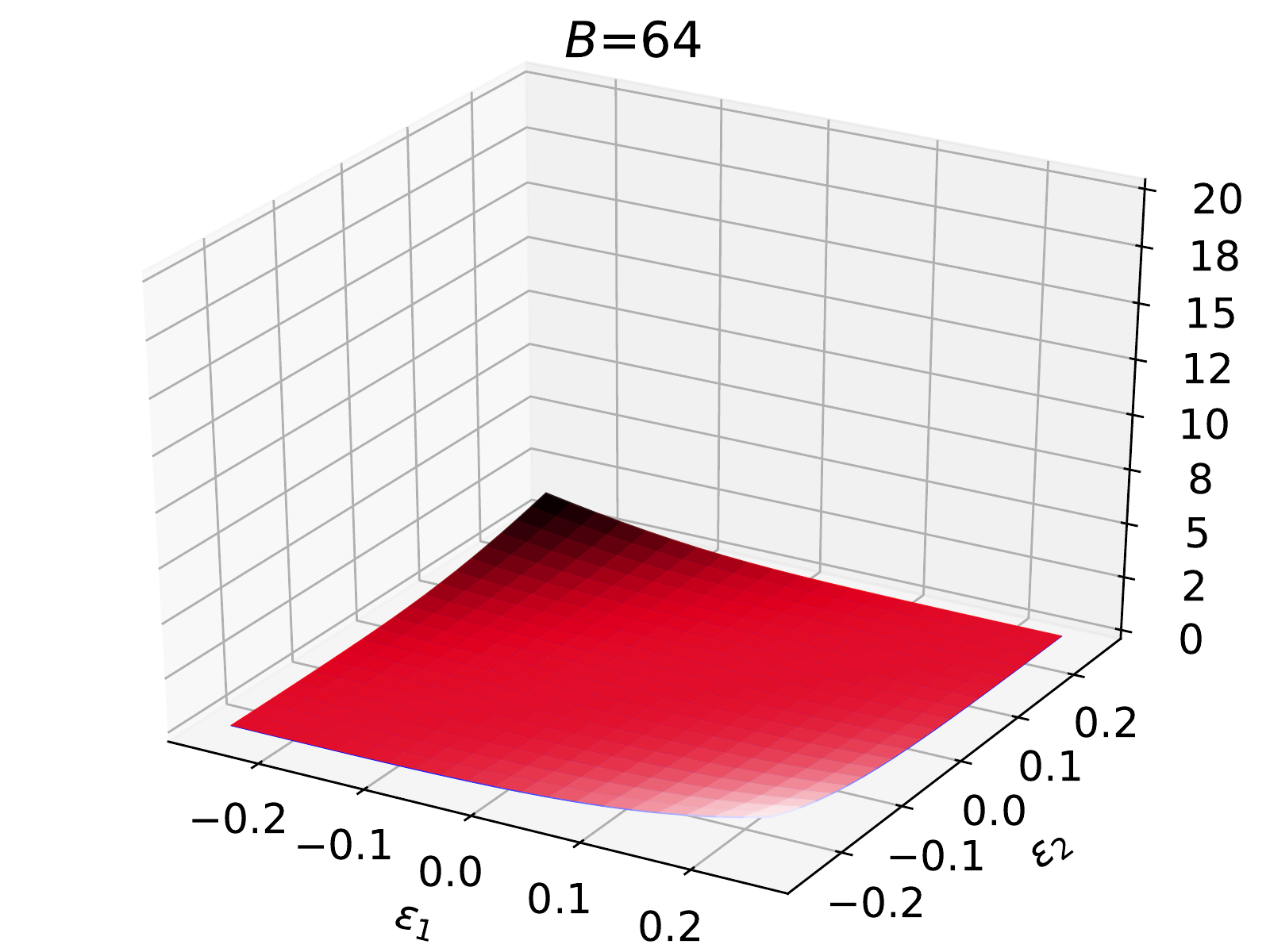} 
  \includegraphics[width=.31\textwidth]{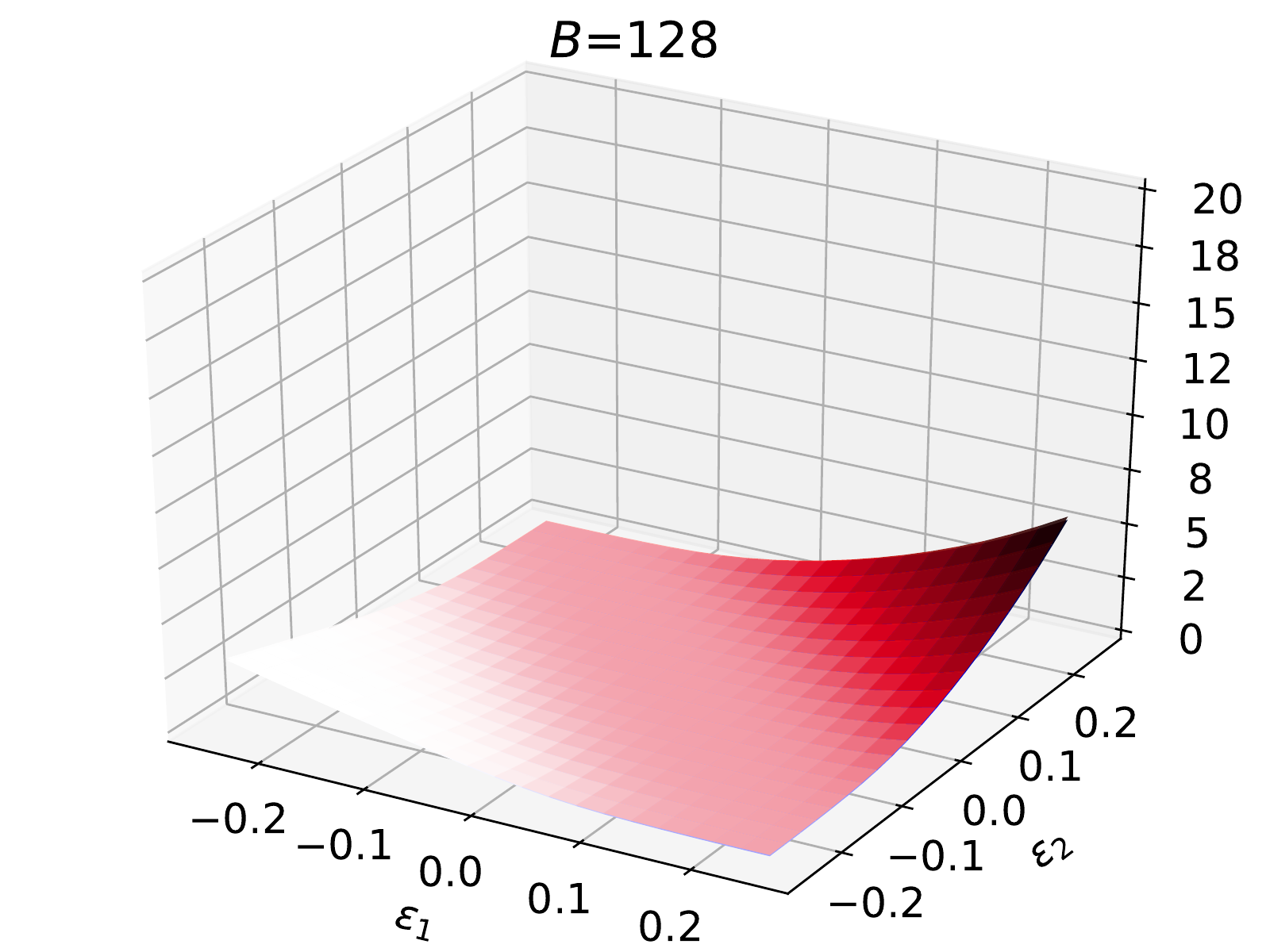}
  \includegraphics[width=.31\textwidth]{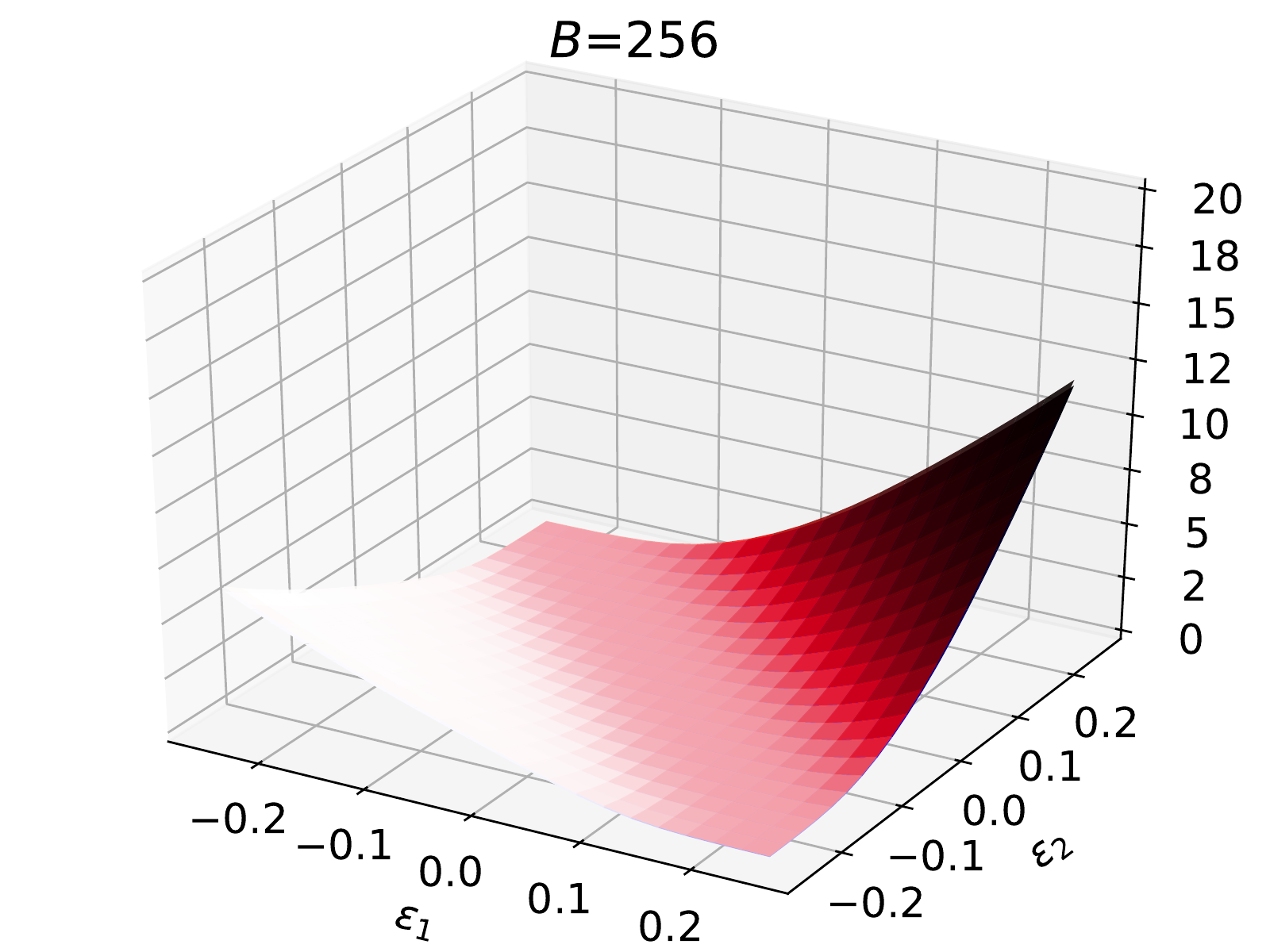}\\
  \includegraphics[width=.31\textwidth]{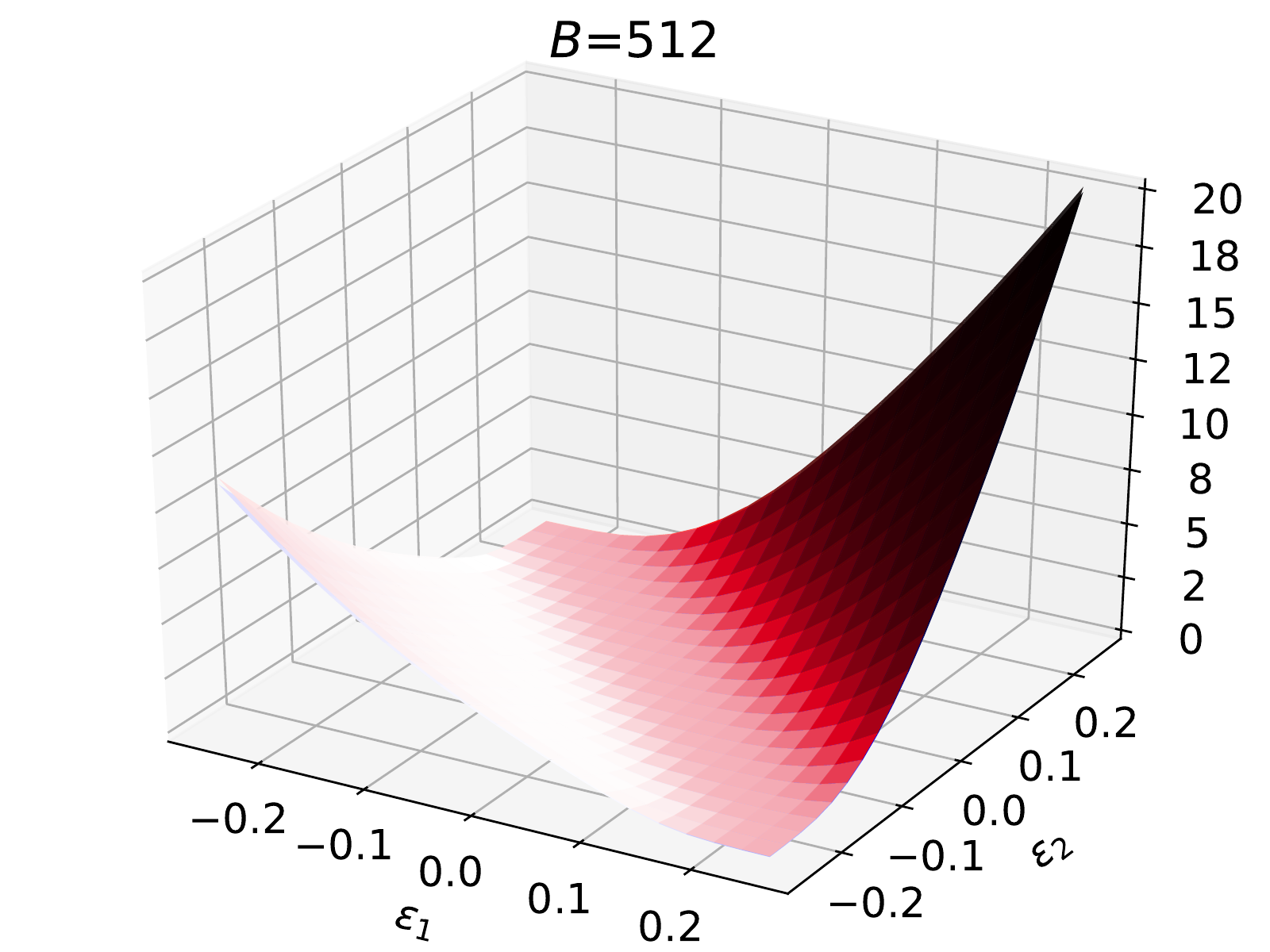}
  \includegraphics[width=.31\textwidth]{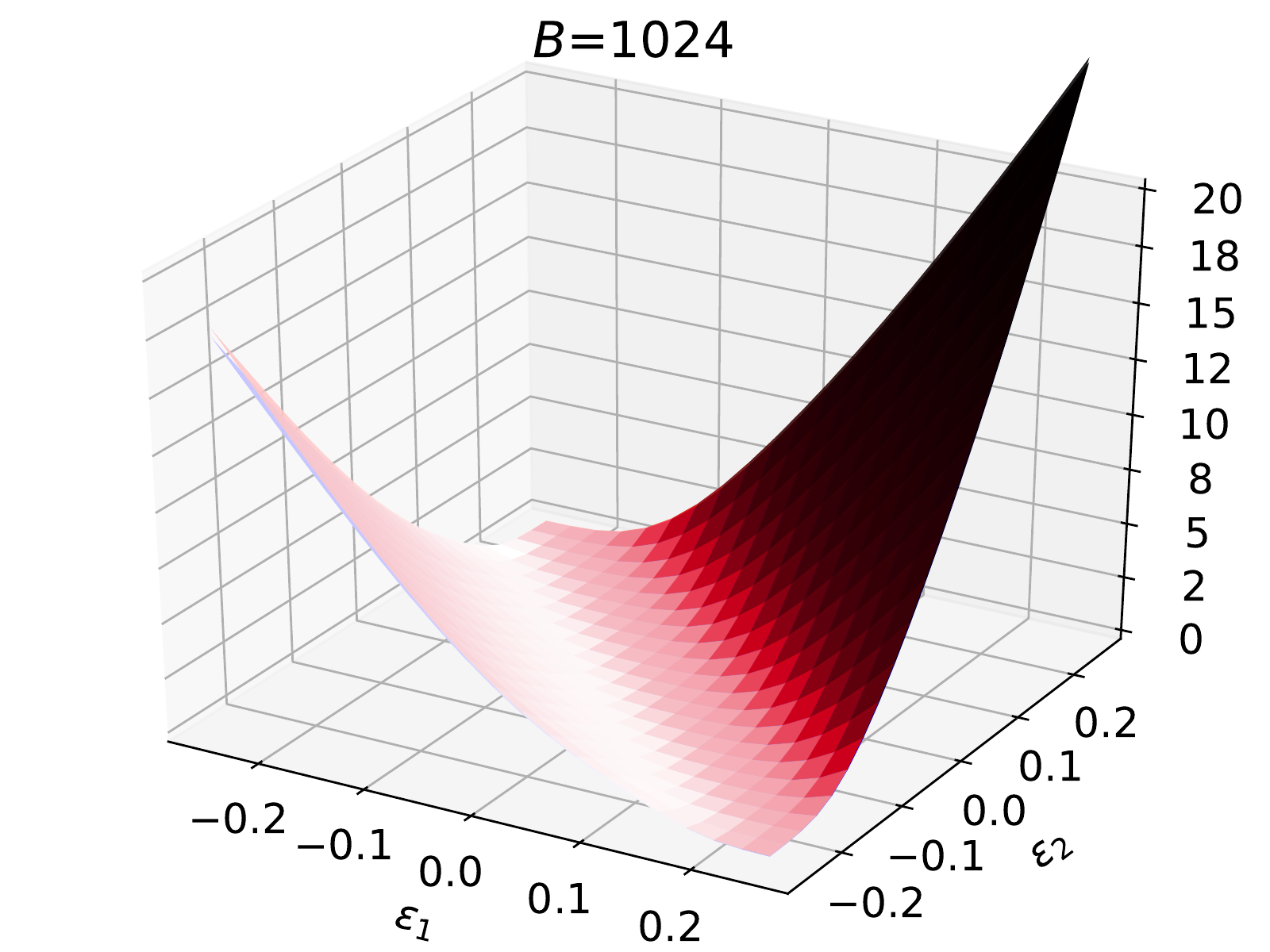}
  \includegraphics[width=.31\textwidth]{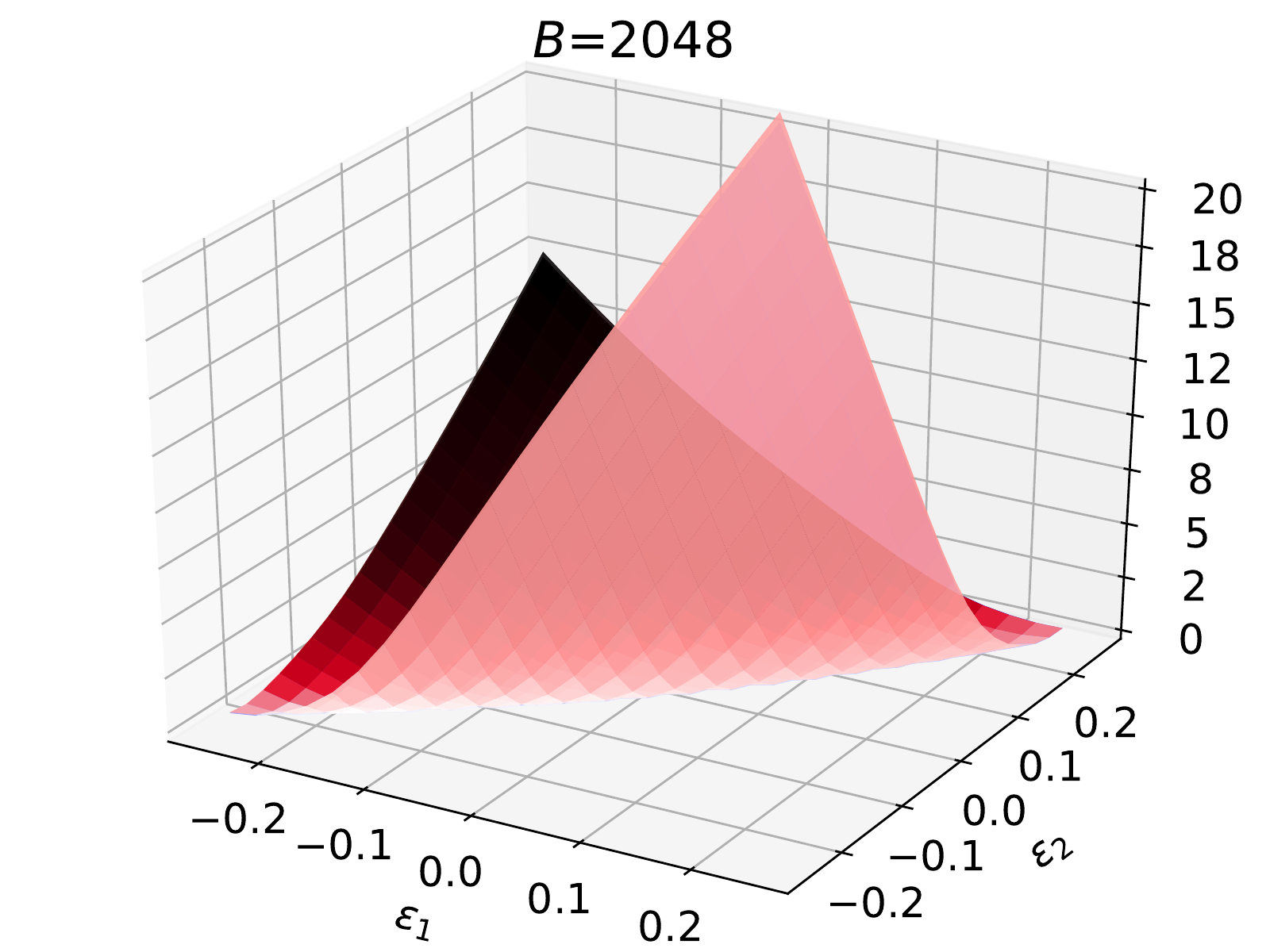}
\end{center}
\caption{
  The landscape of the loss functional is shown when the M1 model parameters are changed along the first two dominant eigenvectors of the Hessian.
  Here $\epsilon_1,\ \epsilon_2$ are scalars that perturbs the model parameters along the first and second dominant eigenvectors.
}
\label{f:surface_largebatch_mnist_app}
\end{figure*}
% ---------------------------------------------%
% ---------------------------------------------%
\begin{figure*}[tbp]
\begin{center}
  \includegraphics[width=.45\textwidth]{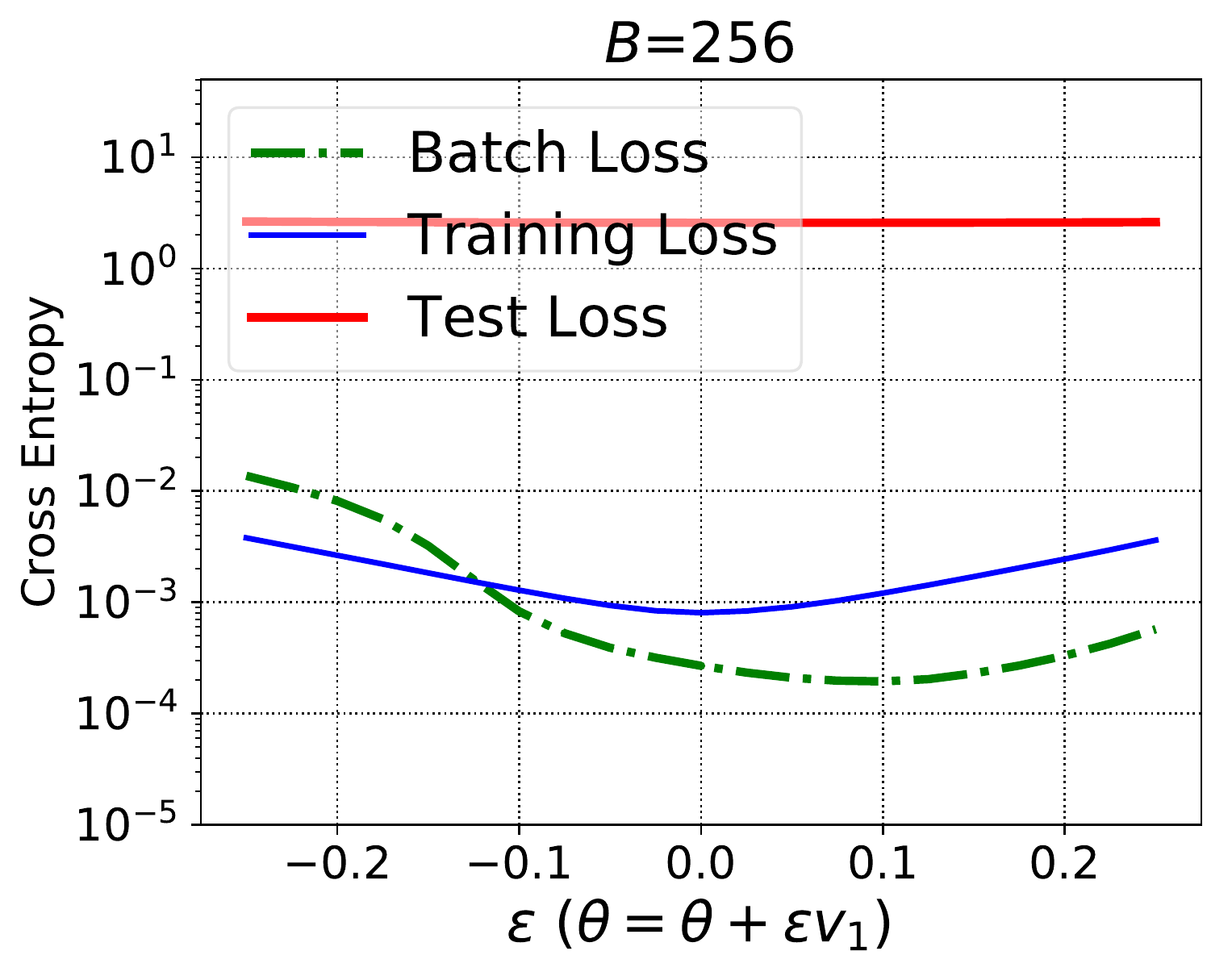}
  \includegraphics[width=.45\textwidth]{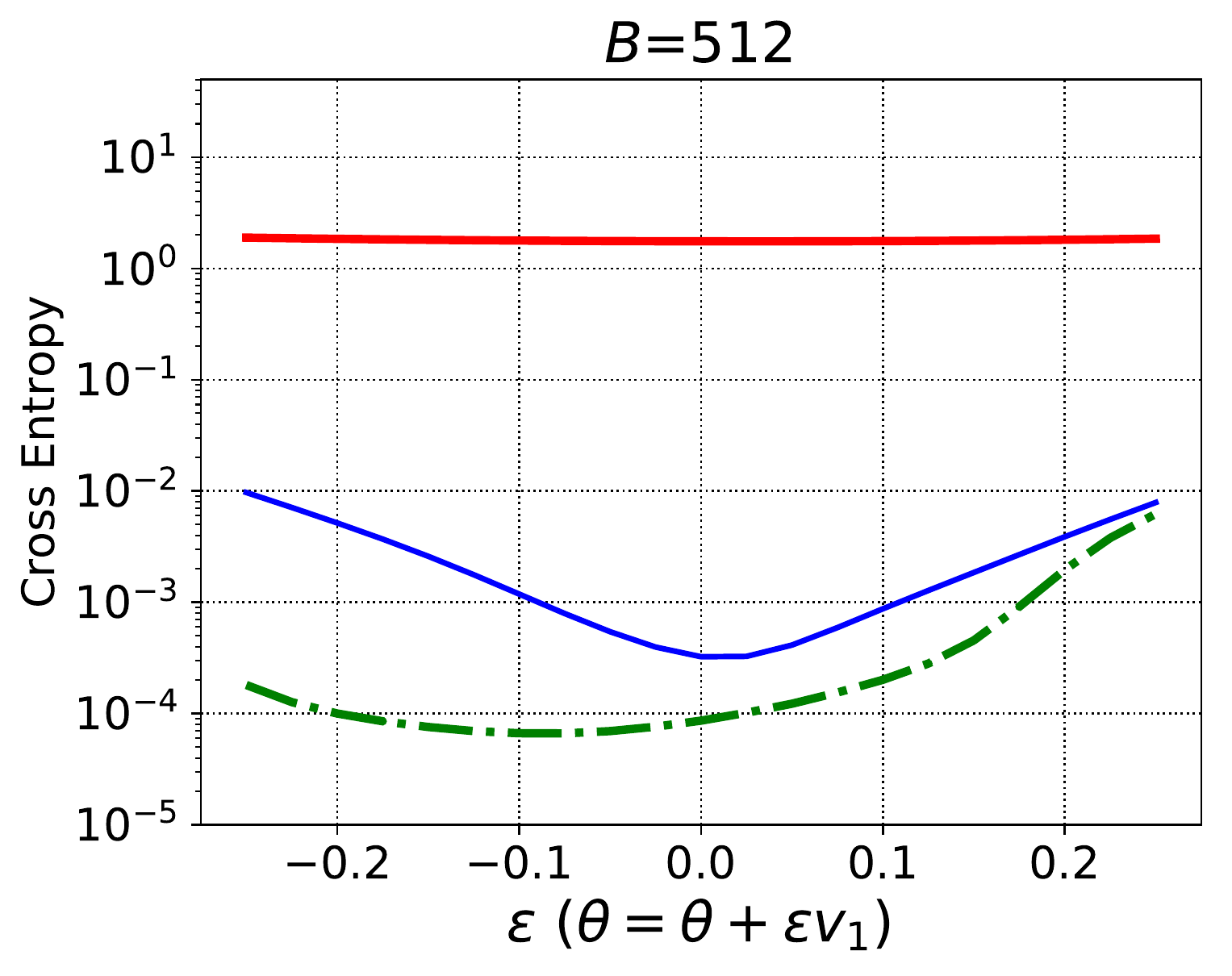}\\
  \includegraphics[width=.45\textwidth]{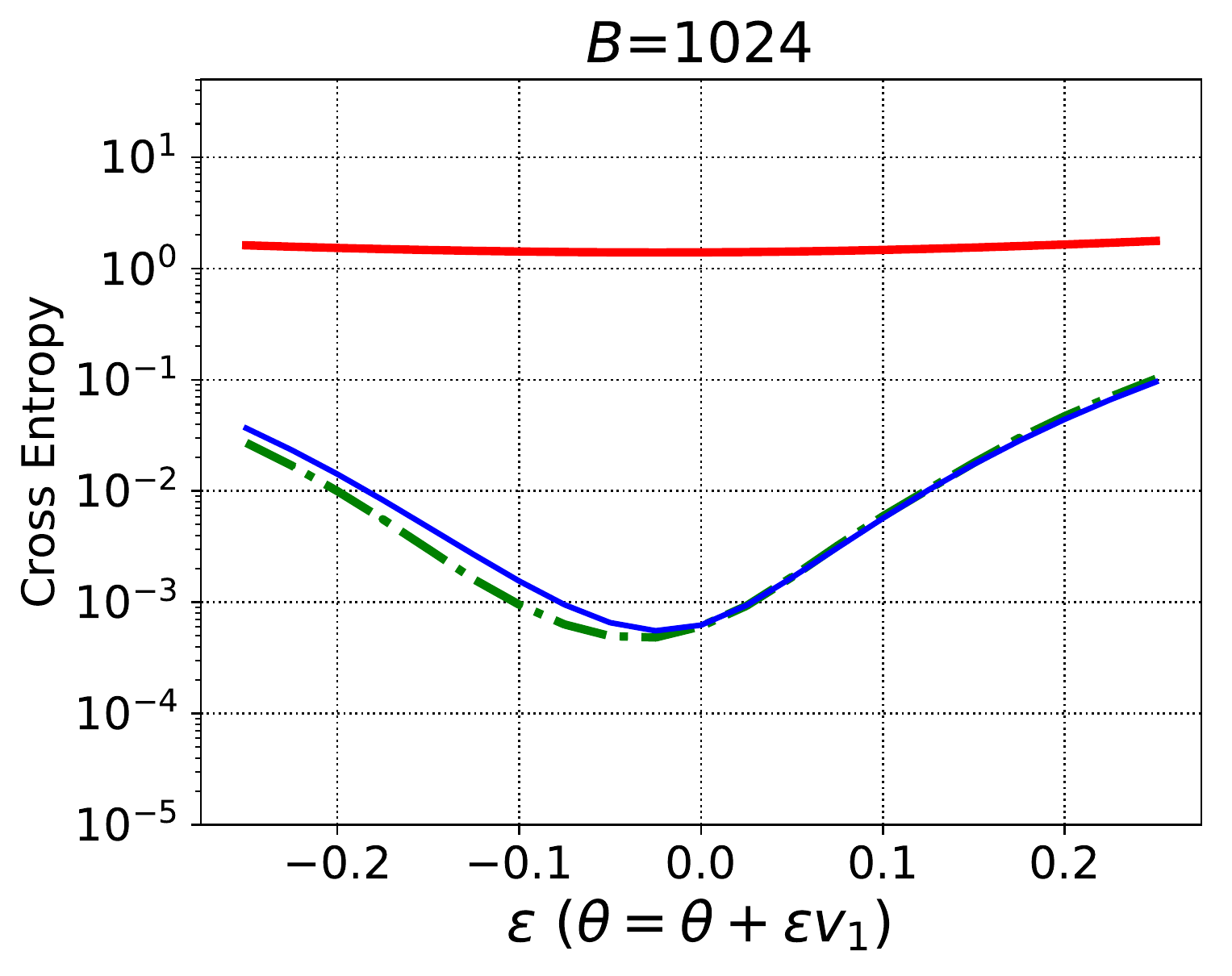}
  \includegraphics[width=.45\textwidth]{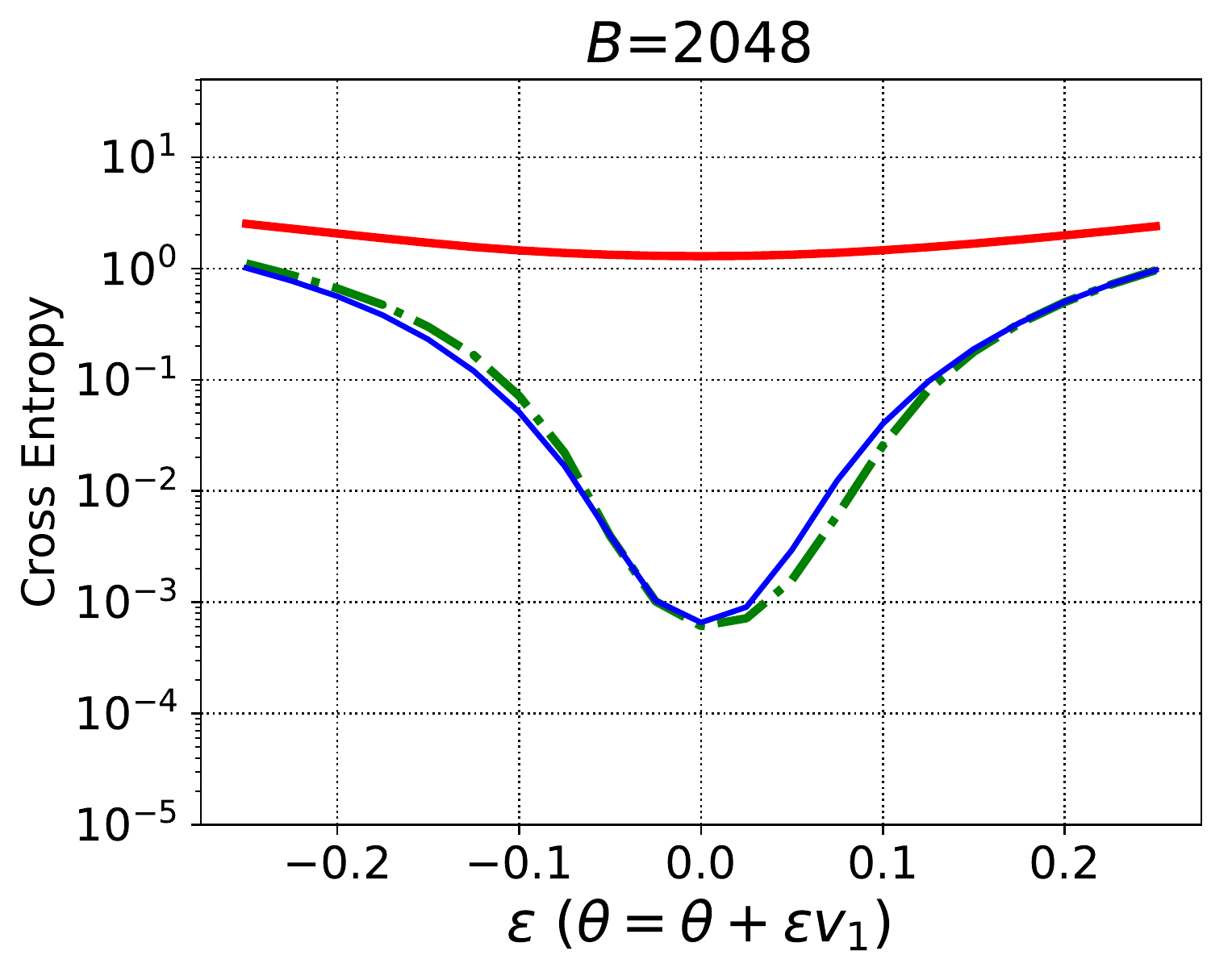}
\end{center}
\caption{
  The landscape of the loss functional is shown along the dominant eigenvector of the Hessian for C2
 architecture on CIFAR-10 dataset.
  Here $\epsilon$ is a scalar that perturbs the model parameters along the dominant eigenvector denoted by $v_1$.
}
\label{f:landscape_largebatch_anet_app}
\end{figure*}

%\begin{figure*}[tbp]
%  \centering

% ---------------------------------------------%
% ---------------------------------------------%
\begin{figure*}[tbp]
  \centering
\includegraphics[width=.32\textwidth]{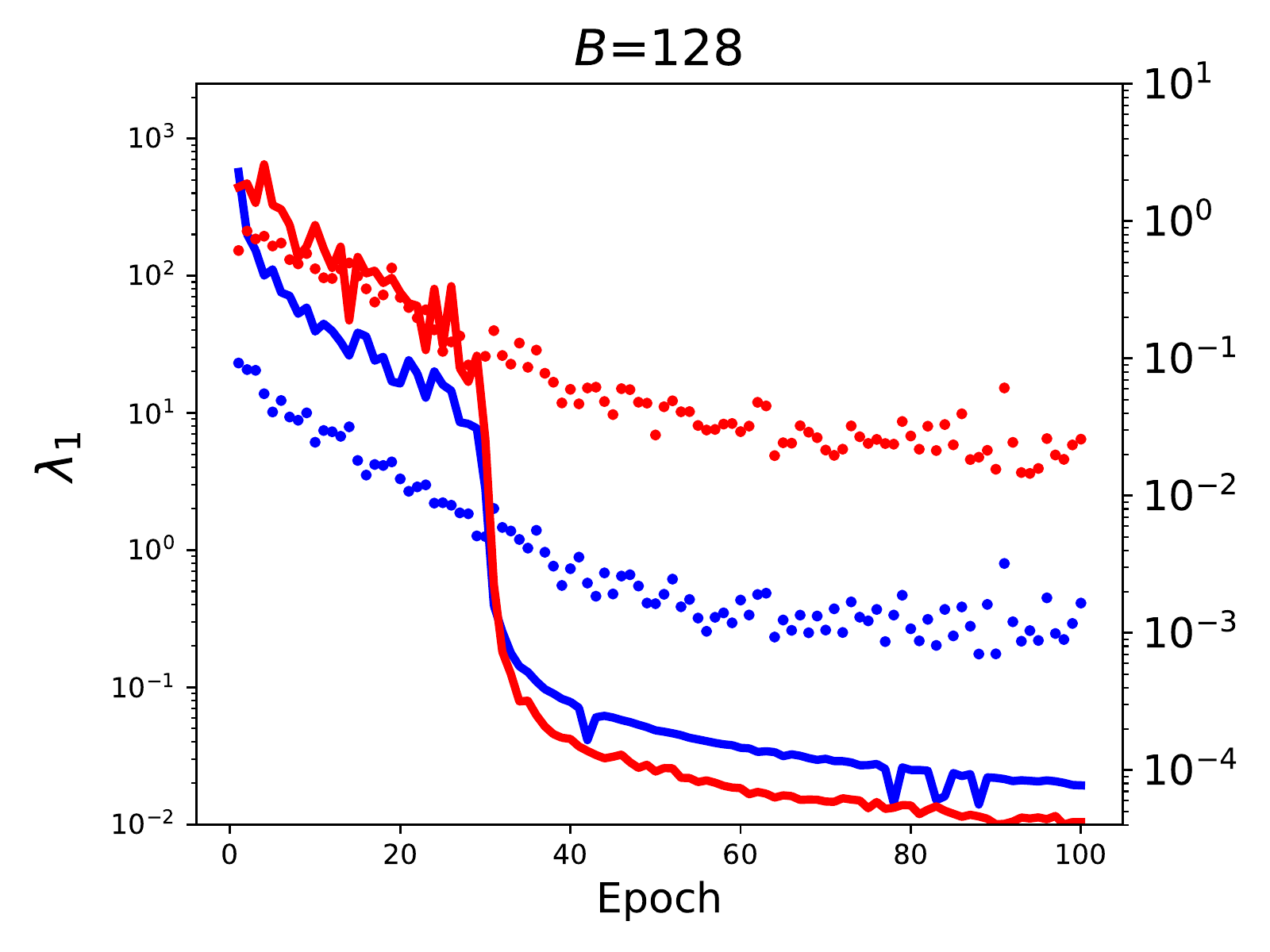} 
\includegraphics[width=.32\textwidth]{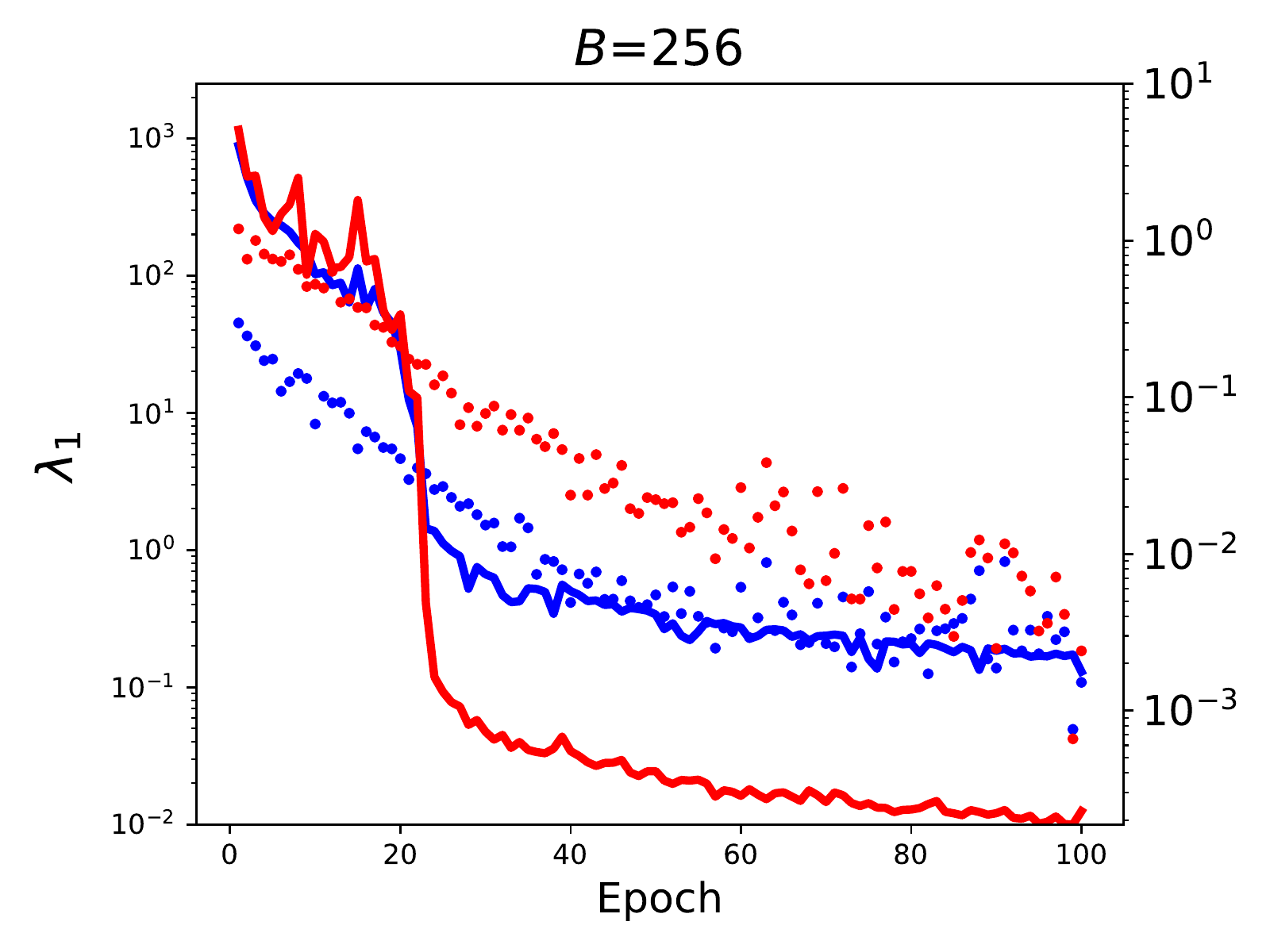}
\includegraphics[width=.32\textwidth]{figures/net100_params_b_512_eta_0.02_robust.pdf}\\
\includegraphics[width=.32\textwidth]{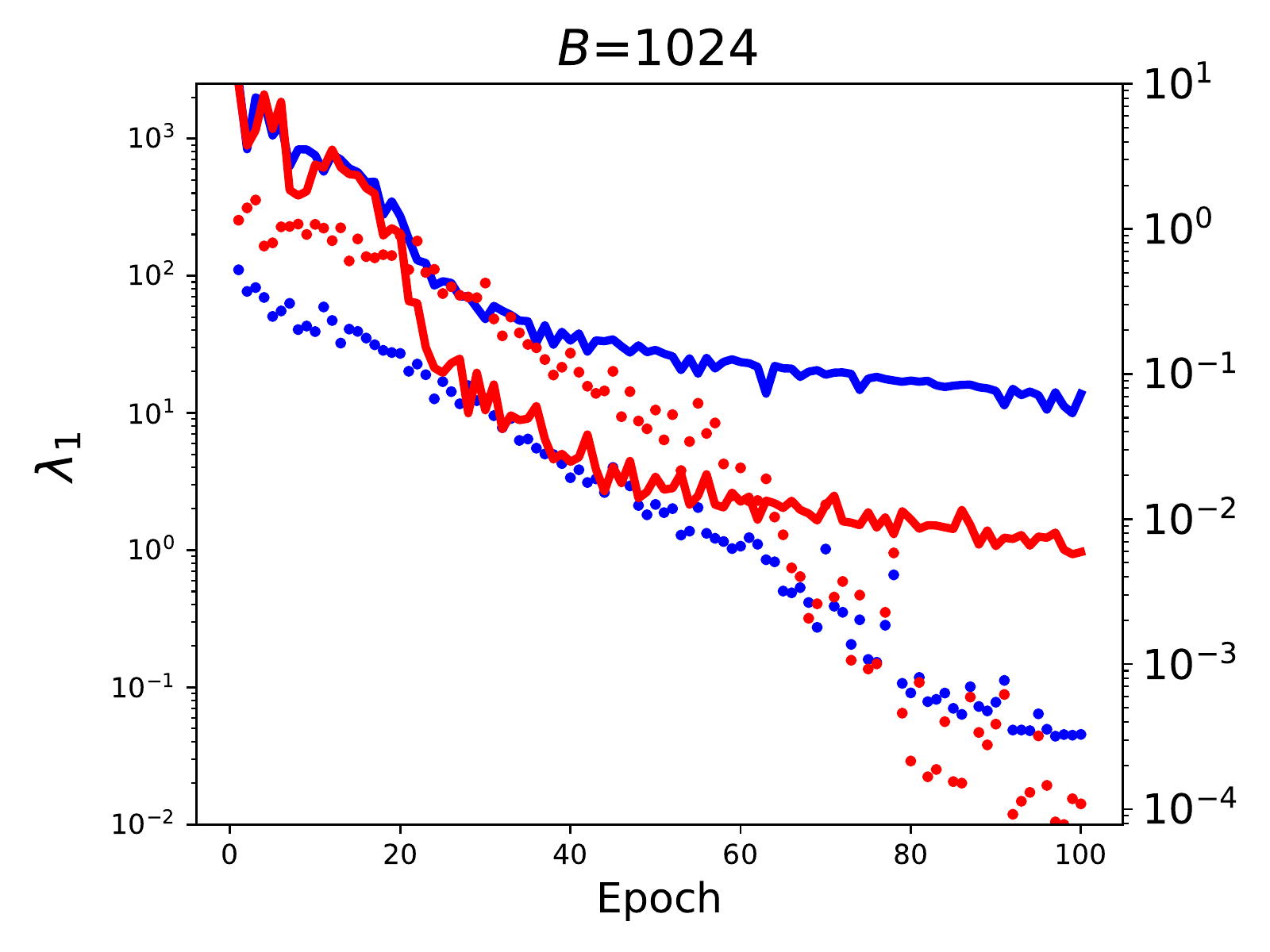}
\includegraphics[width=.32\textwidth]{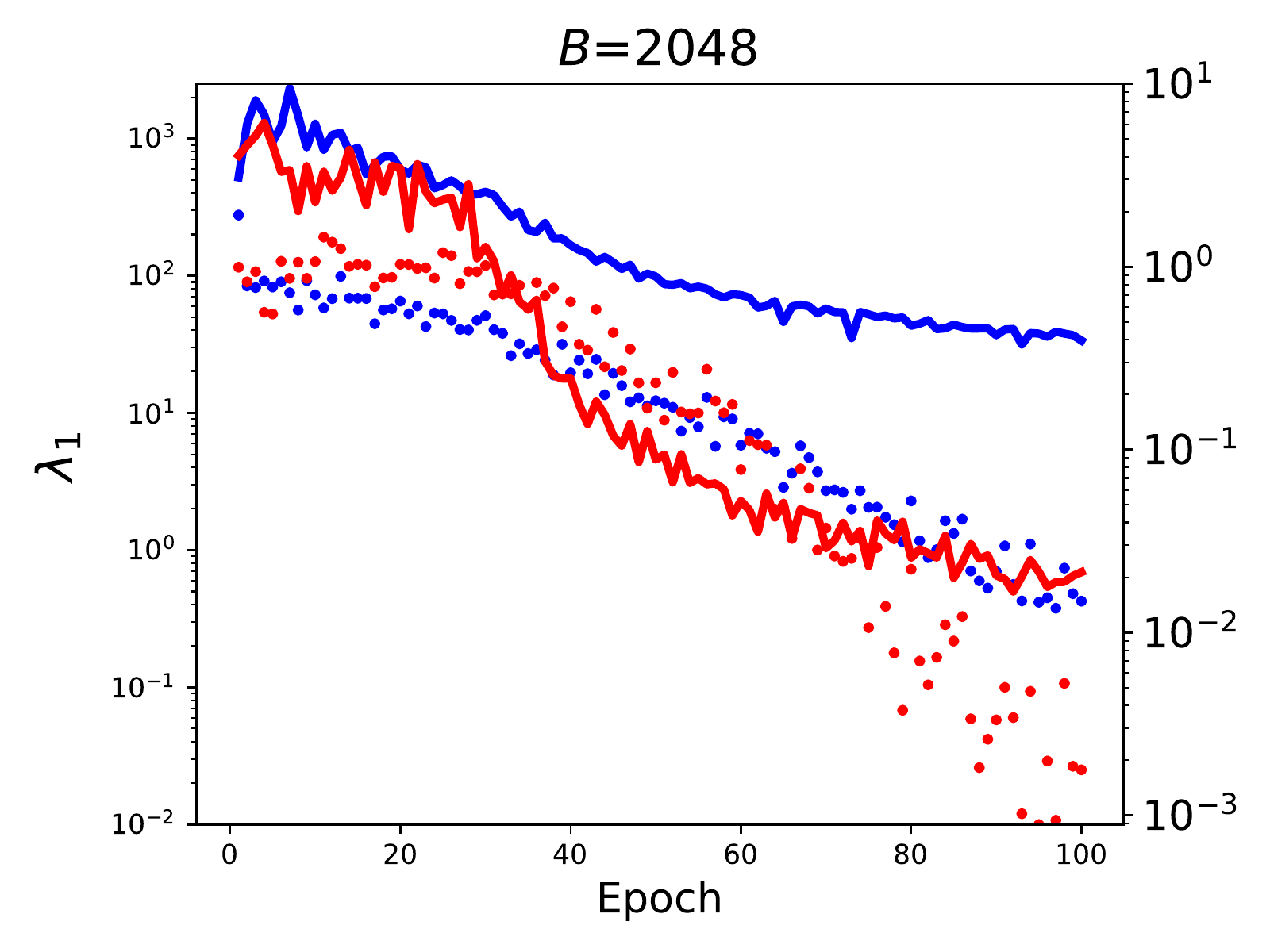}
\includegraphics[width=.32\textwidth]{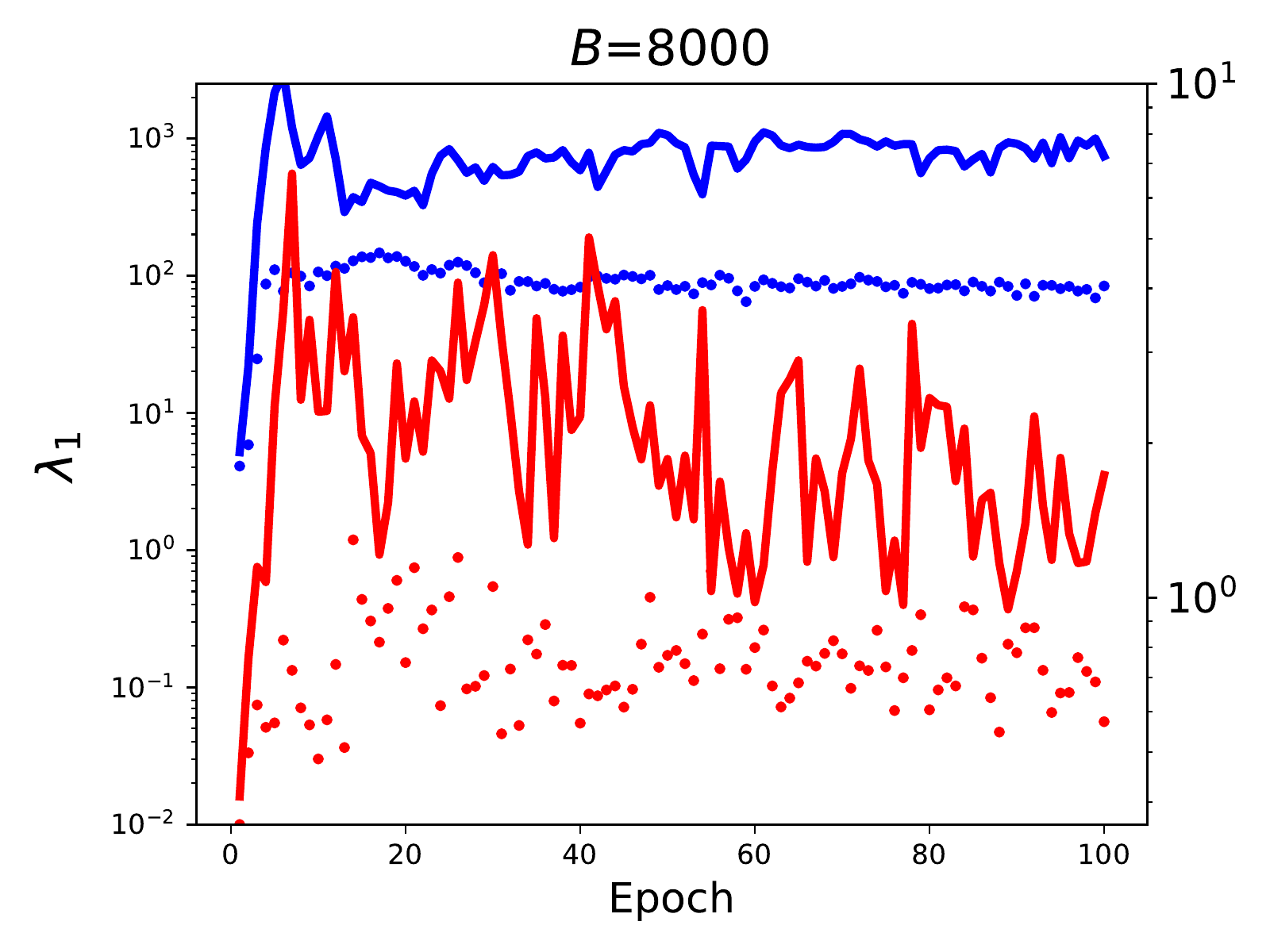}\\
\includegraphics[width=.32\textwidth]{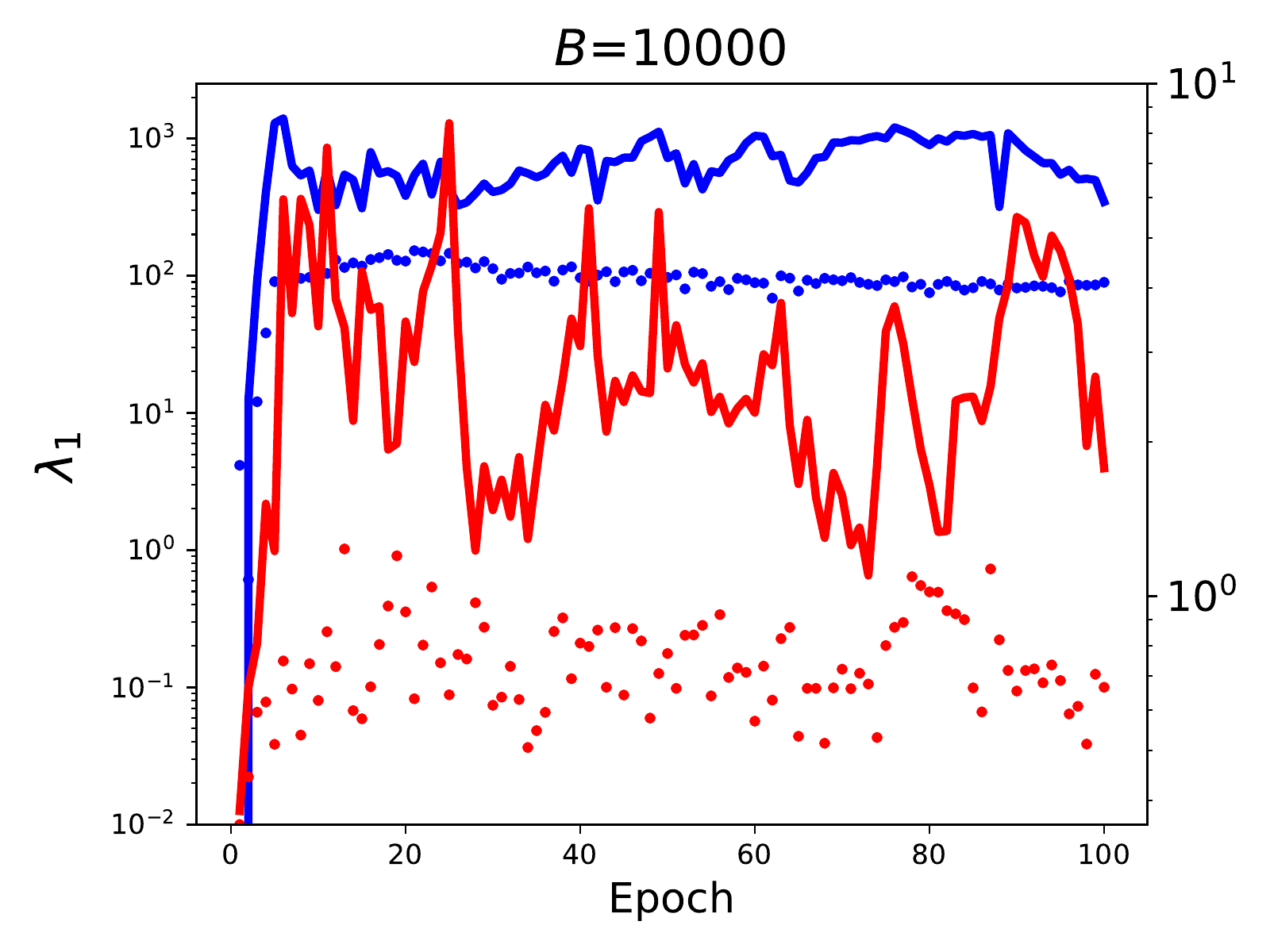}
\includegraphics[width=.32\textwidth]{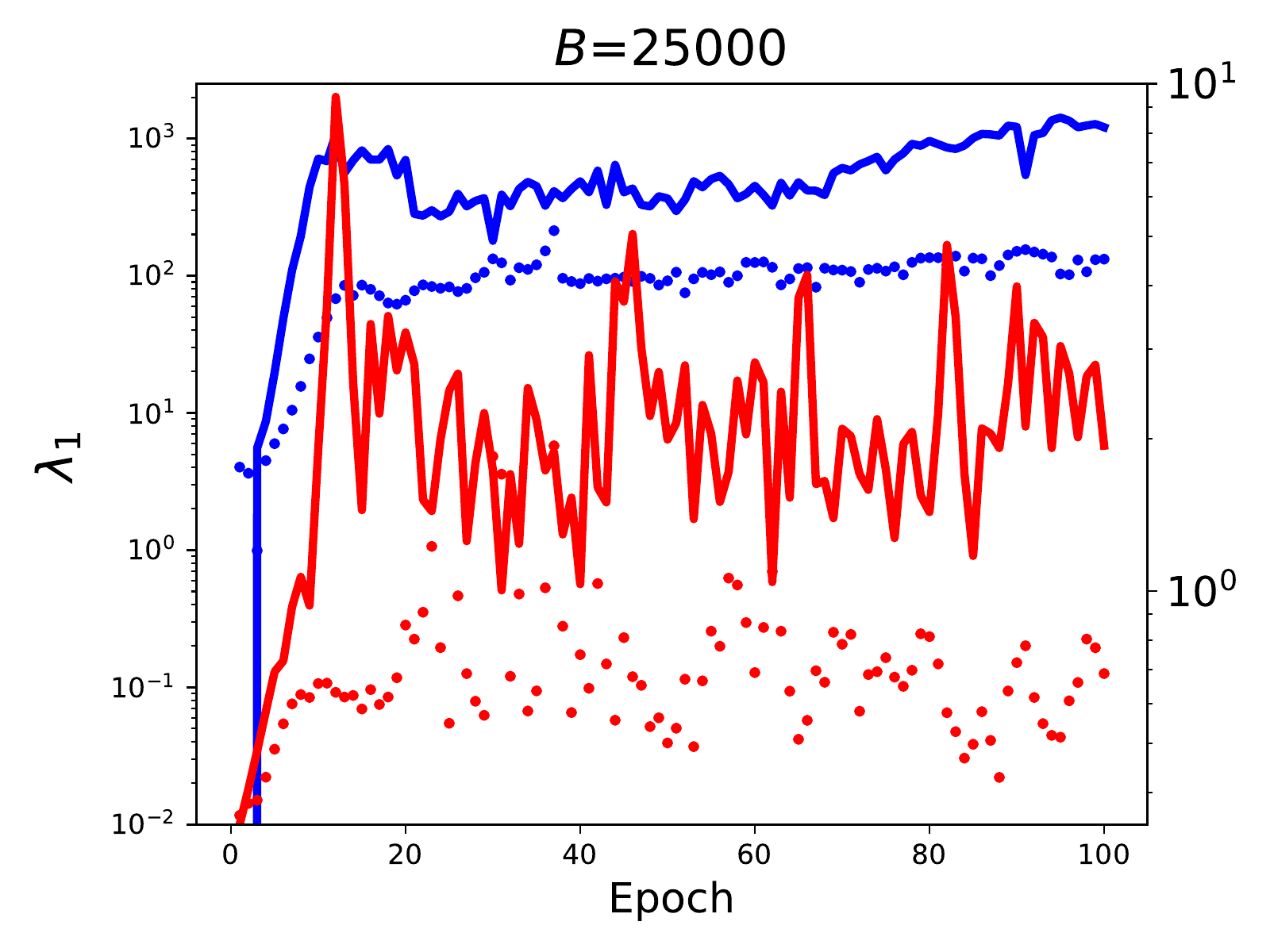}
\includegraphics[width=.32\textwidth]{figures/net100_params_b_50000_eta_0.02_robust.pdf}
\caption{
  Changes in the dominant eigenvalue of the Hessian w.r.t weights and the total gradient is shown for
  different epochs during training. Note the increase in $\lambda_1^{\theta}$ (blue curve) for large batch
  vs small batch. In particular, note that the values for total gradient along with the Hessian spectrum show
  that large batch does not get ``stuck'' in
  saddle points, but areas in the optimization landscape that have high curvature.
  The dotted points show the corresponding results when we use robust optimization. We can see that this
  pushes the training to flatter areas. This clearly demonstrates the potential to use robust optimization
  as a means to avoid sharp minimas.
}
\label{fig:qalex_h_logger_appendix}
\end{figure*}
% ---------------------------------------------%
% ---------------------------------------------%

% ---------------------------------------------%
\begin{table*}[h]
\centering
\caption{
Baseline accuracy is shown for large batch size for C1 model along with results aciheved with
scaling learning rate method proposed by~\cite{goyal2017accurate} (denoted by "FB Acc"). The last column
shows results when training is performed with robust optimization. As we can see, the performance
of the latter is actually better for large batch size. We emphasize that the goal is to perform analysis
to better understand the problems with large batch size training.
More extensive tests are needed before one could claim that robust optimization performs better
than other methods.
}
\label{t:robust_potential}
\small
\begin{tabular}{c|c|c|c}
\toprule
 Batch  	&	Baseline Acc &	FB Acc & Robust Acc \\
\midrule
\Ga  8000  	&	     0.7559  &	0.752  & 0.7612     \\
\Gc 10000  	&	     0.7561  &	0.1    & 0.7597     \\
\Ga 25000  	&	     0.7023  &	0.1    & 0.7409     \\
\Gc 50000  	&	     0.5523  &	0.1    & 0.7116     \\
\bottomrule
\end{tabular}
\end{table*}

\end{document}